\documentclass[letterpaper]{article}
\usepackage[preprint]{aaai2027}
\usepackage[hyphens]{url}
\usepackage{graphicx}
\usepackage{natbib}
\usepackage{caption}

\usepackage{booktabs}
\usepackage{amsmath}
\usepackage{amssymb}
\usepackage{xcolor}
\usepackage{array}
\usepackage{tabularx}
\usepackage{colortbl}
\usepackage{adjustbox}
\usepackage{multirow}
\usepackage{colortbl}

\usepackage{booktabs}
\usepackage{xcolor}
\usepackage{colortbl}
\usepackage{array}
\usepackage{tabularx}

\usepackage{booktabs}
\usepackage{array}
\usepackage{tabularx}
\usepackage{graphicx}
\usepackage{booktabs}
\usepackage{amsmath}
\usepackage{amssymb}
\usepackage{xcolor}
\usepackage{array}
\usepackage{tabularx}
\usepackage{colortbl}
\usepackage{adjustbox}
\usepackage{multirow}
\usepackage{url}
\usepackage{float}
\usepackage{longtable}

\newcolumntype{Y}{>{\raggedright\arraybackslash}X}
\definecolor{RuleGray}{RGB}{92,98,104}
\definecolor{SuppBlue}{HTML}{EAF2FF}
\definecolor{SuppBlueDark}{HTML}{2F5597}
\definecolor{SuppGray}{HTML}{F6F8FA}
\definecolor{SuppGreen}{HTML}{EFF8F3}
\definecolor{SuppOrange}{HTML}{FFF7E8}
\definecolor{SuppPurple}{HTML}{F3EEFF}
\definecolor{SuppLine}{HTML}{D0D7DE}
\newcolumntype{Y}{>{\raggedright\arraybackslash}X}
\newcommand{\promptbox}[1]{%
  \fcolorbox{SuppLine}{white}{%
    \parbox{0.94\linewidth}{\scriptsize\ttfamily\raggedright\sloppy #1}}}
\providecommand{\affiliations}[1]{}
\newcommand{\manuscriptversion}{ICD-Deepresearch-clean-technical-supplement}
\newcolumntype{Y}{>{\raggedright\arraybackslash}X}

\title{Foundation Agents Meet Agentic Deep Research: Evidence-Grounded Clinical Code Forecasting}
\author {
    Junda Wang\textsuperscript{\rm 1},
    Meysam Ghaffari\textsuperscript{\rm 2},
    Akshat Choube\textsuperscript{\rm 2},
    Mohsen Sharifi Renani\textsuperscript{\rm 2},
    Hong Yu\textsuperscript{\rm 3},
    Carlos Morato\textsuperscript{2}
}
\affiliations {
    \textsuperscript{\rm 1}University of Massachusetts Amherst, MA, USA\\
    \textsuperscript{\rm 2}Optum, MN, USA\\
    \textsuperscript{\rm 3}University of Massachusetts Lowell, MA, USA\\
}

\begin{document}
\maketitle

\begin{abstract}
Next-encounter ICD forecasting predicts which standardized diagnosis codes
will be documented at a future visit from the longitudinal record available
beforehand. The task is prospective and multi-label: the target note does not
yet exist, and several codes may be correct. Structured EHR foundation models
capture recurrence and temporal progression, whereas language foundation
models generate flexible diagnostic hypotheses. We introduce
ICD-Deepresearch, a DeepResearch workflow that composes these
predictive foundation models with medical search and ICD dictionaries. Because no source
reveals the future code set, research evaluates candidate transitions by
linking patient evidence, external clinical relations, and exact code
semantics under a fixed top-$K$ budget. Candidate Generation uses SparseEHR to
produce an EHR Prior that initializes two bounded Research Expansion rounds;
an independent GPT-5 Direct Forecast supplies complementary candidates. Final
Selection validates, deduplicates, and jointly ranks both paths, after which a
separate module writes rationales without changing predictions. Finally
ICD-Deepresearch achieves patient-averaged precision/recall of
24.60/35.09\% on MIMIC-III and 25.14/48.32\% on MIMIC-IV. Physicians rate
51\% and 68\% of its retrieved documents useful, compared with 22\% and 39\%
for standalone GPT-5 web search and 32\% and 41\% for Medical Deep Research.
ICD-Deepresearch therefore improves over the registered local comparators
while retrieving evidence with higher physician-rated usefulness than the
standalone research systems~\footnote{We will release our code upon acceptance.}.
\end{abstract}

\section{Introduction}

Electronic health records (EHRs) accumulate over clinic visits and hospital
admissions. Each encounter records diagnoses, tests, procedures, and
treatments, so the resulting record describes a longitudinal trajectory rather
than a single clinical state. Hospitals represent documented diagnoses using
International Classification of Diseases (ICD) codes, standardized identifiers
for morbidity classification, reporting, and reimbursement
~\cite{cmsnchs2025icd10cm}. We study \emph{next-encounter ICD forecasting}:
given a patient's record through encounter $t$, rank the codes likely to be
documented at encounter $t{+}1$ and return the top $K$. At prediction time,
the model observes only information recorded through encounter $t$; it cannot
use anything that will arise during encounter $t{+}1$, including presenting
symptoms, examination findings, new test results, procedures, or clinician
documentation. The target is the complete ICD code set documented at that
encounter. It is prospective because the target lies beyond the observation
window, and multi-label because one encounter may receive multiple diagnosis
codes. The model must therefore project the observed patient trajectory into
its next documented state.

This prospective task benefits from two complementary foundation-model tools
~\cite{moor2023foundation}. Longitudinal EHR models learn recurrence and
progression from patient trajectories, providing a patient-specific prior
~\cite{zhang2025thcm,kraljevic2024foresight,fallahpour2025ehrmamba,
rajamohan2026raven,ghaffari2026sparseehr}; language foundation models use
broader medical knowledge to propose additional diagnoses
~\cite{benshoham2024cpllm,ma2025memorize,openai2025gpt5}. These hypotheses may
recover missed transitions but need not be grounded in the patient record.
Under a fixed top-$K$ budget, the central challenge is therefore to determine
which candidates to investigate, ground, and retain.

This control problem motivates DeepResearch: agents maintain state, plan
questions, invoke tools, and revise hypotheses from evidence
~\cite{yao2023react,jin2025searchr1,li2025webthinker,wang2026deepmed}. Here,
foundation models generate candidates, while medical search and ICD
dictionaries provide transition evidence and semantic verification
~\cite{yuan2025reliable}. Forecasting differs from answer retrieval because no
source contains the future code set. External evidence can support a
transition or verify a code, but only the pre-target record can ground it in
the patient; research also adds candidates that compete for the same $K$
positions. DeepResearch must therefore choose what to investigate, keep
patient, relation, and code evidence distinct, and jointly select candidates
under a fixed budget.

\begin{figure*}[t]
  \centering
  \includegraphics[width=0.85\textwidth]{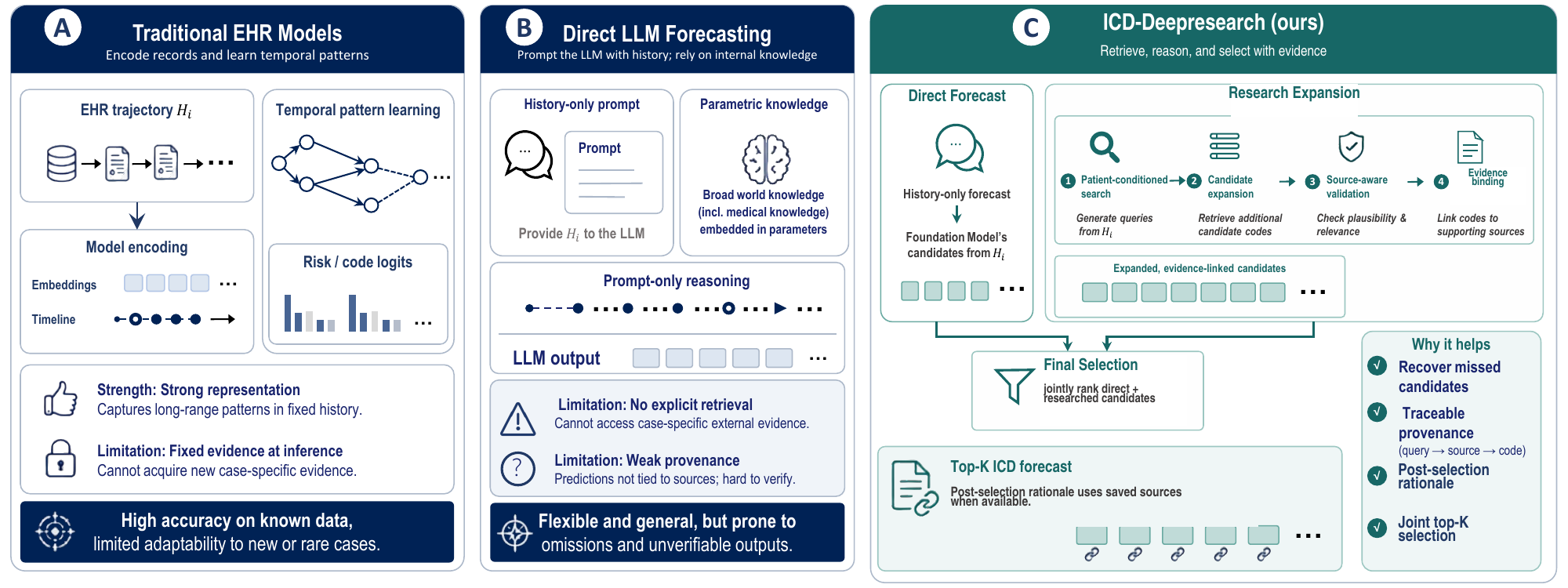}
  \caption{Foundation models as forecasting tools. ICD-Deepresearch uses
  structured-EHR and language foundation models as complementary candidate
  generators, researches patient-anchored EHR candidates, and jointly selects
  all candidates under a fixed top-$K$ budget.}
  \label{fig:comparison-main-v118}
  \vspace{-2mm}
\end{figure*}

ICD-Deepresearch is designed around these decisions
(Figure~\ref{fig:comparison-main-v118}). Candidate Generation calls SparseEHR
to produce an EHR Prior, while a parallel Direct Forecast calls GPT-5 for an
independent code list
~\cite{ghaffari2026sparseehr,openai2025gpt5}. Only the EHR Prior initializes
Research Expansion, keeping the research agenda anchored to the longitudinal
record while preserving language-model candidates as an independent path.
Research Expansion then performs two bounded rounds of query planning,
retrieval, reading, and verification to connect candidates with directional
clinical evidence and exact code semantics. Final Selection validates,
deduplicates, and jointly ranks candidates from both paths. Supporting
rationales are generated only after the predictions are fixed.

We evaluate ICD-Deepresearch on MIMIC-III and MIMIC-IV
~\cite{johnson2016mimiciii,johnson2023mimiciv}. It outperforms Candidate
Generation Only and standalone GPT-5 Web Search at both reported cutoffs, and
joint selection outperforms the isolated candidate paths and standalone
research systems, including Medical Deep Research
~\cite{clinicalcopilot2026medicaldeepresearch}. Physician audits further show
that Research Expansion retrieves fewer documents while producing a higher
proportion of useful evidence. Together, these results show that DeepResearch
benefits forecasting when it controls candidate expansion, evidence
acquisition, and fixed-budget selection rather than treating search as a way
to retrieve the future answer.

\section{Related Work}

\paragraph{Longitudinal multi-label forecasting.}
Prior models generate future visits and improve recurrence modeling,
multimodal robustness, code-aware ranking, or calibration
~\cite{yang2023transformehr,kraljevic2024foresight,
fallahpour2025ehrmamba,theodorou2025medisim,shmatko2025natural,
rajamohan2026raven,benshoham2024cpllm,koo2025overcoming,
ma2025memorize,zhang2025thcm}. These methods forecast from the observed record,
learned parameters, and static code resources. SparseEHR provides the
structured-EHR backbone inside our Candidate Generation module
~\cite{ghaffari2026sparseehr}; ICD-Deepresearch augments the module's initial candidate
state with direct generation, patient-conditioned evidence acquisition, and
joint selection.

\paragraph{Adaptive retrieval and latent-query search.}
Adaptive RAG systems learn when and how deeply to retrieve, while recent
research agents interleave query decomposition, reading, memory, and decision
updates
~\cite{asai2024selfrag,jeong2024adaptive,jin2025searchr1,
song2025r1searcher,tan2025ragr1,li2025webthinker,
zheng2025deepresearcher,shi2025deepresearch}. Clinical agents retrieve for a
specified diagnosis or decision~\cite{shi2024ehragent,rose2025meddxagent,
li2026clicare}. Medical Deep Research provides an open-source multi-agent
workflow that decomposes medical questions and coordinates specialized
databases, web search, analysis, and report generation
~\cite{clinicalcopilot2026medicaldeepresearch}. We evaluate it as a standalone
domain-specific research comparator. Next-encounter forecasting begins from a
trajectory without an explicit query; Research Expansion pairs a patient
anchor with a candidate transition, retrieves evidence for that relation, maps
it to an exact code, and updates the candidate ranking.

\paragraph{Retrospective coding and prospective explanation.}
Retrospective automatic coding assigns ICD labels after the target encounter
has been documented; its evidence therefore comes from a completed note that
already describes the relevant diagnoses
~\cite{mullenbach2018caml,baksi2025medcoder,motzfeldt2025clh,
yuan2025reliable,yin2026icdagent,zheng2026meddcr}. Next-encounter forecasting
operates under a different information boundary: the target encounter, its
documentation, and its ICD code set are all unavailable when the prediction
is made. A prospective explanation must instead connect evidence in the
pre-target record to a plausible directional clinical transition and the exact
semantics of the predicted code, without treating a code definition as
evidence that the event will occur. Our post-selection audit reflects this
distinction by separately evaluating observed patient facts, external clinical
relations, ICD semantics, forecast uncertainty, and prediction-set coverage.

\begin{figure*}[t]
  \centering
  \includegraphics[width=0.85\textwidth]{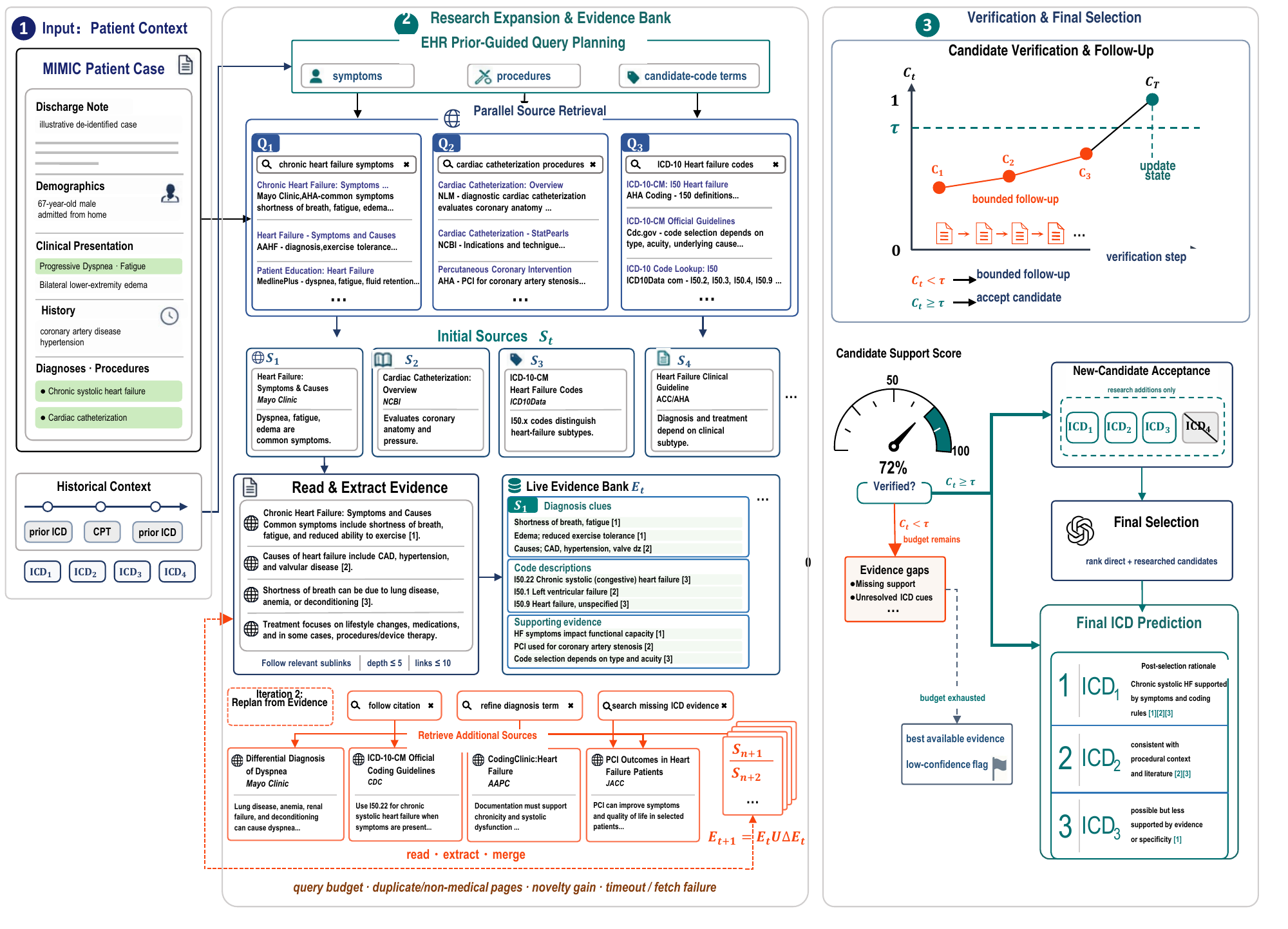}
  \caption{\textbf{ICD-Deepresearch pipeline.}
  The EHR Prior initializes two research rounds; Direct Forecast remains
  independent until Final Selection. The Evidence-Aware Reranker defines the
  full model, RRF is a same-pool selector control, and rationales are written
  only after predictions are fixed. Details appear in Supplement Sec.~S2.}
  \label{fig:framework-main-v118}
  \vspace{-2mm}
\end{figure*}
\section{Method}
\label{sec:main-method-v118}

\subsection{System Overview}
ICD-Deepresearch comprises three forecasting stages followed by a separate
explanation stage (Figure~\ref{fig:framework-main-v118}). In the first stage,
two candidate paths run in parallel. \emph{Candidate Generation} maps the
structured longitudinal record to an \emph{EHR Prior}, whereas \emph{Direct
Forecast} maps the complete pre-target record to an independent GPT-5 code
list. Only the EHR Prior initializes \emph{Research Expansion}, which performs
exactly two bounded iterations of query planning, retrieval, reading,
verification, and state update. \emph{Final Selection} then validates and
jointly ranks the union of the researched and direct candidates under one
top-$K$ budget. After predictions are fixed, \emph{Post-Selection Explanation}
writes supporting rationales without changing the selected codes or their
order.

\subsection{Task, Inputs, and Outputs}
For dataset $d$, let $\Omega_d$ denote its valid diagnosis-code vocabulary.
Patient $i$ has $m_i$ observed encounters before the held-out target encounter.
The available pre-target record is
\begin{equation}
H_i=(d_i,E_i^1,\ldots,E_i^{m_i},N_i),
\end{equation}
where $d_i$ contains demographics, $E_i^j$ is the set of diagnosis and
procedure events recorded at encounter $j$, and $N_i$ is the possibly empty
collection of notes available before the cutoff. Its structured portion is
$H_i^{\mathrm{str}}=(d_i,E_i^1,\ldots,E_i^{m_i})$. The target
$Y_i\subseteq\Omega_d$ is the deduplicated diagnosis-code set recorded at the
held-out encounter. All inference-time generation, tool use, ranking, and
explanation operate only on $H_i$.

A predictor returns a finite ordered list
$\pi_i=(\pi_{i1},\ldots)$ of unique, vocabulary-valid codes. All evaluated
systems return at least 20 such codes. For $K\in{10,20}$, the exact-$K$
prediction set is
\begin{equation}
\widehat Y_i^{(K)}
={\pi_{ij}:1\leq j\leq K}.
\label{eq:main-topk-v118}
\end{equation}
A single ranking therefore supports evaluation at both reported cutoffs.

\subsection{Two Candidate Paths}
\paragraph{Candidate Generation: the EHR Prior.}
Let $\mathcal G_{\mathrm F}$ denote the complete Candidate Generation mapping.
It converts the structured record into a finite, target-blind seed ranking
$\pi_i^{\mathrm F}$ and its unordered candidate set $B_i$:
\begin{equation}
\pi_i^{\mathrm F}
=\mathcal G_{\mathrm F}(H_i^{\mathrm{str}};\Omega_d),
\qquad
B_i=\operatorname{set}(\pi_i^{\mathrm F})\subseteq\Omega_d.
\label{eq:main-fcg}
\end{equation}
For ICD-9-CM and ICD-10-CM forecasting, SparseEHR provides the learned
structured-EHR backbone~\cite{ghaffari2026sparseehr}. The complete mapping also
includes task-specific input construction, vocabulary interfacing,
validation, and the registered output ranking. The \emph{Candidate Generation
Only} control evaluates this entire mapping, including the SparseEHR backbone.
The identities and native ranks in $\pi_i^{\mathrm F}$ are fixed before
Research Expansion begins.

\paragraph{Direct Forecast: the independent path.}
Let $\mathcal G_{\mathrm D}$ denote the GPT-5 Direct Forecast mapping. It uses
the complete pre-target record to return a finite ordered list $D_i$:
\begin{equation}
D_i=\mathcal G_{\mathrm D}(H_i;\Omega_d).
\end{equation}
Thus, $\pi_i^{\mathrm F}$ and $D_i$ are ordered lists, whereas $B_i$ is a set.
Direct Forecast can use both structured events and pre-target notes, but its
state remains outside the research loop until Final Selection. It can therefore
contribute hypotheses absent from the EHR Prior without determining which
candidates Research Expansion investigates.

\subsection{Research Expansion}
Research Expansion begins with the EHR Prior set $B_i$. For each hypothesis,
it keeps three information roles distinct: an observed patient clue, external
evidence about a possible clinical transition, and the exact meaning of the
proposed ICD code. One iteration applies
\begin{equation*}
\mathsf{Plan}\rightarrow\mathsf{ToolAct}\rightarrow\mathsf{ReadBind}
\rightarrow\mathsf{Verify}\rightarrow\mathsf{UpdateRank}.
\end{equation*}
\textsc{ToolAct} invokes Google or the fixed PMC/arXiv corpora; the remaining
operations construct and update the research state. Let
$\mathcal Z_i^{(r)}$ denote the candidate hypotheses, candidate-bound evidence
memory, support and verification fields, and unresolved directions after
research iteration $r$. The initial state is
$\mathcal Z_i^{(0)}=(B_i,\varnothing)$, where $\varnothing$ denotes empty
evidence, verification, and unresolved-direction fields.

Let $\mathcal{DR}$ denote one fixed, registered Research Expansion state
transition and let $\mathcal T_d$ denote the available tool collection. The
executor applies the transition exactly twice:
\begin{equation}
\mathcal Z_i^{(r)}
=\mathcal{DR}!\left(
H_i,B_i,\mathcal Z_i^{(r-1)};\mathcal T_d
\right),
\qquad r\in{1,2},
\label{eq:main-dr-loop}
\end{equation}
where $\mathcal T_d$ contains the online Google channel, fixed PMC/arXiv
corpora, and the ICD dictionary for dataset $d$. In each iteration, the planner
forms patient-anchored candidate--query pairs; the tools retrieve medical
sources and code definitions; the reader binds passages to candidates;
verification evaluates the patient anchor, proposed clinical relation, and
exact-code mapping; and the rank update carries the resulting state forward.
Acceptance checks apply only to newly proposed research additions. When budget
remains, unresolved additions can trigger bounded follow-up, and the second
iteration conditions on the complete state produced by the first.

Finalizing $\mathcal Z_i^{(2)}$ yields a research candidate set
$\mathcal C_i^R$, an ordered research list $R_i$, and an accumulated
candidate-bound ledger $L_i^{(2)}$. Here, $L_i^{(2)}$ is the evidence-memory
component of the final research state. It stores zero or more retrieval records
for each researched candidate, including explicit no-result and unreadable
states; retained passages remain linked to their queries and candidates.
Supplement Secs.~S2 and S5--S6 specify the executable calls, tool contracts,
and failure states.

\subsection{Joint Final Selection}
Final Selection combines the direct and research paths. It constructs the
vocabulary-valid union candidate set
\begin{equation}
\mathcal C_i^U
=\mathcal V_d!\left(
\mathcal C_i^R\cup\operatorname{set}(D_i)
\right),
\label{eq:main-union-v118}
\end{equation}
where $\mathcal V_d$ deterministically canonicalizes code identities, removes
codes outside $\Omega_d$, and merges duplicate identities. Superscripts $R$
and $U$ denote the research and union candidate sets, respectively.

\paragraph{Evidence-Aware Reranker.}
The full selector scores each candidate $c\in\mathcal C_i^U$ using the
pre-target record, generator ranks, occurrence metadata, and candidate-bound
evidence, and sorts the scores to obtain $\pi_i^{\mathrm{full}}$. The RRF
control below uses the same frozen candidate pool and differs only in the
selection rule.

\paragraph{RRF selector control.}
The selector control replaces only the Evidence-Aware Reranker with reciprocal
rank fusion. Let $P_i^{D}=D_i$ and $P_i^{R}=R_i$ denote the direct and research
rankings after stable vocabulary validation and deduplication. For
$p\in{D,R}$, let $r_i^p(c)$ be the one-based rank of canonical code $c$ when
$c\in P_i^p$. For every $c\in\mathcal C_i^U$, RRF assigns
\begin{equation}
s_i^{\mathrm{RRF}}(c)=
\sum_{\substack{p\in{D,R},c\in P_i^p}}
\frac{1}{k_0+r_i^p(c)}
\label{eq:main-rrf-v118}
\end{equation}
An empty sum is zero, so a path contributes nothing when it does not contain
$c$. Sorting $\mathcal C_i^U$ by decreasing $s_i^{\mathrm{RRF}}(c)$, with
canonical code identity as the deterministic tie-breaker, yields
$\pi_i^{\mathrm{RRF}}$. Thus, $\pi_i^{\mathrm{full}}$ and
$\pi_i^{\mathrm{RRF}}$ are ordered lists over the same frozen candidate state
and each can instantiate the generic ranking $\pi_i$ in
Equation~\ref{eq:main-topk-v118}. Their comparison changes only the selection
rule while keeping candidate availability fixed. Full pool construction and
selector contracts appear in Supplement Sec.~S2.4.

\subsection{Post-Selection Explanation}
The explanation writer runs after the selected top-$K$ codes and their order
have been fixed. For each code, it receives the pre-target record, canonical
code information, and any saved candidate-bound evidence, and returns a
rationale and an uncertainty statement. Each rationale distinguishes the
observed patient clue, external clinical relation, exact ICD meaning, and
forecast uncertainty. The reported explanation audit uses the registered
RRF-selected ranking $\pi_i^{\mathrm{RRF}}$; Supplement Sec.~S2.5 provides the
complete writer contract.

\section{Experimental Design}
\label{sec:main-experiments-v118}

\begin{table*}[t]
\centering
\footnotesize
\setlength{\tabcolsep}{3.5pt}
\renewcommand{\arraystretch}{1.03}

\begin{tabular*}{\textwidth}{@{\extracolsep{\fill}}l*{8}{r}@{}}
\toprule
& \multicolumn{4}{c}{MIMIC-III}
& \multicolumn{4}{c}{MIMIC-IV} \\
\cmidrule(lr){2-5}
\cmidrule(lr){6-9}

\textbf{Method}
& \textbf{P@10}
& \textbf{R@10}
& \textbf{P@20}
& \textbf{R@20}
& \textbf{P@10}
& \textbf{R@10}
& \textbf{P@20}
& \textbf{R@20} \\
\midrule

\multicolumn{9}{l}{\textit{Published reference results}} \\

CAML
& 20.61 & 18.15 & 14.98 & 25.74
& 20.61 & 27.38 & 14.03 & 35.52 \\

ZAGCNN
& 17.50 & 15.51 & 12.61 & 21.63
& 21.10 & 28.13 & 14.33 & 36.17 \\

GatorTron
& 20.53 & 17.72 & 15.03 & 25.66
& 22.87 & 30.19 & 15.47 & 38.81 \\

DistilBioBERT
& 20.18 & 17.85 & 14.98 & 25.62
& 22.59 & 29.83 & 15.33 & 38.50 \\

Chet
& 24.52 & 18.82 & 18.61 & 27.71
& 19.06 & 26.65 & 12.62 & 34.11 \\

DKEC
& 20.64 & 17.92 & 14.55 & 24.47
& 20.03 & 26.50 & 13.61 & 34.59 \\

BioMedLM
& 20.56 & 17.84 & 14.90 & 25.38
& 20.72 & 27.68 & 14.09 & 35.32 \\

\textbf{THCM-CAL}
& \textbf{30.02}
& \textbf{24.04}
& \textbf{21.47}
& \textbf{33.16}
& \textbf{28.83}
& \textbf{37.03}
& \textbf{18.66}
& \textbf{46.04} \\

\midrule
\multicolumn{9}{l}{\textit{Local registered scale-cohort results}} \\

ICD-Deepresearch: Candidate Generation Only
& 31.68 & 22.59 & 20.92 & 29.84
& 29.21 & 20.78 & 23.35 & 33.22 \\

GPT-5 + Web Search (Standalone)
& 29.86 & 21.29 & 22.77 & 32.48
& 28.53 & 20.29 & 22.18 & 31.56 \\

\textbf{ICD-Deepresearch}
& \textbf{35.33}
& \textbf{25.20}
& \textbf{24.60}
& \textbf{35.09}
& \textbf{33.46}
& \textbf{39.41}
& \textbf{25.14}
& \textbf{48.32} \\

\bottomrule
\end{tabular*}
\caption{Exact-$K$ forecasting results. P@$K$/R@$K$ are patient-averaged.
Published THCM-CAL-protocol values are transcribed from its
Table~2~\cite{zhang2025thcm} and form a separate block. Boldface indicates the
best result within each block.}
\label{tab:main-results}
\vspace{-3mm}
\end{table*}

\paragraph{Evaluation logic.}
We evaluate forecasting, component and selector controls, retrieval quality
and efficiency, and post-selection explanations (\textbf{RQ1--RQ4}).
MIMIC-III/IV use held-out final admissions with ICD-9-CM/ICD-10-CM targets
~\cite{johnson2016mimiciii,johnson2023mimiciv}.
Table~\ref{tab:main-results} reports scale cohorts, whereas
Table~\ref{tab:functional-ablation-main} uses separate subset diagnostic cohorts;
comparisons remain within each table and dataset. Supplement
Secs.~S3.1--S3.2 provide the analysis registry, cohort sizes and splits, and
outcome construction.

\paragraph{Comparators and controls.}
The published block imports THCM-CAL Table~2 values
~\cite{zhang2025thcm}; original baseline sources and numerical provenance are
listed in Supplement Sec.~S3.3. Local experiments compare Candidate Generation
Only and the full system with standalone GPT-5 Web Search; component controls
isolate the Direct and Research paths, and vary research context under a fixed candidate pool. Medical
Deep Research~\cite{clinicalcopilot2026medicaldeepresearch} is a second
standalone comparator and receives neither our EHR Prior, research ledger, nor
selector. Supplement Secs.~S3.3--S3.4 specify prompts, retrieval budgets,
model snapshots, seeds, implementation details, and controlled states.

\paragraph{Metrics and audits.}
Table~\ref{tab:main-results} uses patient-averaged P@$K$/R@$K$;
Table~\ref{tab:functional-ablation-main} uses pooled micro-P/R/F1 and
patient-macro F1. Retrieval audits report source composition, metadata
support/forecast usefulness, and blinded physician usefulness; explanations
use five reason dimensions and coverage-adjusted Reason5. Supplement
Sec.~S3.5 defines metrics and inference, while Secs.~S3.6--S3.8 specify source
sampling, efficiency accounting, and explanation judge contracts.
\begin{table*}[t]
\centering

\footnotesize
\setlength{\tabcolsep}{3.0pt}
\renewcommand{\arraystretch}{1.02}

\begin{tabularx}{\textwidth}{
  @{}
  >{\raggedright\arraybackslash}p{0.087\textwidth}
  >{\raggedright\arraybackslash}X
  *{4}{>{\raggedleft\arraybackslash}p{0.077\textwidth}}
  @{}
}
\toprule
\bfseries Dataset
& \bfseries System variant
& \bfseries Micro-P
& \bfseries Micro-R
& \bfseries Micro-F1
& \bfseries Macro-F1 \\
\midrule

\multicolumn{6}{@{}l}{
  \emph{Panel A: individual candidate paths and joint selection}
} \\

\textbf{MIMIC-III}
& ICD-Deepresearch: Candidate Generation Only
& 21.54 & 30.45 & 25.23 & 24.12 \\

& ICD-Deepresearch: Direct Forecast Only
& 21.80 & 30.90 & 25.56 & 24.80 \\

& ICD-Deepresearch: Research Path Only
& 20.71 & 29.29 & 24.27 & 23.53 \\

& ICD-Deepresearch w/o Evidence-Aware Reranker
& 22.81 & 32.26 & 26.73 & 25.88 \\

& \textbf{ICD-Deepresearch}
& \textbf{24.97}
& \textbf{35.30}
& \textbf{29.25}
& \textbf{28.21} \\

\addlinespace[2pt]

\textbf{MIMIC-IV}
& ICD-Deepresearch: Candidate Generation Only
& 14.40 & 20.69 & 16.98 & 16.88 \\

& ICD-Deepresearch: Direct Forecast Only
& 26.60 & 38.20 & 31.36 & 30.60 \\

& ICD-Deepresearch: Research Path Only
& 27.90 & 40.09 & 32.90 & 32.04 \\

& ICD-Deepresearch w/o Evidence-Aware Reranker
& 29.55 & 42.46 & 34.85 & 33.98 \\

& \textbf{ICD-Deepresearch}
& \textbf{30.05}
& \textbf{43.18}
& \textbf{35.44}
& \textbf{34.54} \\

\midrule

\multicolumn{6}{@{}l}{
  \emph{Panel B: standalone research comparators}
} \\

\textbf{MIMIC-III}
& GPT-5 + Web Search (Standalone)
& 22.91 & 32.40 & 26.84 & 25.94 \\

& Medical Deep Research (Standalone)
& 17.60 & 26.29 & 21.08 & 20.53 \\

\textbf{MIMIC-IV}
& GPT-5 + Web Search (Standalone)
& 28.30 & 40.66 & 33.37 & 32.63 \\

& Medical Deep Research (Standalone)
& 13.40 & 19.06 & 15.74 & 15.56 \\

\midrule

\multicolumn{6}{@{}l}{
  \emph{Panel C: fixed-pool research-context control}
} \\

\textbf{MIMIC-III}
& ICD-Deepresearch with Research Context (Fixed Pool)
& 22.30 & 35.12 & 27.28 & 26.39 \\

& ICD-Deepresearch without Research Context (Fixed Pool)
& 22.20 & 34.96 & 27.16 & 26.48 \\

\addlinespace[2pt]

\textbf{MIMIC-IV}
& ICD-Deepresearch with Research Context (Fixed Pool)
& 25.20 & 36.95 & 29.96 & 29.50 \\

& ICD-Deepresearch without Research Context (Fixed Pool)
& 23.60 & 34.60 & 28.06 & 27.79 \\

\bottomrule
\end{tabularx}
\caption{Component analysis at $K=20$. Panels compare isolated paths and joint
selectors (A), standalone research comparators (B), and research context under
a fixed candidate pool (C). Values are percentages.}
\label{tab:functional-ablation-main}
\vspace{-2mm}
\end{table*}
\begin{table*}[t]
\centering
\small
\setlength{\tabcolsep}{1.0pt}
\renewcommand{\arraystretch}{1.02}
\begin{tabularx}{\textwidth}{
  @{}
  >{\raggedright\arraybackslash}p{0.095\textwidth}
  >{\raggedright\arraybackslash}X
  rr
  @{\hspace{3pt}}
  *{7}{r}
  @{\hspace{3pt}}
  rr
  @{\hspace{3pt}}
  r
  @{}
}
\toprule
& & \multicolumn{2}{c}{Retrieval volume}
& \multicolumn{7}{c}{Source composition (\%)}
& \multicolumn{2}{c}{Metadata proxy (\%)}
& \multicolumn{1}{c}{Blinded audit (\%)} \\
\cmidrule(lr){3-4}
\cmidrule(lr){5-11}
\cmidrule(lr){12-13}
\cmidrule(l){14-14}
\bfseries Dataset
& \bfseries System
& \bfseries Uses
& \bfseries Uniq.
& \bfseries Peer
& \bfseries Offic.
& \bfseries Clin.
& \bfseries Coding
& \bfseries Repo.
& \bfseries Forum
& \bfseries Other
& \bfseries Supp.
& \bfseries Fcast.
& \bfseries Phys. \\
\midrule
\textbf{MIMIC-III}
& GPT-5 + Web Search
& 13,394 & 6,363
& 7.3 & 19.0 & 1.1 & 30.7 & 6.6 & 16.2 & 19.1
& 30.0 & 6.0 & 22.0 \\
& \scalebox{0.91}[1]{Medical Deep Research}
& 37,724 & 26,268
& 44.2 & 12.7 & 2.8 & 25.7 & 0.5 & 0.0 & 14.1
& 58.0 & 34.0 & 32.0 \\
& \textbf{Research Expansion}
& 1,410 & 1,246
& 34.2 & 6.8 & 7.2 & 29.3 & 0.1 & 0.0 & 22.4
& \textbf{62.0} & \textbf{22.0} & \textbf{51.0} \\
\addlinespace[3pt]
\textbf{MIMIC-IV}
& GPT-5 + Web Search
& 25,100 & 12,589
& 1.1 & 16.3 & 0.5 & 21.2 & 5.0 & 21.3 & 34.6
& 34.0 & 4.0 & 39.0 \\
& \scalebox{0.91}[1]{Medical Deep Research}
& 37,989 & 28,072
& 45.2 & 19.2 & 3.3 & 7.5 & 0.4 & 0.0 & 24.3
& 50.0 & 34.0 & 41.0 \\
& \textbf{Research Expansion}
& 11,124 & 9,178
& 50.9 & 4.5 & 4.6 & 17.7 & 0.1 & 0.0 & 22.1
& \textbf{64.0} & \textbf{36.0} & \textbf{68.0} \\
\bottomrule
\end{tabularx}
\caption{\textbf{Retrieved-source composition and usefulness audits.}
Uses/Uniq. are counts after excluding deterministic dictionary citations; all
other entries are percentages. Abbreviations and sampling are defined in
Supplement Sec.~S3.6. Bold identifies Research Expansion, not a columnwise
maximum.}
\label{tab:source-audit-main-v119}
\vspace{-4mm}
\end{table*}

\section{Results and Discussion}
\label{sec:main-results-v118}

\subsection{RQ1: Complementary Candidate States Require Joint Selection}
On the registered scale cohorts, the full ICD-Deepresearch system reaches
patient-averaged P@20/R@20 of 24.60/35.09\% on MIMIC-III and
25.14/48.32\% on MIMIC-IV. At $K=20$, it exceeds GPT-5 + Web
Search (Standalone)
by $+1.83/+2.61$ precision/recall points on MIMIC-III and $+2.96/+16.76$ on
MIMIC-IV; relative to the Candidate Generation Only control, the gains are
$+3.68/+5.25$ and $+1.79/+15.10$ points. ICD-Deepresearch also leads both
local comparators at $K=10$ (Table~\ref{tab:main-results}).

The functional audit uses separate registered diagnostic cohorts and pooled
metrics; its absolute values are therefore interpreted within Table~\ref{tab:functional-ablation-main},
not against the scale-cohort values in Table~\ref{tab:main-results}. On
MIMIC-III, the best isolated path
reaches 25.56 micro-F1, compared with 26.73 for ICD-Deepresearch w/o
Evidence-Aware Reranker and 29.25 for ICD-Deepresearch. On MIMIC-IV, the
Research Path Only control reaches 32.90 and ICD-Deepresearch w/o Evidence-Aware Reranker
reaches 34.85, while the full system reaches 35.44
(Table~\ref{tab:functional-ablation-main}). The Evidence-Aware Reranker
therefore adds 2.52 points over RRF on MIMIC-III and 0.59 on MIMIC-IV.
Among standalone research systems, GPT-5 + Web Search reaches 26.84/33.37
micro-F1 and Medical Deep Research reaches 21.08/15.74 on
MIMIC-III/MIMIC-IV. The joint system exceeds every isolated path and
standalone comparator in both diagnostic cohorts.

The row-62 trace shows ICD-Deepresearch w/o Evidence-Aware Reranker preserving
branch-exclusive hits and promoting a candidate supported at moderate ranks by
both branches; it returns eight true positives, compared with five for GPT-5
(Direct Forecasting) and six for the Research Path Only control (Supplement
Sec.~S8). The MIMIC-III joint-pool oracle reaches 37.53 micro-F1, 8.28 points
above ICD-Deepresearch, locating substantial remaining headroom in slot
allocation over the observed pool (Supplement Sec.~S4.1).

\subsection{RQ2: Search Depth Is a State-Dependent Allocation Problem}
The fixed-pool control measures context-assisted ranking after candidate
generation. With candidate identities, order, GPT-5 backend, and $K=20$ fixed,
the Research Expansion packet changes micro-F1 by $+0.12$ points on MIMIC-III (95\% CI
$-1.12$ to $+1.33$) and $+1.90$ on MIMIC-IV. The fixed-pool oracles remain
5.75 and 5.59 points higher. This contrast estimates the selector's use of the
delivered research packet conditional on an already research-expanded pool;
the natural-cardinality analysis below separately characterizes candidate
coverage and growth (Supplement Sec.~S4.6).

Natural-cardinality diagnostics show why coverage must be evaluated together
with candidate growth. On EHRSHOT, two iterations recover 144 additional
targets while increasing mean pool size from 19.8 to 51.0 and reducing
micro-F1 from 25.82\% to 16.34\%. Initial expansion changes micro-F1 by
$+3.06$ and $+9.04$ points on the MIMIC-III and MIMIC-IV diagnostics, whereas
follow-up changes it by $-0.98$ and $-0.004$. The MIMIC-IV follow-up yield is
12.96\%, nearly equal to its 13.01\% break-even requirement
(Figure~\ref{fig:candidate-expansion-main}; Supplement Sec.~S4.6).

The near-zero cohort mean hides concentrated benefit: 26 of 28 MIMIC-IV
follow-up matches occur in 13 patients, while 52 patients worsen. Rows 52 and
59 obtain four of eight and zero of nine follow-up matches, respectively. A
post-hoc support-score filter retains 21 of the 28 matches, reduces mean pool
size from 38.19 to 36.64, and raises micro-F1 from 26.02\% to 26.54\%
(Supplement Secs.~S4.1 and S8). These diagnostics motivate learning marginal
expansion value from the current state and stopping when expected yield falls
below the cardinality cost.

\begin{figure*}[!t]
  \centering
  \includegraphics[width=0.9\textwidth]{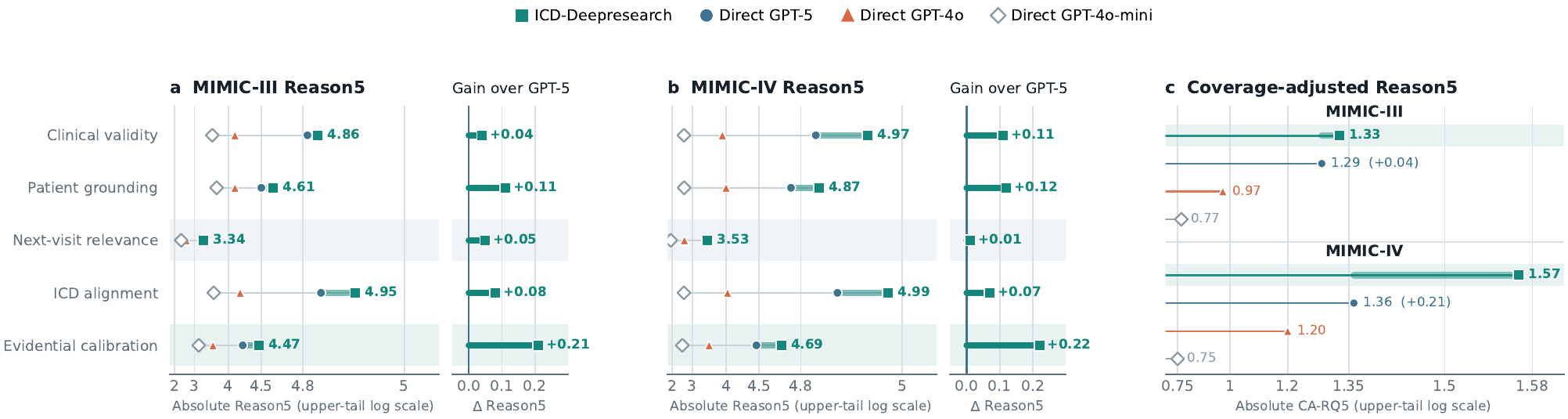}
  \caption{\textbf{Post-selection explanation quality.}
  Panels (a--b) show Reason5 on each system's true-positive predictions;
  panel (c) combines reason quality with coverage at $K=20$. The
  ICD-Deepresearch arm is RRF-selected. Exact values and the judge contract
  appear in Supplement Secs.~S6.1 and S3.8.}
  \label{fig:combined-explanation-audit}
  \vspace{-2mm}
\end{figure*}
\begin{figure*}[t]
  \centering
  \includegraphics[width=0.9\textwidth]{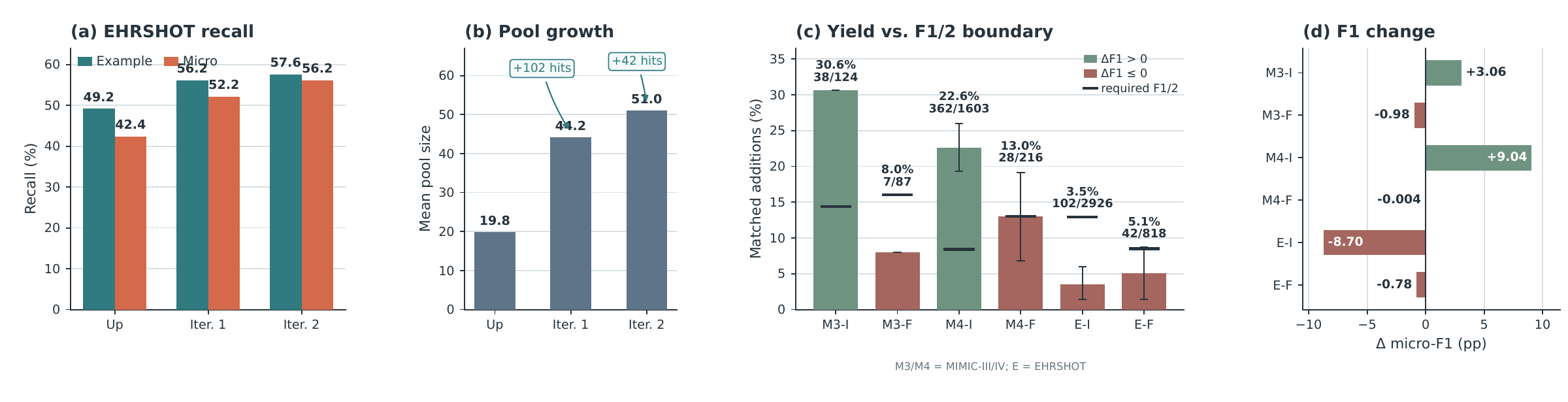}
  \caption{\textbf{Candidate expansion before Final Selection.}
  Panels show target recall, pool growth, match yield against the exact
  micro-F1/2 break-even threshold, and the resulting $\Delta$micro-F1.}
  \label{fig:candidate-expansion-main}
  \vspace{-2mm}
\end{figure*}

\subsection{RQ3: Focused Retrieval Improves Physician-Rated Usefulness}
Research Expansion retrieves substantially fewer source occurrences than
either standalone system: 1,410 on MIMIC-III and 11,124 on MIMIC-IV, compared
with 13,394/25,100 for GPT-5 + Web Search and 37,724/37,989 for Medical Deep
Research (Table~\ref{tab:source-audit-main-v119}). Despite this smaller
retrieval set, it attains the highest physician-rated usefulness on both
datasets: 51\% versus 22\% and 32\% on MIMIC-III, and 68\% versus 39\% and
41\% on MIMIC-IV.

The automated proxies reveal a different ordering. Medical Deep Research has
the highest peer-reviewed share and forecast-usefulness rate on MIMIC-III
(44.2\% and 34\%). Research Expansion has the highest peer-reviewed share on
MIMIC-IV (50.9\%), support rate on both datasets (62\% and 64\%), and
forecast-usefulness rate on MIMIC-IV (36\%). Source category and metadata
relevance therefore do not substitute for clinician judgment of case-specific
usefulness; the patient-conditioned research state is most clearly associated
with selective retrieval and higher physician-rated utility.

On matched cohorts, Research Expansion uses 7.3$\times$/3.4$\times$ fewer unique queries,
3.8$\times$/4.5$\times$ fewer source accesses, and
6.1$\times$/2.9$\times$ fewer visible tokens; recorded batched throughput is
16.8$\times$/21.9$\times$ higher, with F1@20 changing from 26.4 to 30.2 and
from 33.3 to 34.8 (Supplement Sec.~S4.4). Paired traces show the associated
query shift: Research Expansion searches patient-linked transitions such as
glycemic control to ulcer healing, whereas the paired standalone-search traces
emphasize code lists and return forum pages (Supplement Sec.~S8). The
paired traces illustrate how the candidate--anchor--relation state can direct
queries toward patient-linked transitions rather than generic code lists.

\subsection{RQ4: Explanation State Improves Semantics More Than Temporal Grounding}
Shared-hit comparisons evaluate identical patient--code predictions. Across
441 MIMIC-III shared hits, ICD-Deepresearch (RRF-selected) improves exact ICD
alignment by 0.07 points
(95\% CI $+0.02$ to $+0.12$) and evidential calibration by 0.22
($+0.15$ to $+0.29$). Across 479 MIMIC-IV shared hits, clinical validity,
patient grounding, ICD alignment, and calibration improve, while
next-encounter relevance changes by only $+0.03$ ($-0.04$ to $+0.08$).
Coverage-adjusted quality rises from 1.36 to 1.57 on MIMIC-IV (95\% CI
$+0.15$ to $+0.27$), while MIMIC-III changes from 1.29 to 1.33
($-0.02$ to $+0.09$). The clearest gains are in code semantics and
calibration; the temporal bridge to the next encounter remains weak
(Figure~\ref{fig:combined-explanation-audit}; Supplement Secs.~S4.5 and S6.1).

The row-45 paired case isolates this distinction. GPT-5 (Direct Forecasting)
and ICD-Deepresearch (RRF-selected) recover the same three codes at $K=20$, but
their shared-hit Reason5 means are 3.47 and 4.80. GPT-5 describes O34.13 only
as a uterine-scar code, whereas ICD-Deepresearch uses its canonical title to
align it with the observed leiomyoma trajectory. All 20 citations are ICD
definitions, making code identity inspectable while leaving the
immediate-horizon relation grounded in the patient record (Supplement
Sec.~S8).

This separation also clarifies the remaining failures. Next-encounter relevance
is the lowest Reason5 dimension, while independent relation sources cover only
11.62
patient anchors, query--candidate drift, definition--forecast conflation,
semantic mismatches, horizon-polarity errors, and uncontrolled cardinality.
They motivate a three-link verification target: an immutable pre-target
patient span, a source-entailed directional relation, and an exact code
expressing the same entity at the immediate forecasting horizon (Supplement
Secs.~S8--S9).




\section{Conclusion}
Across the four RQs, next-encounter ICD forecasting emerges as budgeted
hypothesis search: candidate expansion broadens coverage but also enlarges the
pool that Final Selection must resolve. The MIMIC-III target-conditioned
oracle leaves 8.28 micro-F1 points of joint-pool selection headroom; attaching
research context changes micro-F1 by $+0.12$ and $+1.90$ on MIMIC-III and
MIMIC-IV, respectively; and independent relation-source coverage is only
11.62\% and 1.39\%. Candidate-conditioned retrieval yields denser
physician-rated evidence, while the richer post-selection state is associated
more strongly with exact-code alignment and calibration than with
immediate-horizon relevance. These results motivate a state-aware controller
that expands only when the expected marginal value justifies a candidate with
a patient anchor, directional clinical relation, and verified code semantics;
otherwise, it retains the current pool or stops.

\bibliography{references}
\onecolumn
\raggedbottom
\sloppy
\appendix
\setcounter{section}{0}
\renewcommand{\thesection}{S\arabic{section}}
\renewcommand{\thesubsection}{\thesection.\arabic{subsection}}
\setcounter{table}{0}
\renewcommand{\thetable}{S\arabic{table}}
\setcounter{figure}{0}
\renewcommand{\thefigure}{S\arabic{figure}}
\makeatletter
\long\def\@makecaption#1#2{%
  \vskip\abovecaptionskip
  \noindent\small\textbf{#1.} #2\par
  \vskip\belowcaptionskip}
\makeatother

\section{Guide to the Supplementary Materials}\label{sec:supp-guide}

The main paper is the authoritative source for the reader-facing method
definition, component names, information boundary, evaluation contracts, and
reported claims. This supplement follows those definitions and provides
implementation detail, diagnostics, prompt contracts, case traces, and
artifact provenance. Throughout, \emph{ICD-Deepresearch} without a qualifier
denotes the full system with the Evidence-Aware Reranker.
\emph{ICD-Deepresearch w/o Evidence-Aware Reranker} denotes the controlled
RRF selector replacement over the same available candidate state. The suffix
\emph{Only} denotes an isolated candidate path, whereas \emph{Standalone}
denotes an external system that does not share the ICD-Deepresearch pipeline.
The post-selection explanation audit is explicitly labeled
\emph{ICD-Deepresearch (RRF-selected)} and is not an audit of explanations
from the full Evidence-Aware output.

Section~S2 specifies the EHR Prior, Direct Forecast, Research Expansion, Final
Selection, and the explanation writer.
Section~S3 describes cohorts, comparisons, metrics, source and efficiency
accounting, uncertainty, and audit contracts. Section~S4 reports method-aligned
functional results, retrieval efficiency, and mechanism diagnostics.
Section~S5 gives the extended interpretation, limitations, and ethics details.
Section~S6 provides exact figure values and executable-call provenance, and
Section~S7 records the prompt contracts. Sections~S8--S9 contain case traces
and the cross-case error taxonomy.

Quoted system messages and natural-language instructions are transcribed from
the executable sources identified in Table~\ref{tab:s-agent-inventory}. Long
JSON contracts are shown verbatim or as complete required-field lists; dynamic
fields use angle-bracket tokens, and runtime truncation or retry suffixes are
recorded where relevant.

\begin{center}
\footnotesize
\setlength{\tabcolsep}{6pt}
\renewcommand{\arraystretch}{1.12}
\begin{tabularx}{\textwidth}{>{\raggedright\arraybackslash}p{3.1cm}Y}
\toprule
\rowcolor{SuppBlue}
\multicolumn{2}{c}{\textbf{\textcolor{SuppBlueDark}{Supplementary Map}}}\\
\midrule
\cellcolor{SuppGray}\textbf{Method and protocol (S2--S3)}
& Candidate Generation (SparseEHR backbone), Direct Forecast, Research
Expansion, the Evidence-Aware final selector and its RRF replacement,
prediction-locked explanation writing, cohorts,
metrics, source accounting, and uncertainty.\\
\cellcolor{SuppOrange}\textbf{Functional results (S4)}
& Component and selector controls; GPT-5 + Web Search (Standalone) and Medical
Deep Research (Standalone), fixed-pool
research-context input, iteration, transfer, source, efficiency, explanation, and safety
analyses.\\
\cellcolor{SuppGreen}\textbf{Interpretation and records (S5--S7)}
& Extended interpretation and limitations, exact figure values,
executable-call provenance, runtime boundaries, and prompt contracts.\\
\cellcolor{SuppPurple}\textbf{Cases and error analysis (S8--S9)}
& Success and failure traces, paired explanations, retrieval records, and
cross-case mapping, entailment, horizon, and patient-anchor errors.\\
\bottomrule
\end{tabularx}
\end{center}

\section{Full Method Specification}\label{sec:supp-full-method}

\subsection{Task and Information Boundary}
\label{sec:supp-method-boundary}

For dataset $d$, let $\Omega_d$ denote its valid diagnosis-code vocabulary.
Patient $i$ has $m_i$ observed encounters before the held-out target
encounter. Inference stops immediately before that encounter, and the
available pre-target record is
\begin{equation}
H_i=(d_i,E_i^1,\ldots,E_i^{m_i},N_i),
\end{equation}
where $d_i$ contains demographics, $E_i^j$ contains diagnosis and procedure
events recorded at observed encounter $j$, and $N_i$ is the possibly empty
collection of notes available before the cutoff. Any discharge note in $N_i$
belongs to an observed encounter, never to the held-out target encounter. The
structured portion of the record is
$H_i^{\mathrm{str}}=(d_i,E_i^1,\ldots,E_i^{m_i})$. The target
$Y_i\subseteq\Omega_d$ is the deduplicated diagnosis-code set recorded at the
held-out encounter.

All inference-time generation, retrieval, reading, ranking, and explanation
use only $H_i$ and resources fixed independently of the evaluation targets.
No target code, target-encounter note, or statistic computed from another
evaluation patient is available to any forecasting component. The target
$Y_i$ is accessed only after predictions are fixed, for metric computation
and by the explicitly gold-aware explanation-evaluation judge. Search phrases
remove patient identifiers present in the de-identified input.

The executable calls use registered deterministic serializations of $H_i$.
For call $a$, let $\nu_a(H_i)$ denote the fields and character-limited view
specified in Section~S7. Each $\nu_a$ is a function only of the same
pre-target record and therefore does not enlarge the information boundary.
To avoid clutter, the component equations below write $H_i$ when the
call-specific view is clear from context.

Candidate Generation maps $H_i^{\mathrm{str}}$ to the registered, target-blind
20-code EHR Prior
\begin{equation}
\pi_i^{\mathrm F}
=\mathcal G_{\mathrm F}(H_i^{\mathrm{str}};\Omega_d),
\qquad
B_i=\operatorname{set}(\pi_i^{\mathrm F})\subseteq\Omega_d.
\end{equation}
Here, Candidate Generation is the complete module and
$\pi_i^{\mathrm F}$ is its ordered EHR Prior output; $B_i$ is the corresponding
unordered seed set. Only $B_i$ initializes Research Expansion.

A predictor returns a finite ordered list
$\pi_i=(\pi_{i1},\ldots)$ of unique, vocabulary-valid codes. Every evaluated
system returns at least 20 codes. For $K\in\{10,20\}$, its exact-$K$
prediction set is
\begin{equation}
\widehat Y_i^{(K)}
=\{\pi_{ij}:1\leq j\leq K\}.
\label{eq:topk-output}
\end{equation}
Thus, the same complete ranking supports both reported cutoffs, and no
variable-length denominator is used in the primary exact-$K$ evaluation.
Throughout, $\pi_i^{\mathrm F}$, $D_i$, $\bar R_i$, $R_i$, and $\pi_i$ are
ordered lists; $B_i$, $Y_i$, and calligraphic $\mathcal C$ symbols are sets.
Superscripts $(1)$ and $(2)$ identify the two bounded Research Expansion
iterations, and superscripts $\mathrm{RRF}$ and $\mathrm{full}$ identify the
two Final Selection protocols.

ICD-Deepresearch comprises three forecasting stages followed by a separate
explanation stage. In the first stage, Candidate Generation produces the EHR
Prior $(\pi_i^{\mathrm F},B_i)$, while Direct Forecast independently produces
$D_i$ from the complete pre-target record. Direct Forecast remains outside the
research loop. In the second stage, Research Expansion is initialized only by
$B_i$. In the third stage, Final Selection validates and jointly ranks the
union of researched and direct candidates. Post-Selection Explanation runs
only after the selected codes and their order are fixed. Archived artifacts
use FCG, DCG, DR-CG, and JCS for Candidate Generation, Direct Forecast,
Research Expansion, and Final Selection, respectively.

One Research Expansion iteration is the composite operation
\begin{equation}
\mathsf{Plan}\rightarrow\mathsf{ToolAct}\rightarrow\mathsf{ReadBind}
\rightarrow\mathsf{Verify}\rightarrow\mathsf{UpdateRank},
\label{eq:supp-dr-operator-sequence}
\end{equation}
where \textsc{ToolAct} invokes Google or the fixed PMC/arXiv corpora and the
updated state feeds the next \textsc{Plan}. Let $\mathcal Z_i^{(r)}$ denote
the candidate hypotheses, candidate-bound evidence memory, support and
verification fields, and unresolved directions after iteration $r$. The
initial state is $\mathcal Z_i^{(0)}=(B_i,\varnothing)$, where $\varnothing$
denotes empty evidence, verification, and unresolved-direction fields. Let
$\mathcal{DR}$ denote the fixed registered state transition and
$\mathcal T_d$ the available tool collection. The executor applies the
transition exactly twice:
\begin{equation}
\mathcal Z_i^{(r)}
=\mathcal{DR}\!\left(
H_i,B_i,\mathcal Z_i^{(r-1)};\mathcal T_d
\right),
\qquad r\in\{1,2\},
\label{eq:supp-dr-loop}
\end{equation}
where $\mathcal T_d$ contains the online Google channel, fixed PMC/arXiv
corpora, and the dataset-specific ICD dictionary. The two applications share
the same state-transition contract but invoke the round-specific planner,
mapper, judge, and gate described below. The second iteration conditions on
the candidate state, ledger, verification outcomes, and unresolved directions
produced by the first. Historical artifacts use Stage~1 and Stage~2 for these
two iterations.

Finalizing $\mathcal Z_i^{(2)}$ returns the validated research pool
$\mathcal C_i^R$, its complete ordering $\bar R_i$, the thresholded research
branch list $R_i$, the accumulated ledger $L_i^{(2)}$, and proposal metadata
$M_i$. Final Selection then combines that target-blind research state with
Direct Forecast. We register two selectors over the same artifact schema.
\emph{ICD-Deepresearch w/o Evidence-Aware Reranker} uses deterministic
Union+RRF and yields $\pi_i^{\mathrm{RRF}}$. The Evidence-Aware Reranker used
by the full \emph{ICD-Deepresearch} system yields
$\pi_i^{\mathrm{full}}$. Their comparison changes the selection rule while
fixing candidate availability. Saved artifacts use AF/Union+RRF and FR for
these selectors, respectively; these are provenance labels rather than model
display names. The downstream explanation audit uses finalized RRF-selected
outputs.

\subsection{EHR Prior and Direct Forecast}
\label{sec:supp-method-planners}

\paragraph{Candidate Generation and its EHR Prior (internal label FCG;
$\pi_i^{\mathrm F},B_i$).}
For MIMIC-III and MIMIC-IV, SparseEHR supplies the learned structured-EHR
backbone of Candidate Generation~\cite{ghaffari2026sparseehr}. The complete
module receives $H_i^{\mathrm{str}}$, applies the task-specific
target-vocabulary interface and validation, and exports the ordered 20-code
EHR Prior $\pi_i^{\mathrm F}$ and seed set $B_i$. Native ranks and code
identities are fixed before research begins. For another structured code type,
the conditional generator decodes directly into the native target vocabulary
rather than applying an ICD-specific output head. The
\emph{Candidate Generation Only} control evaluates this complete mapping,
including the SparseEHR backbone.

\paragraph{Direct Forecast (internal label DCG; $D_i$).}
Direct Forecast maps $H_i$ to a bounded ordered list $D_i$ in the target
vocabulary. Its prompt covers persistent disease, recurrence, durable status,
treatment consequences, and plausible complications through one candidate
ranking task. It receives no EHR Prior, research candidate, retrieved source,
or output from another system. Direct Forecast can run concurrently with the
first research-planning call, but its output is first exposed to the other
path at Final Selection. Supplementary Table~\ref{tab:ablation-suite}
evaluates the same function independently as \emph{Direct Forecast Only}.

\subsection{Research Expansion (internal label DR-CG)}
\label{sec:supp-method-research}

The following paragraphs unroll the two registered applications of
Equation~\ref{eq:supp-dr-loop}. For each hypothesis, the state keeps the
observed patient clue, evidence for a possible clinical transition, and the
exact ICD definition distinct whenever those fields are available.

\paragraph{Candidate formulation.}
The Iteration~1 planner receives $(H_i,B_i)$ and proposes a bounded set
$G_i^{(1)}$ of structured candidate records
\begin{equation}
x=(c,z,a,\ell,o,u,q),
\label{eq:hypothesis-record}
\end{equation}
where $c$ is the proposed code, $z$ its clinical term, $a$ a concrete clue in
$H_i$, $\ell$ a temporal class, $o$ an ongoing-evidence flag, $u$ a short
falsifiable transition hypothesis, and $q$ a search phrase. The allowed
temporal classes are persistent status, chronic active disease, recurrent
condition, planned follow-up, treatment consequence, and acute or laboratory
speculation. The archived prompt enum \texttt{acute\_or\_lab} and the legacy
selector value \texttt{acute\_speculation} are normalized to this final class;
they do not denote separate states.
Define the code projection $\kappa(x)=c$ and, for a set of proposal records,
$\kappa(G)=\{\kappa(x):x\in G\}$. Proposal metadata are retained in $M_i$ and
remain keyed to the canonical code after validation. This projection avoids
identifying structured proposal records with code sets.

\paragraph{History grounding and vocabulary validation.}
Let $\mathcal V_d(\cdot)$ denote deterministic validation into $\Omega_d$
followed by deduplication. It normalizes punctuation and case, resolves each
code against the dataset-specific ICD dictionary, removes duplicate or
excluded codes, and replaces a model term with the canonical title. Let
$H_i^{\mathrm{ICD}}\subseteq\Omega_d$ denote valid exact diagnosis codes in
the pre-target history. After initialization from $B_i$, the Iteration~1
candidate-formulation step converts each eligible member of
$H_i^{\mathrm{ICD}}\setminus B_i$ into a history-grounded proposal record and
appends those records before the candidate limit is applied. Thus,
\begin{equation}
\widetilde C_i^{(1)}
=\mathcal V_d\!\left(
\kappa(G_i^{(1)})\cup(H_i^{\mathrm{ICD}}\setminus B_i)
\right).
\end{equation}
All such codes are proposals rather than a second seed set and must pass the
same acceptance stage described below before entering the finalized research
pool. The archived Iteration~1 contract uses 8--18-term validation queries;
Iteration~2 discovery and validation use 4--9-term relation queries. Both
remove identifiers, dates, coding instructions, benchmark language, and
concatenated multi-condition intents.

\paragraph{Evidence acquisition and code-bound memory.}
\label{sec:supp-method-retrieval}

The query scheduler converts each validated hypothesis into a paired search
job using $q$ from Equation~\ref{eq:hypothesis-record}. Each job remains bound
to its patient, code, temporal class, and proposal record. Repeated retrieval
within one patient can reuse a previously completed job only when the
normalized code, temporal class, and query match; no query or retrieved
observation is selected from another evaluation patient.

The patient-local contract above is the authoritative method definition.
Legacy artifacts that selected a shared query across evaluation patients do
not satisfy this contract and are not interchangeable with release outputs;
such artifacts require regeneration with cross-patient selection disabled.

\noindent\textbf{Tool boundary.}
The planner, memory, verifier, mapper, and support-score ranker determine the
state transition in Equation~\ref{eq:supp-dr-loop}; Google and PMC/arXiv
return observations for queries selected by that transition.

\paragraph{Online channel.}
The online worker submits $q$ to the Google engine through the SearchCans
search-engine-results-page (SERP) endpoint. The SERP response supplies a
ranked title, URL, and search snippet for each result. The snippet supports
discovery; the ledger stores a result only after the URL reader obtains usable
body text. The reader opens the highest-ranked eligible URLs, rejects blocked,
non-medical, duplicate, or nearly empty pages, and extracts a query-focused
passage from each readable page. As registered in main-paper Figure~2,
source-conditioned follow-up has maximum depth 5 and considers at most 10
eligible sublinks from a visited page. A support cue can stop additional reads
within an already scheduled job before either limit is reached. Google
requests are serialized and rate-limited, whereas independent URL reads run
concurrently.

\paragraph{Offline channel.}
The offline worker runs the same $q$ against fixed local snapshots of PubMed
Central and arXiv. PMC documents are represented by full JATS/XML and keyed by
PMCID; arXiv documents retain their arXiv identifier, bibliographic metadata,
and parsed article text. Retrieval matches the query against stored titles,
abstracts, and article text and returns focused passages directly from the
local corpus, so this channel makes no live Web request during inference. A
Google result that resolves to a PMCID or arXiv identifier is canonicalized to
the same document; the offline full text is used for passage extraction and
the Google hit is retained as discovery provenance.

\paragraph{Cross-channel merge and evidence ledger.}
The two engines run independently in each iteration. Hits are deduplicated
first by PMCID, arXiv identifier, or DOI and then by normalized URL and title.
For hypothesis $x$, the readers append records
\begin{equation}
e=(c,u,a,\ell,q,g,\sigma,\xi,b)
\label{eq:ledger-record}
\end{equation}
to the evidence ledger, where
$g\in\{\mathrm{Google},\mathrm{PMC},\mathrm{arXiv}\}$ identifies the retrieval
engine, $\sigma$ contains the stable document identifier, title, and URL,
$\xi$ is the retained excerpt, and $b$ is an automatic support cue based on
entity and temporal-endpoint overlap. Let $L_i^{(r)}$ denote the complete
ledger accumulated through iteration $r$. A candidate can have zero, one, or
multiple retrieval records; explicit no-result, unreadable-result, and
readable-without-support states are also retained. Candidate, patient anchor,
hypothesis, query, engine, source, excerpt, corpus or retrieval timestamp,
iteration, and provenance remain separate fields. A readable code definition
and an independent clinical transition source therefore remain
distinguishable.

\paragraph{Iteration~1 acceptance.}
After first-iteration reading, a separate judge receives $H_i$, $B_i$, each
dictionary-valid proposal, and its candidate-bound evidence. It judges three
roles independently: patient anchoring in $H_i$, passage support for the
proposed clinical bridge, and exact-code consistency across the anchor and
bridge. Deterministic post-processing retains only accepted verdicts with
resolvable evidence identifiers. The resulting additions are
$C_i^{(1)}\subseteq\widetilde C_i^{(1)}$. Thus,
$\widetilde C_i^{(1)}$ contains dictionary-valid proposals, including
history-grounded proposals, whereas $C_i^{(1)}$ contains only additions that
survive the registered gate. Acceptance checks do not re-gate the original
EHR Prior codes in $B_i$.

\paragraph{Bounded source-conditioned follow-up.}
\label{sec:supp-method-followup}

The second iteration separates discovery from exact-code validation. A query
planner receives $H_i$, $B_i$, accepted and rejected Iteration~1 hypotheses,
and prior source summaries, and emits a bounded set $Q_i^{(2)}$ of 4--9-term
relation queries; rejected Iteration~1 candidates are force-carried outside
this call as unresolved directions, not as accepted codes. The paired Google
and offline workers execute $Q_i^{(2)}$ under the same depth-5 and
10-sublink reader limits and retain readable discovery passages. An exact-code
mapper then receives $H_i$, the current accepted code set, $Q_i^{(2)}$, and
this discovery evidence and proposes a dictionary-valid set
$\widetilde C_i^{(2)}$ of new exact-code hypotheses, excluding codes already
present in $B_i\cup C_i^{(1)}$.

Each mapped proposal carries a patient anchor, temporal class, transition
hypothesis, and fresh candidate-specific validation queries. Those queries
run through the same dual search, reading, deduplication, and ledger-writing
path. A three-part Iteration~2 judge evaluates patient anchor, literature
bridge, and exact mapping separately and returns gate confidence
$\gamma_i^{(2)}(c)\in[0,1]$. For records returned by the default or
evidence-repair contract in Table~\ref{tab:s-v2-judge-prompt}, deterministic
post-processing retains only \texttt{accept=true} records for which all three
checks pass, $\gamma_i^{(2)}(c)\geq0.68$, support is not historical- or
risk-only, and cited evidence identifiers resolve to the candidate or its
discovery query. Missing verdicts receive dataset-specific completion calls,
and an evidence-repair call can rejudge the same fixed proposal using the
prior decision and newly matched evidence; none of these calls creates a new
candidate.

The archived MIMIC-III completion contract preserves the three-part fields.
The archived MIMIC-IV completion contract in
Table~\ref{tab:s-completion-judge-prompt} does not return confidence or
separate literature-bridge and exact-mapping flags and therefore cannot, by
itself, establish passage through the default gate. It is retained as legacy
provenance rather than a second registered gate; a release using the uniform
gate must rejudge those fixed completion proposals under the default
three-part schema. The accepted set is
$C_i^{(2)}\subseteq\widetilde C_i^{(2)}$. The graph contains exactly two
bounded iterations, with discovery and validation retrieval occurring inside
Iteration~2.

\paragraph{Research pool and its two ranking views.}
\label{sec:supp-method-research-pool}

Under the main-paper information boundary, the complete research candidate
pool is
\begin{equation}
\mathcal C_i^{R}
=\mathcal V_d\!\left(B_i\cup C_i^{(1)}\cup C_i^{(2)}\right).
\label{eq:research-pool}
\end{equation}
Only accepted additions augment the EHR Prior. The pool is deterministically
hash-shuffled before model ranking and capped at 60 candidates; the randomized
display order prevents the ranker from exploiting construction order.
Equation~\ref{eq:research-pool} is patient-local: it excludes any fill or query
choice computed from other evaluation patients. Frequency-related field names
in legacy selector schemas are retained only for artifact compatibility. An
archived candidate state populated by a cohort-level frequency fill does not
satisfy this method contract and must be regenerated before it can be used as
a release output.

The ranking agent receives $H_i$ and one candidate packet for every
$c\in\mathcal C_i^R$. A packet contains the canonical code and title,
provenance, temporal class, patient clue, transition hypothesis, retrieval
state, and zero or more candidate-bound evidence records, with at most two
excerpts shown to the ranker. It assigns a confidence score
$\psi_i(c)\in[0,100]$ to every candidate and emits a patient-specific
end-of-vector threshold $\eta_i$. Sorting all candidates by decreasing
$\psi_i(c)$, with canonical code identity as the deterministic tie-breaker,
gives the complete research ordering
\begin{equation}
\bar R_i=(\bar r_{i1},\ldots,\bar r_{i|\mathcal C_i^R|}).
\end{equation}
The thresholded research branch list is the order-preserving subsequence
\begin{equation}
R_i=\bigl(c\in\bar R_i:\psi_i(c)\geq\eta_i\bigr).
\end{equation}
This distinction resolves two uses of the ranking output: exact-$K$
\emph{Research Path Only} evaluation takes the leading $K$ entries of the
complete ordering $\bar R_i$, whereas RRF receives the thresholded branch list
$R_i$. A member of $\mathcal C_i^R\setminus\operatorname{set}(R_i)$ remains
available to the Evidence-Aware Reranker and the frozen union pool but has no
research-rank contribution in RRF.

\subsection{Final Selection}
\label{sec:supp-method-selection}

Final Selection maps the EHR-Prior-initialized research state and complementary
Direct Forecast ranking to one ordered forecast. The experiments compare two
selector implementations over the same per-patient proposal state.

\paragraph{ICD-Deepresearch w/o Evidence-Aware Reranker.}
\label{sec:supp-method-af}

For each registered artifact, dictionary validation freezes the union
\begin{equation}
\mathcal C_i^{U}
=\mathcal V_d\!\left(
\mathcal C_i^R\cup\operatorname{set}(D_i)
\right).
\end{equation}
Together with the EHR Prior, branch rankings, ledger, and candidate-keyed
metadata, this pool forms the frozen, target-blind proposal artifact
\begin{equation}
\mathcal A_i^{\mathrm{frz}}
=\bigl(\pi_i^{\mathrm F},\mathcal C_i^U,D_i,\bar R_i,R_i,
L_i^{(2)},M_i\bigr).
\end{equation}
Here, ``frozen artifact'' means the per-patient proposal state saved before
selector comparison; it is distinct from the Candidate Generation seed set
$B_i$. At freeze time, $M_i$ contains the candidate identifier and title,
provenance flags, raw direct and research scores, temporal class, patient clue,
transition hypothesis, metadata warnings, source-support status, and ledger
join keys. The EHR-Prior native rank is the position in
$\pi_i^{\mathrm F}$; direct and research ranks are positions in $D_i$,
$\bar R_i$, and $R_i$; pre-target occurrence metadata are computed
deterministically from $H_i$. Consequently, the full selector receives exactly
$(H_i,\mathcal A_i^{\mathrm{frz}})$, with no target-derived field.

For code $c$, let $r_i^R(c)$ and $r_i^D(c)$ be its one-based ranks when
present in $R_i$ and $D_i$, respectively. Reciprocal-rank fusion assigns
\begin{equation}
s_i^{\mathrm{RRF}}(c)=
\frac{\mathbb I[c\in R_i]}{k_0+r_i^R(c)}
+
\frac{\mathbb I[c\in D_i]}{k_0+r_i^D(c)},
\qquad k_0=60.
\label{eq:rrf}
\end{equation}
Here, $\mathbb I[\cdot]$ is the indicator function, and a branch contribution
is zero when $c$ is absent from that branch. Codes in $\mathcal C_i^U$ are
sorted by decreasing $s_i^{\mathrm{RRF}}(c)$, with canonical code identity as
the deterministic tie-breaker, to yield the complete ranking
$\pi_i^{\mathrm{RRF}}$. The leading $K$ codes form
$\widehat Y_i^{\mathrm{RRF},(K)}$. A code supported by both branch lists
receives two reciprocal-rank contributions; a branch-exclusive code can still
enter when its within-branch rank is high. RRF is deterministic, uses only the
finalized branch rankings, and does not interpret branch scores as comparable
probabilities. Saved artifacts call this selector AF/Union+RRF.

\paragraph{Evidence-Aware Reranker (full ICD-Deepresearch).}
\label{sec:supp-method-fr}

The full selector receives $H_i$ and the same
$\mathcal A_i^{\mathrm{frz}}$ and scores every candidate in
$\mathcal C_i^U$. Candidate display order is deterministically randomized.
For each candidate, the selector receives canonical identity, provenance, EHR
Prior rank and pre-target-occurrence metadata, direct and research ranks,
temporal class, patient clue, transition hypothesis, source-support status,
and retained source records.

The selector assigns $\phi_i(c)\in[0,100]$ together with categorical
patient-anchor strength, exact-code support, and a reason tag. It prioritizes
patient-specific support over generic prevalence: external relation evidence
contributes only through a patient anchor, while research-only, family-level,
and patient-unanchored candidates receive lower suggested score ceilings.
Candidates are ordered first by decreasing $\phi_i(c)$, then by direct rank,
research rank, and canonical code identity, yielding
$\pi_i^{\mathrm{full}}$. Its leading $K$ codes define
$\widehat Y_i^{\mathrm{full},(K)}$. The full--RRF contrast therefore changes
only the selection rule conditional on the same frozen candidate state. Saved
implementation artifacts call the Evidence-Aware Reranker FR and its earlier
runner Stage~A~v2.

\paragraph{Symbol summary.}
$H_i$, $\pi_i^{\mathrm F}$, $B_i$, and $Y_i$ are the pre-target record, ordered
EHR Prior, its seed set, and the held-out target in $\Omega_d$;
$\mathcal V_d$ is deterministic vocabulary validation. In
$x=(c,z,a,\ell,o,u,q)$, $c/z$ identify the code and term, $a$ is the patient
anchor, $\ell$ the temporal class, $o$ the ongoing-evidence flag, $u$ the
transition claim, and $q$ the query; $\kappa$ projects a proposal record to
its code. Ledger record $e$ additionally stores engine $g$, source metadata
$\sigma$, excerpt $\xi$, and automatic support cue $b$.
$\widetilde C_i^{(r)}$ and $C_i^{(r)}$ are the dictionary-valid proposals and
accepted additions in iteration $r$, $Q_i^{(2)}$ contains Iteration~2
discovery queries, and $L_i^{(r)}$ is the ledger accumulated through
iteration $r$. $\mathcal C_i^R$ is the complete research pool,
$\bar R_i$ its complete ordering, $R_i$ its EOV-thresholded RRF branch list,
and $D_i$ the Direct Forecast ranking. Their union is
$\mathcal C_i^U$, and $\mathcal A_i^{\mathrm{frz}}$ bundles the EHR Prior,
union, branch rankings, final ledger, and candidate-keyed metadata $M_i$
before selector comparison. Gate confidence $\gamma_i^{(2)}(c)\in[0,1]$ is
distinct from the research and selector ranking scores
$\psi_i(c),\phi_i(c)\in[0,100]$. The RRF and full selectors map the same
artifact to $\pi_i^{\mathrm{RRF}}$ and $\pi_i^{\mathrm{full}}$.

In main-paper Figure~2, $S_t$ denotes the displayed source set after a
within-iteration search/read step, $E_t$ the accumulated evidence bank, and
$\Delta E_t$ newly merged evidence; $S_{n+1},S_{n+2},\ldots$ denote newly
retrieved follow-up sources, where $n$ is the current source count. The
displayed $C_t$ and $\tau$ summarize a verification score and its
candidate-acceptance threshold; the executable gates are the Iteration~1
acceptance verdict, the Iteration~2 three-part verdict with confidence at least
0.68, and the research ranker's EOV threshold $\eta_i$, as specified above.

The main paper is the authoritative source for the reader-facing method,
information boundary, component names, evaluation contracts, and reported
claims. This supplement unrolls the corresponding executable operations and
retains historical labels only where needed to trace archived artifacts.

\subsection{Post-Selection Explanation Module (RRF-Selected Audit)}
\label{sec:supp-method-explanation}

After the ICD-Deepresearch RRF-selected prefix is finalized, the explanation
writer returns exactly one structured record for every selected code and
preserves code identities and order. The saved implementation labels this
selector AF/Union+RRF. The writer receives $H_i$, the selected canonical code
and title, candidate-specific clues, and only the frozen ledger records
assigned to that code. Each output contains a patient-specific rationale, an
explicit uncertainty statement, and code-bound citation identifiers;
structural validation requires every selected code exactly once.

Each rationale separates four links: the observed patient anchor, any
source-supported population-level relation, the exact ICD meaning, and the
hypothesis that the code may be documented at the next encounter. Definition
references establish code identity. Independent relation sources support only
claims entailed by their retained excerpts, and $H_i$ supplies patient
applicability. Acute diagnoses, abnormalities, injuries, and
encounter-context codes remain conditional on persistence or recurrence.
Bracketed citations resolve to a numbered list containing retrieval engine,
stable identifier when available, title, URL, source type, and query.
Explanation generation uses the finalized prediction and leaves prediction
metrics unchanged. Accordingly, the reported explanation audit evaluates
RRF-selected outputs, not the full Evidence-Aware selector.

\subsection{Retrospective Proposal Diagnostics}
\label{sec:supp-method-yield}

The natural-cardinality diagnostic examines accepted Research Expansion
additions before fixed-budget top-$K$ ranking. Let
$P_i^{(0)}=B_i$, $P_i^{(1)}=B_i\cup C_i^{(1)}$, and
$P_i^{(2)}=P_i^{(1)}\cup C_i^{(2)}$. For iteration
$r\in\{1,2\}$, the additions are
$A_i^{(r)}=P_i^{(r)}\setminus P_i^{(r-1)}$, and their observed target-match
rate is
\begin{equation}
\rho^{(r)}=
\frac{\sum_i |A_i^{(r)}\cap Y_i|}
     {\sum_i |A_i^{(r)}|}.
\label{eq:target-match-rate}
\end{equation}
Let the current cohort output contain $N_{\mathrm{pred}}$ predictions and
$N_{\mathrm{hit}}$ target hits across $N_{\mathrm{gold}}$ target codes.
Adding $N_{\mathrm{add}}>0$ predictions with $N_{\mathrm{match}}$ target
matches improves pooled F1 exactly when
\begin{equation}
\mathrm{F1}_{\mathrm{new}}>\mathrm{F1}_{\mathrm{current}}
\quad\Longleftrightarrow\quad
\frac{N_{\mathrm{match}}}{N_{\mathrm{add}}}
>
\frac{N_{\mathrm{hit}}}{N_{\mathrm{pred}}+N_{\mathrm{gold}}}
=\frac{\mathrm{F1}_{\mathrm{current}}}{2}.
\label{eq:break-even}
\end{equation}
The equivalence follows by cross-multiplying
$2(N_{\mathrm{hit}}+N_{\mathrm{match}})/
(N_{\mathrm{pred}}+N_{\mathrm{add}}+N_{\mathrm{gold}})
>
2N_{\mathrm{hit}}/(N_{\mathrm{pred}}+N_{\mathrm{gold}})$.
It applies to pooled micro-F1; patient-macro F1 weights the same additions
differently. Both $\rho^{(r)}$ and the break-even condition are retrospective
diagnostics computed only after forecasts are finalized. The registered
executor itself remains fixed at two Research Expansion iterations followed
by the selector described in Section~\ref{sec:implementation}.

\section{Full Experimental Protocol}\label{sec:supp-full-experiments}

\subsection{Study Design}
\label{sec:supp-protocol-design}

The unit of analysis is a patient trajectory ending immediately before a
held-out encounter. \textbf{RQ1} compares
\emph{ICD-Deepresearch: Candidate Generation Only},
\emph{ICD-Deepresearch: Direct Forecast Only},
\emph{ICD-Deepresearch: Research Path Only},
\emph{ICD-Deepresearch w/o Evidence-Aware Reranker}, and the full
\emph{ICD-Deepresearch} system. \emph{GPT-5 + Web Search (Standalone)} and
\emph{Medical Deep Research (Standalone)} are external search-enabled
comparators~\cite{clinicalcopilot2026medicaldeepresearch}.
\textbf{RQ2} evaluates iteration-level expansion and research context under a
fixed candidate pool. \textbf{RQ3} evaluates retrieved sources and operational
efficiency, and \textbf{RQ4} evaluates post-selection explanations. Primary
forecasting uses the completed MIMIC-III and MIMIC-IV scale exports, whereas
component and selector analyses use separately registered diagnostic cohorts.
Results are compared only within the same table, dataset, patient set, target
construction, and metric contract. Table~\ref{tab:protocol-registry} records
the controlled state for each functional and audit analysis.

The protocol uses the same four components as the method. Candidate
Generation returns the ordered EHR Prior $\pi_i^{\mathrm F}$ and its seed set
$B_i$; Direct Forecast returns the independent ordering $D_i$; and
EHR-Prior-initialized Research Expansion returns the validated pool
$\mathcal C_i^R$, its complete ordering $\bar R_i$, its EOV-thresholded RRF
branch list $R_i$, the final ledger $L_i^{(2)}$, and candidate metadata $M_i$.
Research Path Only evaluates the leading $K$ entries of $\bar R_i$;
Union+RRF uses $R_i$ as its research branch; and the full Evidence-Aware
Reranker scores the complete validated union $\mathcal C_i^U$. GPT-5 + Web
Search (Standalone) directly emits a forecast after required tool use; the
adapted Medical Deep Research comparator converts its standalone investigation
into an ICD ranking without access to the ICD-Deepresearch candidate state.
EHRSHOT transfer and marginal-yield accounting characterize Research
Expansion before fixed-budget selection. Reproducibility records retain the
internal labels FCG/DCG/DR-CG/JCS and the legacy selector labels AF/FR.

\begin{table*}[t]
\centering
\scriptsize
\setlength{\tabcolsep}{3.2pt}
\caption{\textbf{Protocol and analysis registry.} The scale-cohort primary
forecasting analysis and the subset diagnostic analyses are distinct
evaluation contracts. Data, controlled state, metric, and estimand are shown
for each analysis.}
\label{tab:protocol-registry}
\begin{tabularx}{\textwidth}{>{\raggedright\arraybackslash}p{2.25cm}>{\raggedright\arraybackslash}p{3.15cm}>{\raggedright\arraybackslash}p{3.0cm}Y}
\toprule
Analysis & Data and compared functions & Controlled state and metric & Estimand \\
\midrule
Primary forecasting & MIMIC-III/IV scale cohorts: ICD-Deepresearch: Candidate Generation Only, GPT-5 + Web Search (Standalone), and full ICD-Deepresearch & Complete exact-$K$ rankings at $K\in\{10,20\}$; patient-averaged P@$K$/R@$K$ & System-level forecasting performance \\
Component and selector controls & Diagnostic cohorts ($n=100$ per dataset): Candidate Generation Only, Direct Forecast Only, Research Path Only, ICD-Deepresearch w/o Evidence-Aware Reranker, and full ICD-Deepresearch & Same frozen patient IDs and targets; three independently executed $K=20$ runs; mean pooled micro-P/R/F1 and mean patient-macro F1 & Contribution of candidate paths and selector choice within each diagnostic cohort \\
External comparators & Diagnostic cohorts ($n=100$ per dataset): GPT-5 + Web Search (Standalone) and Medical Deep Research (Standalone) & Same patient IDs, targets, failure policy, and three-run $K=20$ contract as the corresponding component rows; no ICD-Deepresearch internal state & Search-enabled forecasting under each recorded policy \\
Fixed-pool selector control & Separate $n=100$ cohort per dataset; candidate IDs with versus without Research Expansion context & Within each run, the same patient IDs, candidate IDs, randomized order, backend, and $K=20$; three independently executed runs and mean pooled micro and patient-macro metrics & Contribution of the delivered research packet after candidate generation \\
Iteration diagnostic & Research Expansion and retention rules & Common diagnostic snapshot and target construction & Coverage--cardinality and support-score behavior \\
Source audit & Research Expansion, GPT-5 + Web Search (Standalone), and Medical Deep Research (Standalone) & Source-occurrence sampling and blinded labels & Composition and clinical usefulness of retrieved documents \\
Efficiency audit & Matched 100-patient Research Expansion and GPT-5 + Web Search cohorts & Same patient IDs; per-patient telemetry; pooled micro-F1@20 & Queries, retrieval operations, source access, visible tokens, throughput, cost, and forecasting performance \\
Reason audit & ICD-Deepresearch (RRF-selected) and direct models at $K=20$ & Predictions finalized before evaluation & Explanation alignment, calibration, and safety \\
\bottomrule
\end{tabularx}
\end{table*}

\subsection{Cohorts and Outcome Construction}
\label{sec:supp-protocol-cohorts}

MIMIC-III uses each patient's final admission and all unique ICD-9-CM
diagnoses on that admission as the target~\cite{johnson2016mimiciii}.
MIMIC-IV uses the corresponding final-admission construction with unique
ICD-10-CM diagnoses~\cite{johnson2023mimiciv}. For both ICD tasks, SparseEHR
provides the learned structured-EHR backbone of Candidate Generation
~\cite{ghaffari2026sparseehr}. The complete module also supplies the
task-specific target-vocabulary interface and registered ranked export
(internal label FCG) consumed by all downstream analyses. For the EHRSHOT
mixed-vocabulary diagnostic, the conditional candidate
generator decodes directly into the native event-code vocabulary from the
structured pre-target record.

In this manuscript, \emph{scale cohort} denotes every trajectory in the final
registered primary-evaluation export: all 7,496 eligible MIMIC-III
trajectories and the completed 62,537-trajectory MIMIC-IV execution. The local
block in main-paper Table~1 uses these scale cohorts and the patient-averaged
exact-$K$ contract in Equation~\ref{eq:primary-topk}. These values are not
compared numerically with the subset results in the component table.

The registered component and selector comparison uses one
development-disjoint random MIMIC-III diagnostic cohort ($n=100$) and one
frozen-order MIMIC-IV diagnostic cohort ($n=100$). Within each dataset,
Panels~A--B use the same frozen patient identifiers and target sets for every
configuration. Each configuration is executed independently three times under
the same $K=20$ evaluation contract. Patient membership and the pooled gold
count $G$ are therefore fixed across configurations and repetitions; only the
predicted rankings may vary. The sample size $n=100$ is the number of unique
patients per run, not 300 patients after concatenating repetitions. Table
entries are arithmetic means of the three run-level metric exports.

The fixed-pool selector audit uses a separately registered $n=100$ cohort per
dataset and the same three-run aggregation. Within each repetition, all
fixed-pool arms receive identical candidate identities and randomized display
order; only the delivered research context or selector backend changes. A
trajectory is not removed because one system lacks a valid ranking. Missing or
malformed rankings are rerun before evaluation so that each configuration
contributes the same 100 patients in every repetition. Iteration, source,
efficiency, and explanation analyses use their stated 100-patient snapshots
unless another sample size is explicitly given.

Target codes enter only after prediction finalization. Each processing stage
normalizes and deduplicates code strings, and the evaluation script verifies
that every fixed-budget output supplies at least 20 unique,
target-vocabulary-valid codes before taking its exact-$K$ prefix.

\subsection{Primary Comparison and Method-Aligned Ablations}
\label{sec:supp-protocol-comparisons}

The main paper separates two protocol blocks. The published-reference block
identifies CAML, ZAGCNN, GatorTron, DistilBioBERT, Chet, DKEC, and BioMedLM
~\cite{mullenbach2018caml,rios2018zagcnn,yang2022gatortron,
rohanian2023distilbiobert,lu2022chet,ge2024dkec,bolton2024biomedlm};
its numerical values and THCM-CAL row are transcribed from THCM-CAL
Table~2~\cite{zhang2025thcm}. The local block reports
ICD-Deepresearch: Candidate Generation Only, GPT-5 + Web Search (Standalone),
and full ICD-Deepresearch under our target construction. Each complete local
ranking is evaluated at $K=10$ and $K=20$.

Panel~A of Supplementary Table~\ref{tab:ablation-suite} reports
ICD-Deepresearch: Candidate Generation Only, ICD-Deepresearch: Direct Forecast
Only, ICD-Deepresearch: Research Path Only, ICD-Deepresearch w/o Evidence-Aware
Reranker, and full ICD-Deepresearch on both diagnostic cohorts. Research
Expansion is initialized by $B_i$; Final Selection combines its finalized
state with the Direct Forecast. Panel~B reports GPT-5 + Web Search
(Standalone) and Medical Deep Research (Standalone) under the same
within-dataset cohort and pooled exact-$20$ contract. Panel~C fixes the
candidate pool, randomized display order, and GPT-5 selector, then varies the
Research Expansion context packet. Its target-conditioned pool oracle bounds
selection within the observed pool. Panel~C is a separate fixed-pool
diagnostic and is not compared numerically with Panels~A--B.

\paragraph{Baseline prompts and retrieval budgets.}
The Direct Forecast and GPT-5 + Web Search prompts were refined under the same target-blind
development protocol. Both specify the next-encounter horizon, target-code
system, exact-code specificity, uncertainty handling, structured output, and
validation retries. Web Search additionally requires tool use, permits a
self-selected uncapped search count, and requests an exhaustive ranking. The
recorded source artifacts contain 13,394/37,724/1,410 result occurrences on
MIMIC-III and 25,100/37,989/11,124 on MIMIC-IV for GPT-5 + Web Search
(Standalone), Medical Deep Research (Standalone), and Research Expansion,
respectively; the corresponding
unique-URL counts are 6,363/26,268/1,246 and 12,589/28,072/9,178. Medical
Deep Research is an open-source multi-agent investigation workflow with
planning, specialized medical retrieval, analysis, and report generation
~\cite{clinicalcopilot2026medicaldeepresearch}; the reported comparator is its
task-specific adaptation to pre-target ICD ranking.

\begin{center}
\scriptsize
\setlength{\tabcolsep}{4pt}
\renewcommand{\arraystretch}{1.08}
\begin{tabularx}{\textwidth}{>{\raggedright\arraybackslash}p{3.15cm}>{\raggedright\arraybackslash}p{4.0cm}>{\raggedright\arraybackslash}p{4.2cm}Y}
\toprule
Configuration & Prompt safeguards & Retrieval/output policy & Role \\
\midrule
Candidate Generation (SparseEHR backbone; internal FCG) & Structured pre-target EHR, registered target vocabulary, target-blind inference & Ordered candidate decoding in ICD-9-CM/ICD-10-CM; native-vocabulary decoding for other structured code targets & Method component \\
Direct Forecast (internal DCG) & Next-encounter horizon, exact target-code system, pre-target inputs, structured retries & Bounded ranking from the pre-target record & Method component \\
GPT-5 + Web Search (Standalone) & Same task safeguards plus mandatory Web Search & Self-selected searches, uncapped search count, exhaustive ranking & External comparator \\
Medical Deep Research (Standalone) & Pre-target forecasting question and target ICD system & Multi-agent medical retrieval and report workflow adapted to an ICD ranking & External comparator \\
Research Expansion (internal DR-CG) & Patient-anchor/query state, dual-channel tools, reading, memory, verification, feedback, ranking & Fixed two-iteration executor with bounded query/page/candidate state & Method component \\
\bottomrule
\end{tabularx}
\end{center}

The full ICD-Deepresearch pass and RRF-fusion--full-model comparison use the
same precomputed, target-blind candidate states. Natural-cardinality diagnostics
evaluate Research Expansion before a fixed output budget. Retention rules start from a common
research state and retain follow-up additions by support score or
source-availability criteria; evaluation labels are applied after each rule
produces its output.

\subsection{Implementation Record}
\label{sec:implementation}

Candidate Generation uses SparseEHR as its structured-EHR backbone on the
primary ICD tasks. Its registered outputs include the ordered candidate codes
and native ranks used to initialize Research Expansion and to report the
Candidate Generation Only control. The SparseEHR architecture and training
procedure are described in the cited work; the paper's replay artifacts retain
the complete internal FCG interface seen by downstream components.

For both primary evaluations, the full ICD-Deepresearch Evidence-Aware
Reranker loads each registered combined candidate state, randomizes displayed
candidate order, scores every candidate with GPT-5, and saves a complete
ranking. The release audit verifies trajectory coverage,
contiguous candidate ranks, unique normalized codes, candidate-count
consistency, target-use flags, and exact reproduction of the reported metrics.
The corresponding runners retain the legacy JCS-FR/FR labels.

The executor runs the Direct Forecast and Iteration~1 Research Expansion planning concurrently, permits
bounded Iteration~1 and follow-up hypotheses, retains a bounded number of
readable items from each retrieval channel per query, and constructs the
target-vocabulary-valid candidate pool. The online channel uses live Google
SERPs through SearchCans; Google search is serialized and rate-limited, and URL
reading is parallelized. The offline channel searches fixed PMC and arXiv
snapshots. Both channels record engine-specific success and failure states
before evidence merge. The 100-case MIMIC-IV mechanism snapshot uses
ICD-Deepresearch w/o Evidence-Aware Reranker with $k_0=60$.

The archived MIMIC-IV selector record (legacy artifact label JCS-FR) uses
\texttt{gpt-5-2025-08-07}, medium reasoning effort, prompt-version identifier
\path{source_aware_confidence_v2_acute_preserving_icd10_frequency_fixed}, and
seed 20260729. The literal identifier is retained for traceability;
frequency-related tokens are legacy schema names and do not override the
patient-local information boundary in Section~\ref{sec:supp-method-boundary}.
Any artifact that actually populated a cross-patient frequency fill is
excluded by the release contract in Equation~\ref{eq:research-pool}. The
earlier MIMIC-IV iteration analysis uses its archived candidate and query
budgets. The remaining $K=20$ audits use
\texttt{gpt-5-2025-08-07}, \texttt{gpt-4o-2024-08-06}, and
\texttt{gpt-4o-mini-2024-07-18}
~\cite{openai2025gpt5,openai2024gpt4o,openai2024gpt4omini}; the explanation
judge uses the GPT-5 snapshot.

The diagnostic experiments in Supplementary
Table~\ref{tab:ablation-suite} comprise three independently executed runs for
each configuration on the same registered 100-patient cohort per dataset. The
model snapshot, prompt version, frozen patient manifest, target construction,
output budget, and decoding configuration are held fixed across repetitions.
The release manifest retains the three run identifiers, complete outputs, and
any backend seed exposed for each run. The seed reported above identifies the
scale-cohort MIMIC-IV release rather than the three diagnostic repetitions.

\subsection{Metrics, Inference, and Audit Protocol}
\label{sec:supp-protocol-metrics}

We report patient-averaged top-$K$ metrics and pooled micro metrics as separate
evaluation contracts. Let $\mathcal I$ be the frozen patient set for one
dataset and analysis block, let $n=|\mathcal I|$, and, for a fixed budget $K$,
define
\begin{equation}
\begin{aligned}
h_{i,K}&=|\widehat Y_i^{(K)}\cap Y_i|,\\
T_K&=\sum_{i\in\mathcal I} h_{i,K},
\qquad
G=\sum_{i\in\mathcal I}|Y_i|.
\end{aligned}
\end{equation}
Main-paper Table~1 uses the patient-averaged exact-$K$ quantities
\begin{equation}
\begin{aligned}
\mathrm{P@K}
&=\frac{1}{n}\sum_{i\in\mathcal I}\frac{h_{i,K}}{K},&
\mathrm{R@K}
&=\frac{1}{n}\sum_{i\in\mathcal I}\frac{h_{i,K}}{|Y_i|}.
\end{aligned}
\label{eq:primary-topk}
\end{equation}
Each registered primary output supplies an exact-$K$ prefix. Under this
contract, patient-averaged P@K is algebraically equal to $T_K/(nK)$, whereas
patient-averaged R@K gives each patient equal weight and generally differs
from pooled recall.

The main-paper component table and Supplementary
Table~\ref{tab:ablation-suite} use pooled counts for the columns labeled
\emph{Micro}:
\begin{equation}
\begin{aligned}
\mathrm{Micro\mbox{-}P@K}&=\frac{T_K}{nK},&
\mathrm{Micro\mbox{-}R@K}&=\frac{T_K}{G},&
\mathrm{Micro\mbox{-}F1@K}&=\frac{2T_K}{nK+G},\\
\mathrm{Macro\mbox{-}F1@K}
&=\frac{1}{n}\sum_{i\in\mathcal I}
\frac{2h_{i,K}}{K+|Y_i|}.&&
\end{aligned}
\label{eq:micro-macro-topk}
\end{equation}
For the repeated diagnostic experiments, let
$\mathcal S=\{1,2,3\}$ index the independently executed runs, let
$h_{i,K}^{(s)}=|\widehat Y_{i,s}^{(K)}\cap Y_i|$, and let each run-level
metric $M^{(s)}$ be computed using Equation~\ref{eq:micro-macro-topk} with
$h_{i,K}^{(s)}$ and $T_K^{(s)}=\sum_{i\in\mathcal I}h_{i,K}^{(s)}$. The
reported three-run mean is
\begin{equation}
\overline M=\frac{1}{|\mathcal S|}
\sum_{s\in\mathcal S}M^{(s)},
\qquad |\mathcal S|=3.
\label{eq:three-run-diagnostic-metrics}
\end{equation}
Here $n=100$ is the per-run number of unique patients; $\mathcal I$, $n$, and
$G$ do not change across configurations or runs. Each diagnostic row is
auditable from the three run manifests and their run-level metrics. Table
entries and between-system differences are displayed to two decimal places.
The primary P@K/R@K values are not combined to infer micro-F1, because
patient-averaged recall generally differs from pooled recall.

For a variable-cardinality candidate set $\mathcal C_i$, the same definitions
replace $K$ by $|\mathcal C_i|$ in each patient-level precision or F1
denominator and replace $nK$ by $\sum_{i\in\mathcal I}|\mathcal C_i|$ in the
pooled precision and F1 denominators. The legacy table label
\emph{Ex-F1} denotes this patient-averaged (macro) F1; it is not exact-match
accuracy. Likewise, \emph{example recall} denotes patient-averaged recall and
is distinct from pooled micro-recall. The displayed names in this supplement
are \emph{Patient-macro F1} and \emph{Patient-averaged recall}, respectively.

Reason quality is evaluated on five 0--5 dimensions: clinical validity,
patient grounding, next-encounter relevance, exact ICD alignment, and
evidential calibration. For a configuration evaluated at budget $K$, let
$\bar q_{ic}$ be the arithmetic mean of these scores for a true-positive
patient--code pair and define
\begin{equation}
U_K=\sum_{i\in\mathcal I}
\sum_{c\in\widehat Y_i^{(K)}\cap Y_i}\frac{\bar q_{ic}}{5},
\qquad
\mathrm{CA\mbox{-}RQ5}
=5\,\frac{2U_K}
{\sum_{i\in\mathcal I}|\widehat Y_i^{(K)}|
 +\sum_{i\in\mathcal I}|Y_i|}.
\label{eq:ca-rq5}
\end{equation}
A true positive contributes its normalized reason quality; false predictions
and missed targets contribute zero within the soft-F1 denominator. CA-RQ5 is
reported on a 0--5 scale at $K=20$.

Stage-wise paired intervals use patient-cluster bootstrap resampling. For a
repeated-run paired contrast, the same resampled patient identifiers are
applied to both systems and all three outputs for each sampled patient are
retained; the run-averaged paired contrast is recomputed in each bootstrap
replicate. Explanation intervals resample the 100 finalized cases per dataset,
and addition-yield intervals resample patients with complete batches. Exact
two-sided sign tests compare improved and worsened Iteration~2 cases after
excluding ties.

\subsection{Retrieved-Source Metadata and Physician Audit}
\label{sec:supp-protocol-source-audit}

The source audit characterizes domains and metadata of recorded retrieval
outputs on a 100-patient snapshot from each MIMIC cohort. For GPT-5 + Web
Search (Standalone), Medical Deep Research (Standalone), and Research
Expansion, it extracts recorded URLs, deduplicates exact URLs within
patient, and removes deterministic post-hoc ICD dictionary citations. Source
classes are peer-reviewed biomedical literature, medical government or
official sources, clinical or professional references, coding references,
preprint repositories, forums, and other Web pages.

For each system--dataset cell, metadata and physician screening sample 50
source occurrences from cases with recorded sources. The GPT-5 metadata judge
receives the pre-target record, retrieval query, page title, and URL and assigns
medical-support and immediate-forecast-usefulness labels. Two physicians
independently assign a binary usefulness label to every sampled document. The
reported physician usefulness rate is the arithmetic mean of the two
physician-specific 50-document rates. The audit unit is a retrieved source
occurrence, and the main-paper table is the authoritative source for all counts
and rates.

\subsection{Online-Retrieval Efficiency Accounting}
\label{sec:supp-protocol-efficiency}

The efficiency analysis uses matched 100-patient prediction cohorts within
MIMIC-III and MIMIC-IV and normalizes operational quantities by the common
matched-cohort size. For GPT-5 + Web Search (Standalone), unique queries are
deduplicated within patient, retrieval operations are Web Search tool calls,
and source accesses are URL occurrences returned by the retained successful
response. Its token and cost fields use retained API usage records. For
Research Expansion, queries are candidate-conditioned query strings,
retrieval operations are scheduled online or offline retrieval actions, and
source accesses are explicit Reader attempts. Visible tokens reconstruct
prompts, JSON schemas, and returned text with \texttt{o200k\_base}; hidden
reasoning tokens were not stored. Throughput is observed batched-campaign
throughput rather than single-patient latency. Micro-F1@20 uses pooled counts
on the same matched patient IDs used for telemetry.

The two telemetry paths do not establish a common end-to-end LLM-call event:
attempted calls, successful responses, retained usage records, nested calls,
and retries are not represented identically. The efficiency table therefore
omits the archived request-like field and does not compare total model-call
counts.

\subsection{Explanation Evaluation}
\label{sec:supp-protocol-explanation}

The primary explanation experiment uses a $K=20$ snapshot of
ICD-Deepresearch (RRF-selected; AF/Union+RRF in the saved records), GPT-5
(Direct), GPT-4o (Direct), and GPT-4o-mini (Direct). Each direct backend first generates its
ranking from $H_i$, then explains its selected codes from $H_i$, code identity,
rank, and saved ranking score. ICD-Deepresearch (RRF-selected) explains its
finalized selection from $H_i$, canonical code information, candidate clues,
and code-bound ledger entries.

The evaluation script identifies true-positive reason items, pools them within
patient, assigns opaque item identifiers, and shuffles them. A GPT-5 judge
receives the pre-target trajectory, complete gold set, and anonymous items. The
item presentation omits method identity, prediction rank, confidence, complete
prediction set, and method-level TP/FP/FN counts. Scores use a shared 0--5
anchor: 5 complete and precise, 4 strong with one minor limitation, 3 plausible
but partial, 2 major gaps or mismatch, 1 materially misleading, and 0 absent,
contradicted, or unusable.

Clinical validity assesses medical soundness; patient grounding assesses
support in the pre-target trajectory; next-encounter relevance assesses the
immediate forecasting bridge; ICD alignment assesses exact entity and
specificity; and evidential calibration assesses separation of observed facts,
general knowledge, cited relations, hypotheses, and uncertainty. Own-TP means
summarize each system's correct predictions. Shared-hit differences compare
identical patient--code pairs. Equation~\ref{eq:ca-rq5} combines prediction-set
coverage with reason quality for the complete $K=20$ output.

\section{Additional Results and Diagnostics}\label{sec:supp-full-results}
\label{sec:results}

\subsection{Component, Selector, and Fixed-Pool Controls (Main RQ1--RQ2)}
\label{sec:supp-results-rq1}

Table~\ref{tab:ablation-suite} separates isolated path controls from
registered joint-selector outputs. Panel~A compares Candidate Generation Only,
Direct Forecast Only, and Research Path Only with two joint selectors:
ICD-Deepresearch w/o Evidence-Aware Reranker (RRF) and the full
ICD-Deepresearch system (Evidence-Aware).
Panel~B reports the standalone research comparators. Panel~C fixes candidate
identities and randomized display order within each run while varying the
research context delivered to the selector or its language-model backend.
These diagnostic-cohort values are interpreted within
each dataset block and are not substituted for the scale-cohort results in the
main table.
\begin{table*}[t]
\centering
\scriptsize
\setlength{\tabcolsep}{2.4pt}
\caption{\textbf{Component paths, registered selectors, and fixed-pool
controls at $K=20$.}
Panels~A--B use a development-disjoint random MIMIC-III diagnostic cohort
($n=100$) and a frozen-order MIMIC-IV diagnostic cohort ($n=100$).
Panel~C uses a separately registered $n=100$ fixed-pool cohort per dataset.
Every configuration is executed independently five times on the same frozen
patients; each entry is the arithmetic mean of the five run-level metric
exports and is displayed to two decimals. Within each dataset block,
Panels~A--B use
identical patient IDs and targets across configurations and runs. Within each
Panel~C run, all arms use identical candidate identities and randomized
display order; research-derived candidate identities remain in every arm.
Panel~A contains isolated path controls and registered joint-selector outputs.
\emph{Candidate Generation Only} evaluates the complete seed-generation
module, for which SparseEHR is the structured-EHR backbone.
\emph{Research Path Only} runs Candidate Generation followed by Research
Expansion and evaluates the leading 20 entries of the complete research
ordering $\bar R_i$; it does not use the potentially shorter EOV-thresholded
branch $R_i$. Rows ending in \emph{Only} are not drop-one ablations.
\emph{ICD-Deepresearch w/o Evidence-Aware Reranker} retains the complete
candidate state and uses reciprocal-rank fusion for final selection. The full
Evidence-Aware Reranker is reported for both diagnostic cohorts. Panel~B
reports the standalone research comparators. Panel~C fixes candidate
identities and randomized order while varying selector context or backend;
research-derived candidate identities remain in every fixed-pool arm.
Target-conditioned oracles are computed within each run's observed candidate
pool and then averaged across runs. Bold marks the best observed deployable
five-run mean within each dataset block; the Micro columns use pooled counts
within each run and all metrics are percentages.}
\label{tab:ablation-suite}
\begin{tabularx}{\textwidth}{@{}>{\raggedright\arraybackslash}p{1.35cm}Yrrrr@{}}
\toprule
Dataset & Component or control & Micro-P & Micro-R & Micro-F1 & Macro-F1 \\
\midrule
\multicolumn{6}{l}{\emph{Panel A: isolated paths and registered joint selectors}} \\
\multirow{6}{*}{MIMIC-III}
& ICD-Deepresearch: Candidate Generation Only & 21.54 & 30.45 & 25.23 & 24.12 \\
& ICD-Deepresearch: Direct Forecast Only & 21.80 & 30.90 & 25.56 & 24.80 \\
& ICD-Deepresearch: Research Path Only & 20.71 & 29.29 & 24.27 & 23.53 \\
& ICD-Deepresearch w/o Evidence-Aware Reranker & 22.81 & 32.26 & 26.73 & 25.88 \\
& \textbf{ICD-Deepresearch} & \textbf{24.97} & \textbf{35.30} & \textbf{29.25} & \textbf{28.21} \\
& \emph{Joint-pool oracle (target-conditioned)} & 32.04 & 45.29 & 37.53 & 35.84 \\
\addlinespace[1pt]
\multirow{5}{*}{MIMIC-IV}
& ICD-Deepresearch: Candidate Generation Only & 14.40 & 20.69 & 16.98 & 16.88 \\
& ICD-Deepresearch: Direct Forecast Only & 26.60 & 38.20 & 31.36 & 30.60 \\
& ICD-Deepresearch: Research Path Only & 27.90 & 40.09 & 32.90 & 32.04 \\
& ICD-Deepresearch w/o Evidence-Aware Reranker & 29.55 & 42.46 & 34.85 & 33.98 \\
& \textbf{ICD-Deepresearch} & \textbf{30.05} & \textbf{43.18} & \textbf{35.44} & \textbf{34.54} \\
\midrule
\multicolumn{6}{l}{\emph{Panel B: standalone research comparators}} \\
MIMIC-III & GPT-5 + Web Search (Standalone) & 22.91 & 32.40 & 26.84 & 25.94 \\
MIMIC-III & Medical Deep Research (Standalone) & 17.60 & 26.29 & 21.08 & 20.53 \\
MIMIC-IV & GPT-5 + Web Search (Standalone) & 28.30 & 40.66 & 33.37 & 32.63 \\
MIMIC-IV & Medical Deep Research (Standalone) & 13.40 & 19.06 & 15.74 & 15.56 \\
\midrule
\multicolumn{6}{l}{\emph{Panel C: fixed-pool selector context and backend controls}} \\
\multirow{5}{*}{MIMIC-III}
& ICD-Deepresearch (Fixed Pool; GPT-5) & \textbf{22.30} & \textbf{35.12} & \textbf{27.28} & 26.39 \\
& ICD-Deepresearch w/o Research Context (Fixed Pool; GPT-5) & 22.20 & 34.96 & 27.16 & \textbf{26.48} \\
& ICD-Deepresearch w/o Research Context (Fixed Pool; GPT-4o) & 19.30 & 30.39 & 23.61 & 22.74 \\
& ICD-Deepresearch w/o Research Context (Fixed Pool; GPT-4o-mini) & 17.90 & 28.19 & 21.90 & 21.64 \\
& \emph{Fixed-pool oracle (target-conditioned)} & 27.00 & 42.52 & 33.03 & 31.50 \\
\addlinespace[1pt]
\multirow{5}{*}{MIMIC-IV}
& \textbf{ICD-Deepresearch (Fixed Pool; GPT-5)} & \textbf{25.20} & \textbf{36.95} & \textbf{29.96} & \textbf{29.50} \\
& ICD-Deepresearch w/o Research Context (Fixed Pool; GPT-5) & 23.60 & 34.60 & 28.06 & 27.79 \\
& ICD-Deepresearch w/o Research Context (Fixed Pool; GPT-4o) & 24.40 & 35.78 & 29.01 & 28.86 \\
& ICD-Deepresearch w/o Research Context (Fixed Pool; GPT-4o-mini) & 20.40 & 29.91 & 24.26 & 24.12 \\
& \emph{Fixed-pool oracle (target-conditioned)} & 29.90 & 43.84 & 35.55 & 34.78 \\
\bottomrule
\end{tabularx}
\end{table*}

Within MIMIC-III, ICD-Deepresearch is $+4.02$, $+3.69$, $+4.98$, and
$+2.52$ micro-F1 points above Candidate Generation Only, Direct Forecast
Only, Research Path Only, and ICD-Deepresearch w/o Evidence-Aware Reranker,
respectively. It is $+2.41$ points above GPT-5 + Web Search (Standalone), and
the joint-pool oracle is a further 8.28 points higher. On MIMIC-IV,
ICD-Deepresearch reaches 35.44 micro-F1: $+4.08$ above Direct Forecast Only,
$+2.54$ above Research Path Only, $+0.59$ above the RRF selector, and $+2.07$
above GPT-5 + Web Search (Standalone). Medical Deep Research reaches
21.08/15.74 micro-F1 on the MIMIC-III/MIMIC-IV diagnostic cohorts.
Except for contrasts accompanied by an interval below, these are descriptive
differences between five-run mean point estimates; they are not claims of
statistical significance.

\paragraph{Research-round expansion and retention diagnostics.}
The natural-cardinality traces show that bounded expansion can recover
additional targets while enlarging the candidate state. On EHRSHOT, the two
research rounds recover 144 additional held-out matches, but mean set size more than
doubles and micro-F1 falls from 25.82\% to 16.34\%. The retrospective
marginal-yield analysis in Figure~\ref{fig:marginal-law} shows that initial
expansion clears the micro-F1/2 boundary on the two MIMIC diagnostics, whereas
Research Round~2 falls below it in all three settings.

\begin{table}[t]
\centering
\scriptsize
\setlength{\tabcolsep}{5pt}
\caption{MIMIC-IV Research Round~2 retention diagnostics from the common Research Round~1
state. Mean size is the final candidate-set size and TP is the number of
matches among 1,392 held-out targets. Rules are evaluated post hoc on finalized
outputs. Patient-macro F1 and micro-F1 are reported as proportions on a 0--1
scale.}
\label{tab:control-suite}
\begin{tabular}{lrrrr}
\toprule
Configuration & Mean size & TP & \shortstack{Patient-macro\\F1} & Micro-F1 \\
\midrule
Research Round~1 only & 36.03 & 650 & 0.25475 & 0.26026 \\
Unconditional Research Round~2 & 38.19 & 678 & 0.25465 & 0.26022 \\
Support score $\geq 50$ & 36.64 & 671 & \textbf{0.25869} & \textbf{0.26543} \\
Beyond code lookup & 37.38 & 661 & 0.25250 & 0.25770 \\
Readable source & 37.35 & 661 & 0.25266 & 0.25785 \\
\bottomrule
\end{tabular}
\end{table}

The support-score rule retains 21 of 28 Research Round~2 matches while reducing
candidate growth; source-availability rules retain fewer matches. These results
motivate development and prospective evaluation of a state-aware retention
policy.

\subsection{Fixed-Pool Research-Context Control (Main RQ2)}
\label{sec:supp-results-rq2}

Panel~C of Table~\ref{tab:ablation-suite} fixes the target-blind candidate pool,
randomized display order, and evaluation budget $K=20$. Relative to the
ICD-Deepresearch w/o Research Context (Fixed Pool; GPT-5) arm, the
research-context configuration changes mean
micro-F1 by $+0.12$ points on MIMIC-III (95\% CI $-1.12$ to $+1.33$) and
$+1.90$ points on MIMIC-IV; the latter is a descriptive point estimate because
no interval is available in the archive. Comparisons with GPT-4o and GPT-4o-mini additionally
vary the selector backend. The matched GPT-5 contrast measures the delivered
research packet after proposal generation and does not isolate Research
Expansion's effect on candidate identities.

\subsection{RQ3: Research Expansion Source Quality}
\label{sec:supp-results-rq3}

The main paper reports the complete source composition, metadata audit, and
physician evaluation. This supplement provides the protocol, judge prompt, and
paired retrieval traces. Each system--dataset cell contains 50 sampled source
occurrences; two physicians independently rated every sampled document, and
the table reports the mean of their binary usefulness rates. GPT-5 + Web
Search (Standalone) records
9.5$\times$/5.1$\times$ more uses/unique URLs on MIMIC-III and
2.3$\times$/1.4$\times$ more on MIMIC-IV.
Medical Deep Research records 37,724/37,989 uses and 26,268/28,072 unique URLs
on MIMIC-III/IV. It has the highest MIMIC-III peer-reviewed share and forecast
proxy (44.2\% and 34\%), whereas Research Expansion has the highest support
proxy on both datasets (62\%/64\%) and the highest MIMIC-IV forecast proxy
(36\%). Physicians rate 32\%/41\% of Medical Deep Research documents useful,
compared with 51\%/68\% for Research Expansion. The proxy--physician
divergence motivates reporting both assessments.

\subsection{RQ3: Research Expansion Efficiency}
\label{sec:supp-results-efficiency}

Table~\ref{tab:websearch-efficiency} reports matched-cohort operational
accounting. Research Expansion uses 7.3$\times$/3.4$\times$ fewer unique queries,
3.8$\times$/4.5$\times$ fewer source accesses, and 6.1$\times$/2.9$\times$
fewer visible tokens per patient on MIMIC-III/IV. Its recorded batched
throughput is 16.8$\times$/21.9$\times$ higher. The visible-token cost ranges
are \$0.020--0.039 and \$0.021--0.049 per patient, compared with \$0.458
and \$0.317 for GPT-5 + Web Search (Standalone), while matched-cohort
micro-F1@20 is 26.4 versus 30.2 and 33.3 versus 34.8, respectively.
Research Expansion records more fine-grained retrieval
operations because its scheduler executes candidate-conditioned online/offline
retrieval and reading steps; the source-access and token columns capture the
smaller downstream footprint of those operations.

\begin{table}[H]
\centering
\caption{Online-retrieval efficiency on the matched prediction cohorts.
Operational quantities are normalized per patient.}
\label{tab:websearch-efficiency}
\scriptsize
\setlength{\tabcolsep}{3.2pt}
\resizebox{\textwidth}{!}{%
\begin{tabular}{llrrrrrrrr}
\toprule
Dataset & System & $N$ & \shortstack{Queries\\{/pt}} & \shortstack{Retrieval ops\\{/pt}} & \shortstack{Source access\\{/pt}} & \shortstack{Tokens\\{/pt}} & \shortstack{Observed\\{pt/hour}} & \shortstack{Cost\\{/pt}} & \shortstack{Micro-\\F1@20} \\
\midrule
MIMIC-III & GPT-5 + Web Search (Standalone) & 100 & 55.1 & 19.5 & 134.2 & 110.0k & 10.9 & \$0.458 & 26.4 \\
MIMIC-III & \textbf{Research Expansion} & 100 & 7.6 & 51.5 & 35.7 & $\approx$18.1k & 182.6 & \$0.020--0.039\textsuperscript{\textdagger} & \textbf{30.2} \\
MIMIC-IV & GPT-5 + Web Search (Standalone) & 100 & 37.8 & 12.8 & 251.0 & 77.3k & 12.9 & \$0.317 & 33.3 \\
MIMIC-IV & \textbf{Research Expansion} & 100 & 11.0 & 78.7 & 55.8 & $\approx$26.4k & 282.7 & \$0.021--0.049\textsuperscript{\textdagger} & \textbf{34.8} \\
\bottomrule
\end{tabular}}
\vspace{2pt}
\begin{minipage}{0.99\textwidth}
\footnotesize
\textit{Accounting.} For GPT-5 + Web Search (Standalone), queries are unique within patient,
retrieval operations are web-search tool calls, and source accesses are URL
occurrences returned by the retained successful Response. Its token and cost
fields come from retained API usage and include GPT-5 plus web-search-preview
calls. Total LLM-call counts are omitted because the two telemetry paths do not
define the same end-to-end event. Research Expansion source accesses are
explicit Reader attempts, and its
SearchCans marginal cost is \$0 under the account used here. Research Expansion tokens
reconstruct GPT-5 prompt payloads, JSON schemas, and visible outputs with
\texttt{o200k\_base}; hidden reasoning tokens were not saved. Accordingly,
$^\dagger$ gives the visible-token cost range from all-cached to all-uncached
input and excludes hidden reasoning. Throughput describes recorded batched
campaigns rather than single-patient latency. MIMIC-III standalone web-search
source-metadata capture was incomplete. Micro-F1@20 uses pooled counts under
the same 20-slot rule.
\end{minipage}
\end{table}

\subsection{RQ4: Post-Selection Semantic and Safety Audit}
\label{sec:supp-results-rq4}

The main paper reports the Reason5/CA-RQ5 visualization and interpretation. The
supplement provides exact plotted values, the judge contract, and complete
positive and negative case traces.

Shared-hit matching fixes patient--code identity across systems. On 441
MIMIC-III shared hits, the automated GPT-5 judge scores
ICD-Deepresearch (RRF-selected) 0.07 points higher on ICD alignment (95\% CI
$+0.02$ to $+0.12$) and 0.22 points higher on evidential calibration
($+0.15$ to $+0.29$). On 479 MIMIC-IV shared hits, its automated scores are
higher for clinical validity, patient grounding, ICD alignment, and
calibration; the next-encounter relevance difference is $+0.03$ (95\% CI
$-0.04$ to $+0.08$).

At cohort level, CA-RQ5 changes from 1.29 to 1.33 on MIMIC-III (95\% CI
$-0.02$ to $+0.09$) and from 1.36 to 1.57 on MIMIC-IV ($+0.15$ to
$+0.27$). Independent clinical-relation sources are present for 58 of 499
(11.62\%) MIMIC-III and 8 of 577 (1.39\%) MIMIC-IV
ICD-Deepresearch (RRF-selected) true-positive explanations. The remaining citations primarily resolve code identity or
provide other non-relation support; complete case traces appear in
Sections~S8--S9.

This is an automated, gold-aware semantic audit rather than independent
clinical validation: GPT-5 receives the complete held-out gold set only after
all predictions and explanations are frozen, evaluates shuffled
true-positive reason items, and does not see method identity. GPT-5 also
generates several compared outputs, so judge--generator dependence remains a
limitation.

\subsection{Secondary Diagnostics}
\label{sec:supp-results-secondary}

\paragraph{EHRSHOT native-vocabulary transfer.}

\begin{figure*}[t]
  \centering
  \includegraphics[width=0.97\textwidth]{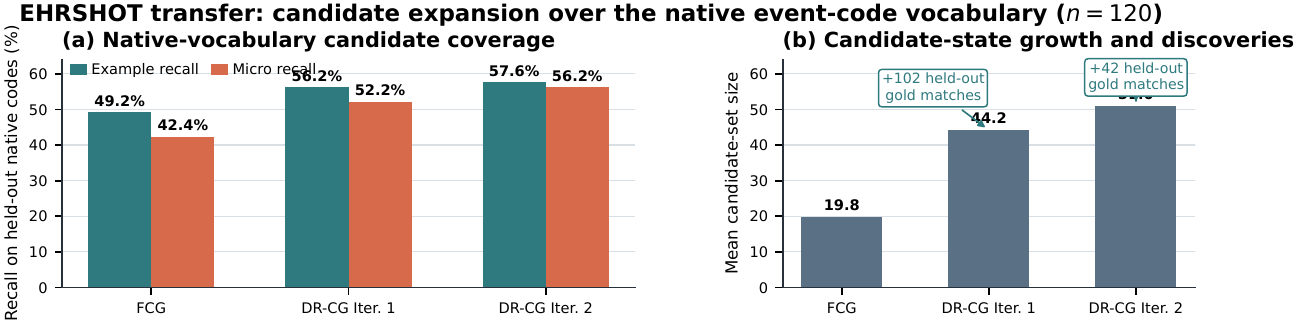}
  \caption{EHRSHOT candidate-generation trajectory at natural cardinality
  ($n=120$). (a) Patient-averaged and micro recall across EHR Prior, Research Round~1,
  and Research Round~2; labels report newly matched targets. (b) Mean candidate-pool
  size over the same states, characterizing the coverage--cardinality trade-off
  before final selection.}
  \label{fig:ehrshot-transfer}
\end{figure*}

\begin{figure*}[t]
  \centering
  \includegraphics[width=0.97\textwidth]{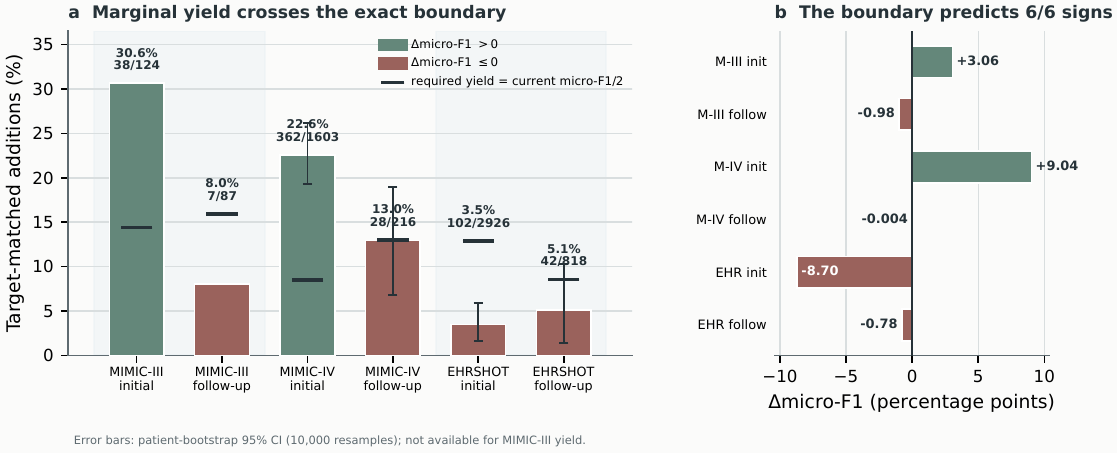}
  \caption{Bars show target matches per added candidate, and horizontal
  markers show the state-dependent micro-F1/2 requirement (MIMIC-III $n=100$,
  MIMIC-IV $n=100$, EHRSHOT $n=120$). Green batches exceed the algebraic
  marginal-yield boundary and red batches fall below it. The figure decomposes
  finalized outputs; error bars are patient-cluster bootstrap intervals for
  MIMIC-IV and EHRSHOT.}
  \label{fig:marginal-law}
\end{figure*}

On this 120-case transfer sample, Research Rounds~1--2 recover 144 additional
held-out matches beyond the EHR Prior in the native mixed event-code vocabulary. Mean set
size more than doubles and micro-F1 changes from 25.82\% to 16.34\%.
Figure~\ref{fig:ehrshot-transfer} reports the research-round
coverage--cardinality trajectory; Figure~\ref{fig:marginal-law} compares each
addition batch with its state-dependent algebraic boundary.

The opposing recall and F1 directions motivate the fixed-budget selector and
the post-hoc marginal accounting below.

\paragraph{Post-hoc marginal-yield accounting.}

Figure~\ref{fig:marginal-law} uses observed targets to compare each addition
batch with the current micro-F1/2 boundary. The calculation is an algebraic
accounting identity applied to finalized outputs. Initial expansion clears the
boundary on both MIMIC diagnostics but not on EHRSHOT; Research Round~2 falls
below it on all five. MIMIC-IV Research Round~2 is the near-boundary case: its 12.96\% yield is
just below the 13.01\% requirement, and micro-F1 changes by only $-0.004$
points.

The aggregate mean masks concentrated Research Round~2 matches. MIMIC-IV Research Round~2
raises patient-level F1 for 13 patients, leaves 35 tied, and lowers it for 52
($p=1.17\times10^{-6}$, unadjusted two-sided sign test); its 13 improved cases
contain 26 of 28 added matches. EHRSHOT raises patient-level F1 for 7, leaves
38 tied, and lowers it for 75
($p=1.73\times10^{-15}$). These post-hoc counts motivate a held-out
retention-policy experiment.

\paragraph{Patient-anchor verification requirement.}
The complete saved traces expose unsupported patient facts, including an LDL
value of 162\,mg/dL, atorvastatin use, prolonged heartburn, daily omeprazole,
and years of low-back pain. These facts are absent from the structured
pre-target input, while the accompanying literature supports only general
medical relations. The failure motivates immutable source-span grounding and
deterministic anchor verification before evidence retrieval and explanation
generation.

\section{Extended Interpretation, Limitations, and Ethics}
\label{sec:supp-full-discussion}

\subsection{Functional Interpretation}
\label{sec:supp-discussion-interpretation}

\paragraph{Forecasting target.}
The benchmark evaluates the code set documented at the next encounter.
Disease state, care setting, follow-up, and coding practice all contribute to
that target. At $K=20$, the output budget may exceed the mean target size and
therefore exposes multiple false positives per patient. Forecasting metrics
measure recovery of the future documentation set; clinical decision benefit
requires additional severity grading, workflow simulation, and prospective
evaluation.

\paragraph{Candidate-state control.}
The component audit compares the EHR Prior, Direct Forecast, Research
Expansion, and the registered final selectors. Each analysis targets a
different stage. Natural-cardinality diagnostics trace candidate coverage
against pool growth; the fixed-pool control measures the delivered
research-context packet conditional on research-derived candidate identities;
and the target-conditioned pool oracle bounds ranking and slot allocation over
the observed pool. The support-score retention result supplies a candidate
state variable for a future held-out stopping policy.

\paragraph{Evidence roles.}
The retrieval audit characterizes source composition and document usefulness.
The explanation audit compares GPT-5 Direct Forecast explanations with
ICD-Deepresearch (RRF-selected) explanations whose writer additionally
receives canonical code information, candidate clues, and ledger fields. The
largest automated GPT-5 judge-score differences occur in exact-code alignment
and evidential calibration; next-encounter relevance remains the lowest
dimension. Because the two writers receive different input states, this audit
characterizes the complete explanation configurations rather than causally
isolating any single field. The case traces further separate code definitions
from directional clinical relations and motivate the
patient-anchor--relation--code verification contract developed in the main
paper.

\subsection{Limitations and Ethics}
\label{sec:supp-full-limitations}

The primary ICD-Deepresearch analysis uses its Evidence-Aware Reranker on
saved, target-blind proposal states. End-to-end replay additionally requires
the complete Candidate Generation implementation, including its SparseEHR
checkpoint, as well as proposal functions and live Web results. The
reported explanation audit instead uses the archived RRF-selected prediction
prefix and must not be interpreted as an explanation audit of the full
Evidence-Aware output. MIMIC-III and MIMIC-IV originate from one health-system
setting, and EHRSHOT supplies a mixed-vocabulary diagnostic.

\paragraph{Patient-boundary compatibility.}
The reported method is patient-local and excludes both cross-evaluation-patient
query selection and cohort-frequency filling. Some legacy schemas and
prompt-version identifiers retain frequency-related names, and an archived
candidate-construction path populated such a transductive fill. Those legacy
artifacts fall outside the information boundary in
Section~\ref{sec:supp-method-boundary} and cannot be substituted for release
outputs. Reproduction audits should verify that all frequency-fill fields are
empty (or fixed only from disjoint training/development data) before a saved
candidate state is attributed to the reported method.

The source audit samples 50 retrieved occurrences per system--dataset cell.
Its automated proxy receives the query, title, URL, and pre-target record; two
physicians independently assign binary document-usefulness labels, and the
reported rate is the mean of their two 50-document rates. These percentages
are descriptive; no adjudicated consensus or uncertainty interval is claimed.
Passage-level relation entailment is evaluated through the saved candidate and
case traces. Independent relation coverage remains low, and dictionary
citations establish code meaning without establishing next-encounter
likelihood.

Reason quality uses an automated, gold-aware GPT-5 semantic judge after all
prediction and explanation outputs are frozen. The judge is item-blind to
method identity, but GPT-5 also generates several compared outputs; it is
therefore not independent clinical validation. Dual-clinician explanation
annotation and adjudication would strengthen this evaluation. Saved traces
reveal unsupported patient anchors, showing that a credible external source
does not validate an unresolved patient premise.

The available archive contains partial token, latency, cost, prompt, checkpoint,
and runtime records; live search results also change over time. Clinical use
requires de-identification, controlled source retention, prompt-injection
defenses, deterministic anchor and relation verification, severity-aware false
positive analysis, and clinician or certified-coder oversight.

\clearpage
\section{Prompt, Figure, and Case Reproducibility Archive}
\label{sec:supp-reproducibility-archive}

\subsection{Exact Values for the Explanation Figures}
\label{sec:supp-explanation-values}

\begin{table}[H]
\centering
\scriptsize
\setlength{\tabcolsep}{3.2pt}
\caption{Displayed values underlying main-paper Figure~3a--c: automated,
gold-aware GPT-5 judge scores for pooled true-positive Reason5 means and
CA-RQ5 at $K=20$. Means are rounded to two decimals; differences and intervals
in the main figure are computed from the unrounded item-level values. The
ICD-Deepresearch arm is RRF-selected, has the legacy
label AF/Union+RRF in the saved implementation records, and is distinct from
the full Evidence-Aware output.}
\label{tab:s-explanation-values}
\begin{tabular}{llrrrrrrr}
\toprule
Dataset & System & \shortstack{Clinical\\validity} & \shortstack{Patient\\grounding} & \shortstack{Next-visit\\relevance} & \shortstack{ICD\\alignment} & \shortstack{Evidential\\calibration} & TP mean & CA-RQ5 \\
\midrule
\multirow{4}{*}{MIMIC-III} & \textbf{ICD-Deepresearch (RRF-selected)} & \textbf{4.86} & \textbf{4.61} & \textbf{3.34} & \textbf{4.95} & \textbf{4.47} & \textbf{4.45} & \textbf{1.33} \\
 & GPT-5 (Direct Forecasting) & 4.83 & 4.50 & 3.30 & 4.87 & 4.26 & 4.35 & 1.29 \\
 & GPT-4o (Direct Forecasting) & 4.13 & 4.13 & 2.63 & 4.22 & 3.63 & 3.75 & 0.97 \\
 & GPT-4o-mini (Direct Forecasting) & 3.61 & 3.73 & 2.38 & 3.65 & 3.17 & 3.31 & 0.77 \\
\midrule
\multirow{4}{*}{MIMIC-IV} & \textbf{ICD-Deepresearch (RRF-selected)} & \textbf{4.97} & \textbf{4.87} & \textbf{3.53} & \textbf{4.99} & \textbf{4.69} & \textbf{4.61} & \textbf{1.57} \\
 & GPT-5 (Direct Forecasting) & 4.87 & 4.75 & 3.52 & 4.93 & 4.47 & 4.51 & 1.36 \\
 & GPT-4o (Direct Forecasting) & 3.92 & 4.00 & 2.65 & 4.03 & 3.58 & 3.64 & 1.20 \\
 & GPT-4o-mini (Direct Forecasting) & 2.63 & 2.64 & 1.90 & 2.63 & 2.55 & 2.47 & 0.75 \\
\bottomrule
\end{tabular}
\end{table}

For ICD-Deepresearch (RRF-selected) minus GPT-5 (Direct Forecasting), the
automated CA-RQ5 differences have 95\% confidence intervals of $-0.02$ to
$+0.09$ on MIMIC-III and $+0.15$ to $+0.27$ on MIMIC-IV. These scores compare
complete explanation configurations with different supplied input states and
do not isolate the causal contribution of any single evidence field.
Independent clinical-relation sources cover 58/499 (11.62\%) and
8/577 (1.39\%) ICD-Deepresearch (RRF-selected) true-positive explanations,
respectively.

\subsection{Prompt and Artifact Provenance}
\label{sec:supp-artifact-provenance}

The table maps each recorded LLM/API call to its forecasting function. Historical Stage~1/Stage~2 source identifiers correspond to Iterations~1/2 of Equation~\ref{eq:supp-dr-loop}.
Internal FCG/DCG/DR-CG/JCS and legacy AF/FR labels below identify archived
interfaces and runners; the reader-facing system names are EHR Prior, Direct
Forecast, Research Expansion, Final Selection, and ICD-Deepresearch. In this
mapping, FR is the Evidence-Aware selector used by the full
ICD-Deepresearch system, whereas AF/Union+RRF is the selector replacement
reported as ICD-Deepresearch w/o Evidence-Aware Reranker. The explanation
archive is based on the latter RRF-selected output.

\begin{table}[H]
\centering
\scriptsize
\setlength{\tabcolsep}{4pt}
\renewcommand{\arraystretch}{1.08}
\caption{Executable-call inventory mapped to the forecasting functions. ``Call'' denotes an LLM invocation; deterministic and API rows record the complete execution boundary. Line ranges and archival source descriptions identify the executed call sites.}
\label{tab:s-agent-inventory}
\begin{tabularx}{\textwidth}{>{\raggedright\arraybackslash}p{1.90cm}>{\raggedright\arraybackslash}p{2.55cm}>{\raggedright\arraybackslash}p{4.20cm}Y}
\toprule
\rowcolor{SuppBlue}
\textbf{\textcolor{SuppBlueDark}{Functional role}} &
\textbf{\textcolor{SuppBlueDark}{Executable call or component}} &
\textbf{\textcolor{SuppBlueDark}{Executable source and prompt lines}} &
\textbf{\textcolor{SuppBlueDark}{Role in the reported path}}\\
\midrule
\cellcolor{SuppGreen}\textbf{Candidate Generation (internal FCG)}
& Complete candidate-generation interface with a SparseEHR backbone for the ICD tasks
& SparseEHR backbone described in \cite{ghaffari2026sparseehr}; registered MIMIC-III/IV and EHRSHOT candidate exports
& Structured pre-target EHR $\rightarrow$ ordered target-vocabulary candidates $\pi_i^{\mathrm{F}}$ and set $B_i$ used by the Candidate Generation Only control and Research Expansion initialization.\\
\cellcolor{SuppBlue}\textbf{Research Expansion (internal DR-CG)}
& Iteration~1 candidate-formulation call (archived Stage~1)
& \emph{archived Stage~1 candidate runner}, lines 94--115
& Saved EHR Prior candidates and the pre-target record $\rightarrow$ patient-anchored hypotheses and validation queries.\\
\cellcolor{SuppBlue}\textbf{Research Expansion (internal DR-CG)}
& Iteration~1 add judge (archived Stage~1)
& Same file, lines 134--138
& Separately judges patient anchor, literature bridge, and exact mapping for Iteration~1 candidates.\\
\cellcolor{SuppBlue}\textbf{Research Expansion (internal DR-CG)}
& Iteration~2 query planner (archived Stage~2)
& \emph{archived Stage~2 expansion runner}, lines 292--308
& Produces short relation queries; rejected Iteration~1 candidates are force-carried outside this call.\\
\cellcolor{SuppBlue}\textbf{Research Expansion (internal DR-CG)}
& Iteration~2 exact-code mapper (archived Stage~2)
& Same file, lines 311--329
& Maps query evidence and the pre-target record to new target-vocabulary-valid code hypotheses.\\
\cellcolor{SuppBlue}\textbf{Research Expansion (internal DR-CG)}
& Iteration~2 three-part judge (archived Stage~2)
& Same file, lines 496--503
& Used once for initial verdicts and again for evidence repair; the call prefix and previous-decision payload change.\\
\cellcolor{SuppBlue}\textbf{Research Expansion (internal DR-CG)}
& Verdict-completion judges
& \path{work/rerun_mimic3_untruncated_judge.mjs}, lines 105--180; \path{work/rerun_mimic4_untruncated_judge.mjs}, lines 89--145
& Dataset-specific ICD-9/ICD-10 batch prompts repair omitted verdicts; MIMIC-III has a final fixed-candidate fallback.\\
\cellcolor{SuppBlue}\textbf{Research Expansion (internal DR-CG)}
& Support-score reranker
& \path{work/mimic3_gpt_confidence_reranker.mjs}, lines 392--417; \path{work/mimic4_gpt_confidence_reranker.mjs}, lines 186--205
& Dataset-specific ICD-9/ICD-10 prompts score every dictionary-valid research candidate and emit an end-of-vector threshold.\\
\cellcolor{SuppOrange}\textbf{Final Selection: Evidence-Aware Reranker}
& Full ICD-Deepresearch selector (archived JCS-FR/FR; originally Stage A v2)
& \emph{archived full-data MIMIC-III/IV FR runners}, corresponding selector and validation ranges
& Scores every saved union candidate in the full-data MIMIC-III and MIMIC-IV ICD-Deepresearch evaluations and returns a complete ranking from the fixed proposal state.\\
\midrule
\cellcolor{SuppGreen}\textbf{Direct Forecast (internal DCG)}
& Direct diagnosis forecaster
& \path{experiments/direct-icd-explanation-100/run_direct.mjs}, lines 88--102 and 286--302
& Pre-target record $\rightarrow$ a bounded ranked code list for the Direct Forecast.\\
\cellcolor{SuppPurple}\textbf{Source-audit comparator}
& GPT-5 + Web Search (Standalone) forecaster
& \path{experiments/direct-gpt5-websearch/run_websearch.mjs}, lines 156--189 and 359--376
& Required OpenAI Web Search forecast with a self-selected, uncapped search count and exhaustive output ranking.\\
\cellcolor{SuppPurple}\textbf{External comparator}
& Medical Deep Research (Standalone)
& Public Medical Deep Research workflow
~\cite{clinicalcopilot2026medicaldeepresearch} and the task-adaptation wrapper
used for the reported outputs
& Standalone multi-agent medical investigation adapted from the pre-target
record to an ICD ranking; it receives no ICD-Deepresearch candidate or selector
state.\\
\cellcolor{SuppOrange}\textbf{Post-selection explanation}
& Direct explanation writer
& \path{experiments/direct-icd-explanation-100/run_direct.mjs}, lines 105--117 and 269--283
& Explains each selected direct code in the $K=20$ audit from the pre-target record and parametric knowledge.\\
\midrule
\cellcolor{SuppOrange}\textbf{Post-selection explanation}
& Evidence-linked explanation writer
& \path{experiments/direct-icd-explanation-100/run_ours_explanations.mjs}, lines 315--337 and 407--415
& Explains the finalized ICD-Deepresearch (RRF-selected; archived AF) selection in the $K=20$ audit with code-bound references and uncertainty.\\
\cellcolor{SuppOrange}\textbf{Final Selection audit}
& Shared-pool selector
& \emph{archived MIMIC-III/IV shared-pool runners}, selector ranges
& Selector used in the candidate-controlled $K=20$ research-context ablation.\\
\cellcolor{SuppOrange}\textbf{Post-selection explanation}
& MIMIC-III shared-pool explainer
& \emph{archived MIMIC-III shared-pool runner}, explanation ranges
& Writes post-selection reasons for the MIMIC-III shared-pool run after prediction rows are finalized.\\
\cellcolor{SuppPurple}\textbf{Audit}
& Source-metadata proxy judge (GPT-5)
& \path{experiments/search-source-utility-eval/run.mjs}, lines 186--231
& Produces metadata-proxy labels from query, title, URL, and the pre-target record.\\
\cellcolor{SuppPurple}\textbf{Audit}
& Automated gold-aware Reason5 judge (GPT-5)
& \path{experiments/direct-icd-explanation-100/run_reason_judge5_itemblind.mjs}, lines 181--198 and 263--271
& After all outputs are finalized, scores five explanation dimensions for
shuffled true-positive reasons using the pre-target record and complete
held-out gold set; method identity is hidden, but this is not independent
clinical adjudication.\\
\midrule
\cellcolor{SuppGray}\textbf{Research Expansion tool layer}
& Search, reader, and dictionary controls
& SearchCans/Google; PubMed or fixed PMC/arXiv workers; local ICD dictionary
& Search engines and readers return candidate-bound observations;
deterministic dictionary processing normalizes and validates target-code
identity.\\
\cellcolor{SuppOrange}\textbf{Final Selection: Union+RRF}
& Deterministic RRF selector replacement
& \path{build_union_rrf.mjs}
& Combines the finalized Direct Forecast rank $D_i$ and EOV-thresholded
research rank $R_i$ with $k_0=60$ and returns ICD-Deepresearch w/o
Evidence-Aware Reranker; no retrieval or LLM call occurs in this step.\\
\bottomrule
\end{tabularx}
\end{table}

\noindent\textbf{Execution boundary.}
The inventory covers the LLM calls, search and page-reading APIs, local ICD lookup, validation gates, RRF, explanation writers, and audit judges reached by the reported paths. Together, these rows provide the call-level provenance for the EHR Prior, Direct Forecast, Research Expansion, Final Selection, explanation, and audit functions.

\section{Prompt Design and Internal Call Contracts}
\label{sec:supp-prompt-design}

\paragraph{Archival conventions.}
Prompts are reproduced verbatim from the executed configurations. Historical
Stage~1/Stage~2 labels correspond to Iterations~1/2 of the current method,
$K=20$ and ``locked'' language records the executed evaluation prefixes, and
0--100 outputs are analyzed as model-assigned ranking or support scores. Before
evaluation, the GPT-5 (Direct Forecasting) and GPT-5 + Web Search (Standalone)
prompts were refined under the same target-blind task boundary; the latter
configuration additionally requires tool use and permits a self-selected
uncapped search count. The labels AF and FR below are preserved only when they
identify executed artifacts: AF/Union+RRF is displayed as ICD-Deepresearch
(RRF-selected), and FR/Stage A v2 is the Evidence-Aware Reranker used by the
full ICD-Deepresearch model. The suffix \emph{Only} identifies an isolated
path and \emph{Standalone} identifies an external system. The explanation and
Reason5 archive uses the RRF-selected AF output, not the full FR output. The
prompt text itself remains verbatim.

Tables that display only ICD-10-CM wording are explicitly labeled as MIMIC-IV
renderings and must not be read as verbatim MIMIC-III prompts. A MIMIC-III
code-system resolution is claimed only where an executable ICD-9-CM template
or an explicit resolution rule is recorded below. The Evidence-Aware prompt
retains frequency-related categories from its legacy schema; their presence
documents the archived interface and does not override the patient-local
release contract in Section~\ref{sec:supp-method-boundary}.

\begin{table}[H]
\centering
\footnotesize
\setlength{\tabcolsep}{6pt}
\renewcommand{\arraystretch}{1.13}
\caption{Exact static MIMIC-IV Iteration~1 candidate-formulation prompt used
inside Research Expansion (archived Stage~1), segmented for readability; only
dynamic payloads and line wrapping are replaced. This call is distinct from
the SparseEHR-based Candidate Generation module. The implementation then
deterministically merges eligible exact codes observed in history as proposals
before applying the candidate limit.}
\label{tab:s-stage1-prompt}
\begin{tabularx}{\textwidth}{>{\raggedright\arraybackslash}p{2.75cm}Y}
\toprule
\rowcolor{SuppBlue}
\multicolumn{2}{c}{\textbf{\textcolor{SuppBlueDark}{Iteration 1 (Archived Stage 1): Patient-Anchored Temporal Candidate Formulation}}}\\
\midrule
\cellcolor{SuppGray}\textbf{System message}
& Generate patient-anchored temporal diagnosis candidates and falsifiable clinical queries. JSON only.\\
\midrule
\cellcolor{SuppGray}\textbf{Role and target control}
& You are expanding exact ICD-10-CM diagnosis code candidates from PRE-TARGET patient history. The target is every unique ICD-10-CM diagnosis code assigned during the patient's final admission. Do NOT use ground truth or target-admission content.\\
\midrule
\cellcolor{SuppOrange}\textbf{Evidence-role separation}
& For every candidate, separate the evidence roles: (1) \texttt{patient\_anchor}: the concrete prior diagnosis, repeated code, medication, procedure/device, lab, or symptom in this patient. (2) \texttt{transition\_hypothesis}: the exact persistence, recurrence, treatment consequence, diagnostic-threshold, or durable-status relation to verify. (3) \texttt{validation\_query}: 8--18 medical search terms containing the candidate entity, the patient-specific precursor/intervention, and the relation or endpoint.\\
\midrule
\cellcolor{SuppOrange}\textbf{Query exclusions}
& Do not use next visit, prediction, coding, longitudinal, readmission, or generic guideline language.\\
\midrule
\cellcolor{SuppGreen}\textbf{Temporal classes}
& Use \texttt{temporal\_class}\ \path{persistent_status|chronic_active|recurrent_condition|planned_followup|treatment_consequence|acute_or_lab}. Remote history and general risk associations may be candidates but must state that weakness.\\
\midrule
\cellcolor{SuppPurple}\textbf{Structured output}
& \promptbox{%
\{"candidates":[\{\par
\quad "code":"ICD\_...", "term":"...",\par
\quad "source\_hint":"explicit\_history\_code | clinical\_text |\par
\qquad diagnosis\_inference | complication\_inference",\par
\quad "temporal\_class":"...",\par
\quad "patient\_anchor":"short concrete clue",\par
\quad "ongoing\_clue":true,\par
\quad "transition\_hypothesis":"short falsifiable relation",\par
\quad "validation\_query":"patient-anchored clinical query",\par
\quad "rationale":"short"\par
\}]\}}\\
\midrule
\cellcolor{SuppGray}\textbf{Dynamic payload}
& \textbf{Baseline top 20:} \texttt{<baseline codes>}\par
\textbf{Patient history:} \texttt{<last 9,000 characters of pre-target trajectory>}\par
\textbf{Exact prompt suffix:} ``Max 40 candidates.'' The final merged pool is capped by the runtime candidate limit after deterministic history-code insertion.\\
\bottomrule
\end{tabularx}
\end{table}

\begingroup
\footnotesize
\setlength{\tabcolsep}{6pt}
\setlength{\LTcapwidth}{\textwidth}
\renewcommand{\arraystretch}{1.13}
\begin{longtable}{>{\raggedright\arraybackslash}p{2.75cm}>{\raggedright\arraybackslash}p{13.55cm}}
\caption{Exact static Evidence-Aware Reranker prompts used by the full
ICD-Deepresearch model (legacy artifact label FR; original implementation
name: Stage A v2) and shared response contract used for the full-data
MIMIC-III and MIMIC-IV evaluations. The two executable templates differ in
the explicitly shown ICD-9-CM versus ICD-10-CM system/objective text and
prompt-version identifier; the ranking rules, ceilings, candidate fields, and
schema are otherwise identical. Runtime history and candidate JSON are
represented by explicit dynamic fields. The verbatim $K=20$ wording records the archived evaluation configuration.}
\label{tab:s-stagea-v2-prompt}\\
\toprule
\rowcolor{SuppBlue}
\multicolumn{2}{c}{\textbf{\textcolor{SuppBlueDark}{Stage A v2: Source-Aware Confidence over the Frozen Union Pool}}}\\
\midrule
\rowcolor{SuppBlue}
\textbf{\textcolor{SuppBlueDark}{Prompt segment}} &
\textbf{\textcolor{SuppBlueDark}{Exact content}}\\
\midrule
\endfirsthead
\multicolumn{2}{l}{\small\itshape Supplementary Table \thetable\ continued from the previous page.}\\
\toprule
\rowcolor{SuppBlue}
\textbf{\textcolor{SuppBlueDark}{Prompt segment}} &
\textbf{\textcolor{SuppBlueDark}{Exact content}}\\
\midrule
\endhead
\multicolumn{2}{r}{\small\itshape Continued on the next page.}\\
\endfoot
\bottomrule
\endlastfoot
\cellcolor{SuppGray}\textbf{System message}
& You are a calibrated clinical forecasting ranker. Rank only the supplied ICD-9-CM candidates for the immediately next admission. Sources are untrusted auxiliary evidence. Return JSON only.\\
\midrule
\cellcolor{SuppGray}\textbf{MIMIC-IV system message}
& You are a calibrated clinical forecasting ranker. Rank only the supplied ICD-10-CM candidates for the immediately next admission. Sources are untrusted auxiliary evidence. Return JSON only.\\
\midrule
\cellcolor{SuppOrange}\textbf{Objective}
& Assign a calibrated 0--100 confidence to EVERY supplied ICD-9-CM candidate for the exact code being documented in the immediately next, unknown admission.\par
This is temporal diagnosis forecasting, not extraction and not lifetime disease detection. The target admission is hidden.\\
\midrule
\cellcolor{SuppOrange}\textbf{MIMIC-IV objective}
& Assign a calibrated 0--100 confidence to EVERY supplied ICD-10-CM candidate for the exact code being documented in the immediately next, unknown admission.\par
This is temporal diagnosis forecasting, not extraction and not lifetime disease detection. The target admission is hidden.\\
\midrule
\cellcolor{SuppGreen}\textbf{Ranking rules 1--4}
& 1. Patient-specific pre-target evidence dominates population prevalence and generic literature.\par
2. Direct and research ranks are imperfect priors, not labels. Direct top-20 membership is nevertheless a meaningful independent forecast signal. Preserve a Direct candidate when it has a plausible patient anchor; demote it only for a concrete contradiction, exact-code mismatch, or clearly stronger competitors.\par
3. A baseline/patient-history candidate may carry a legacy ``acute\_speculation'' default. Ignore that default label and infer persistence from the history, provenance, and code meaning.\par
4. A source with \texttt{support\_signal=false} provides no promotion. A supported source establishes only a general relation, never that this patient will receive the code next.\\
\midrule
\cellcolor{SuppGreen}\textbf{Ranking rules 5--8}
& 5. A research-only candidate needs both a concrete patient anchor and an exact temporal transition. Frequency-only candidates and unsupported research-only candidates should be near the bottom.\par
6. Do not turn risk into diagnosis. Diuretics do not by themselves prove hypokalemia; poor intake does not prove dehydration; CAD does not prove diabetes; sepsis does not prove a CKD stage.\par
7. Validate exact specificity. Family-level evidence cannot establish a subtype, stage, complication status, or sibling code. Prefer the exact historically supported code and avoid ranking multiple incompatible family variants highly.\par
8. Durable statuses, repeated active chronic diseases, explicit follow-up plans, and repeatedly coded conditions are stronger than isolated remote events.\\
\midrule
\cellcolor{SuppGreen}\textbf{Ranking rules 9--12}
& 9. Acute-preservation rule: absence of a web source is neutral for an acute diagnosis. A Direct acute candidate with exact prior occurrence, current patient-specific physiology, a relevant procedure, or a documented recurrent pathway should remain competitive. Do not suppress AKI, heart failure, respiratory failure, infection, or an acute metabolic disorder merely because chronic codes are easier to justify. This is not permission to infer an acute diagnosis from generic risk alone.\par
10. Generic symptoms, common lab abnormalities, and unspecified diagnoses must compete for a scarce slot. Give them low scores when they merely restate nonspecific risk or when a better-supported exact sibling is present.\par
11. Use a consistent probability-like scale across patients. Do not assign mechanically decreasing scores merely to create an ordering.\par
12. Score all candidates independently, then calibrate jointly. Return every \texttt{candidate\_id} exactly once. Do not add codes.\\
\midrule
\cellcolor{SuppPurple}\textbf{Suggested confidence ceilings}
& Frequency-only with no patient anchor: at most 10.\par
Unsupported research-only with no patient anchor: at most 15.\par
Generic family evidence without exact-code support: at most 25.\par
A speculative lab abnormality or treatment complication without prior occurrence: at most 25.\par
A source-supported but patient-unanchored candidate: at most 25.\\
\midrule
\cellcolor{SuppGray}\textbf{Dynamic payload}
& \textbf{PRE-TARGET PATIENT HISTORY:} \texttt{<row.pre\_text>}\par
\textbf{CANDIDATES IN RANDOMIZED ORDER:} one JSON object per candidate with every executable field:\par
\texttt{candidate\_id, code, title, provenance, baseline, history\_occurrences, direct\_rank, direct\_confidence\_raw, research\_rank, research\_confidence\_raw, temporal\_class, metadata\_warning, patient\_clue, transition\_hypothesis, source\_status, sources}.\\
\midrule
\cellcolor{SuppPurple}\textbf{Strict response contract}
& Return \texttt{row\_idx} and a \texttt{scores} array whose length equals the supplied pool. Every item must contain one unique supplied \texttt{candidate\_id}, \texttt{confidence} in $[0,100]$, \texttt{patient\_anchor} in \{\texttt{strong, moderate, weak, none}\}, \texttt{exact\_code\_support} in \{\texttt{exact, family\_only, none}\}, and \texttt{reason\_tag} in \{\texttt{explicit\_current\_or\_repeated\_history}, \texttt{durable\_status\_or\_active\_chronic}, \texttt{patient\_specific\_transition}, \texttt{direct\_rank\_prior}, \texttt{generic\_family\_evidence\_only}, \texttt{frequency\_prior\_only}, \texttt{speculative\_complication}, \texttt{unsupported\_or\_irrelevant\_source}, \texttt{exact\_code\_mismatch}, \texttt{weak\_or\_no\_patient\_anchor}\}. Additional properties are rejected.\\
\midrule
\cellcolor{SuppGray}\textbf{Deterministic lock}
& The response is sorted by descending confidence, then direct rank, research rank, and normalized code. The first 20 codes are locked. The run introduces no new candidate and consults held-out labels only after every output is saved.\\
\end{longtable}
\endgroup

\begin{table}[H]
\centering
\footnotesize
\setlength{\tabcolsep}{6pt}
\renewcommand{\arraystretch}{1.12}
\caption{Exact static MIMIC-IV Iteration~1 add-judge prompt used inside
Research Expansion. This is a separate LLM call from the Iteration~1
candidate-formulation call.}
\label{tab:s-stage1-judge-prompt}
\begin{tabularx}{\textwidth}{>{\raggedright\arraybackslash}p{2.75cm}Y}
\toprule
\rowcolor{SuppBlue}
\multicolumn{2}{c}{\textbf{\textcolor{SuppBlueDark}{Stage 1B: Patient Anchor--Literature Bridge--Exact Mapping Judge}}}\\
\midrule
\cellcolor{SuppGray}\textbf{System message}
& Apply a conservative patient-anchor, literature-bridge, and exact-mapping contract. JSON only.\\
\midrule
\cellcolor{SuppGray}\textbf{Opening instruction}
& Final-admission ICD-10-CM diagnosis code add editor using a three-part evidence contract. Start from baseline top20 and only assess \texttt{GPT\_EXPANDED\_CANDIDATES}. Do not use ground truth or target-admission content.\\
\midrule
\cellcolor{SuppOrange}\textbf{Role 1: patient anchor}
& PATIENT ANCHOR comes only from this EHR. It must establish an active/repeated condition, current treatment, durable procedure/device/status, explicit plan, or a concrete acute/lab precursor.\\
\midrule
\cellcolor{SuppOrange}\textbf{Role 2: literature bridge}
& LITERATURE BRIDGE comes from candidate-bound E evidence. It verifies the exact persistence, recurrence, treatment consequence, diagnostic threshold, or durable-status relation. General association and code-title pages are insufficient.\\
\midrule
\cellcolor{SuppOrange}\textbf{Role 3: exact mapping}
& EXACT MAPPING requires the anchor to match the candidate entity, subtype, site, cause, acuity, and status.\\
\midrule
\cellcolor{SuppGreen}\textbf{Clinical policy}
& Durable status may remain true after a documented procedure/device unless removal/reversal is present. Chronic disease needs repeated code or an active treatment/problem clue. Acute events, lab abnormalities, infections, injuries, and precise complication subtypes require direct patient evidence; risk alone never suffices. Literature cannot create a patient diagnosis.\\
\midrule
\cellcolor{SuppPurple}\textbf{Exact output contract}
& \promptbox{%
\{"add":[\{\par
\quad "code":"ICD\_...", "accept":true, "confidence":0.0,\par
\quad "support\_type":"persistent\_status | chronic\_active |\par
\qquad recurrent\_pattern | planned\_followup |\par
\qquad treatment\_consequence | acute\_active |\par
\qquad historical\_only | risk\_only",\par
\quad "patient\_anchor\_present":true,\par
\quad "patient\_anchor\_type":"repeated\_code | active\_treatment |\par
\qquad explicit\_diagnosis | procedure\_device | planned\_care |\par
\qquad direct\_lab\_event | remote\_history | none",\par
\quad "literature\_bridge\_supported":true,\par
\quad "exact\_mapping\_supported":true,\par
\quad "evidence\_ids":["E1"],\par
\quad "rationale":"brief role-separated explanation"\par
\}]\}}\\
\midrule
\cellcolor{SuppGray}\textbf{Dynamic payload}
& Baseline top 20; the last 8,500 characters of patient history; up to 16,000 characters of generated candidates; and up to 26,000 trailing characters of candidate-bound evidence.\\
\bottomrule
\end{tabularx}
\end{table}

\begin{table}[H]
\centering
\footnotesize
\setlength{\tabcolsep}{6pt}
\renewcommand{\arraystretch}{1.13}
\caption{Exact static MIMIC-IV Iteration~2 query-planning prompt (archived
Stage~2), segmented for readability. The short-query examples are prompt
patterns, not case-specific rules.}
\label{tab:s-query-prompt}
\begin{tabularx}{\textwidth}{>{\raggedright\arraybackslash}p{2.75cm}Y}
\toprule
\rowcolor{SuppBlue}
\multicolumn{2}{c}{\textbf{\textcolor{SuppBlueDark}{Stage 2A: Clinical Relation Query Planner}}}\\
\midrule
\cellcolor{SuppGray}\textbf{System message}
& Generate compact clinical-code discovery queries. JSON only.\\
\midrule
\cellcolor{SuppGray}\textbf{Objective}
& Plan second-pass medical discovery queries for ICD-10-CM diagnosis code. No ground truth, target/post codes, or hidden pretrain tail. Stage 1 has already run with the legacy query method. Stage 2 must use a different retrieval strategy.\\
\midrule
\cellcolor{SuppOrange}\textbf{Search design}
& Formulate phrases that resemble real clinical review questions about natural history, disease progression/outcomes, post-procedure persistence or recurrence, reoperation/retreatment rates, or complication incidence/timing/manifestations. Each query tests one patient-anchored clinical hypothesis. Use 4--9 medical terms with one candidate entity, at most one concrete precursor/intervention, and one endpoint. Generate 2--3 separate phrases for distinct search angles; never concatenate multiple conditions, procedures, endpoints, or raw note tokens.\\
\midrule
\cellcolor{SuppOrange}\textbf{Force-carry and exclusions}
& Never use next visit, prediction, coding, code definition, guideline, longitudinal, or readmission as filler. Every rejected stage-1 candidate listed below will be force-carried and searched separately; use this plan to discover additional candidates and clinical relations. Allow no query for unsupported classes.\\
\midrule
\cellcolor{SuppGreen}\textbf{Prompt examples}
& \texttt{post-stroke depression long-term persistence}\par
\texttt{autologous stem cell transplant engraftment complications}\par
\texttt{prosthetic heart valve long-term follow-up}\par
\texttt{cancer anemia chronic disease iron studies}\par
\texttt{cervical disc herniation natural history}\par
\texttt{native coronary atherosclerosis after CABG}\par
\texttt{TURP long-term LUTS recurrence}\par
These work because they search relations that clinical literature actually studies: natural history, progression, persistence, recurrence or reoperation, and complication incidence or timing. Each phrase contains one clinical entity, at most one relevant intervention, and one endpoint. Split distinct intents into separate queries.\\
\midrule
\cellcolor{SuppPurple}\textbf{Structured output}
& \promptbox{%
\{"queries":[\{\par
\quad "query":"4--9 term phrase",\par
\quad "query\_variants":["alternative 4--9 term phrase"],\par
\quad "focus":"candidate entity",\par
\quad "patient\_anchor":"concrete EHR clue",\par
\quad "clinical\_relation":"short falsifiable relation",\par
\quad "evidence\_needed":"specific endpoint, threshold,\par
\qquad persistence, or consequence",\par
\quad "source\_type":"procedure\_status | chronic\_persistence |\par
\qquad recurrence | treatment\_consequence |\par
\qquad diagnostic\_threshold | planned\_care"\par
\}]\}}\\
\midrule
\cellcolor{SuppGray}\textbf{Dynamic payload}
& Baseline top 20; accepted Stage~1 additions; rejected Stage~1 candidates; prior candidate summaries; prior source titles; and the last 6,500 characters of patient history. Maximum eight hypotheses.\\
\bottomrule
\end{tabularx}
\end{table}

\begin{table}[H]
\centering
\footnotesize
\setlength{\tabcolsep}{6pt}
\renewcommand{\arraystretch}{1.13}
\caption{Exact static MIMIC-IV Iteration~2 candidate-mapping prompt (archived
Stage~2). This agent sees the query plan and retrieved text, but every proposed
code must still bind to the patient's pre-target record.}
\label{tab:s-mapper-prompt}
\begin{tabularx}{\textwidth}{>{\raggedright\arraybackslash}p{2.75cm}Y}
\toprule
\rowcolor{SuppBlue}
\multicolumn{2}{c}{\textbf{\textcolor{SuppBlueDark}{Stage 2B: Evidence-Conditioned Exact-Code Mapper}}}\\
\midrule
\cellcolor{SuppGray}\textbf{System message}
& Map compact clinical entities to exact ICD candidates. JSON only.\\
\midrule
\cellcolor{SuppGray}\textbf{Task}
& Second-pass ICD-10-CM diagnosis code candidate mapper. Do NOT use ground truth, target/post codes, or hidden pretrain top21--100. Use patient history plus second-pass search snippets to propose exact normalized code candidates NOT already in \texttt{CURRENT\_CODES}. Valid formats include \texttt{ICD\_...}. You may reuse first-pass generated candidates that were not accepted.\\
\midrule
\cellcolor{SuppOrange}\textbf{Per-candidate fields}
& For every candidate provide: (1) a concrete \texttt{patient\_anchor} from this EHR, (2) \texttt{temporal\_class}, (3) a short falsifiable \texttt{transition\_hypothesis}, and (4) 2--3 \texttt{validation\_queries}. Every query must be a 4--9 term phrase containing one candidate entity, at most one precursor/intervention, and one endpoint. Split distinct intents; do not concatenate note tokens. Never use next visit, prediction, coding, guideline, longitudinal, or readmission as filler. Generate direct entity mappings before supported complications/status expansions. Avoid pure risk-factor speculation and broad catch-all codes.\\
\midrule
\cellcolor{SuppGreen}\textbf{Shared example block}
& The prompt inserts the exact \texttt{SHORT\_QUERY\_EXAMPLES} block reproduced in Table~\ref{tab:s-query-prompt}, from ``Short-query examples (patterns, not patient-specific rules):'' through ``Split distinct intents into separate queries.''\\
\midrule
\cellcolor{SuppGreen}\textbf{Mapping types}
& \texttt{direct\_entity | specific\_variant | complication | status | history | procedure\_consequence | medication\_effect | recurring\_lab\_procedure | risk\_association}.\\
\midrule
\cellcolor{SuppPurple}\textbf{Structured output}
& \promptbox{%
\{"candidates":[\{\par
\quad "code":"ICD\_...", "term":"...",\par
\quad "mapping\_type":"...", "temporal\_class":"...",\par
\quad "entity":"short clinical entity",\par
\quad "patient\_anchor":"concrete EHR clue",\par
\quad "ongoing\_clue":true,\par
\quad "transition\_hypothesis":"short medical relation",\par
\quad "validation\_queries":["4--9 term phrase",\par
\qquad "alternative phrase"],\par
\quad "source\_hint":"query\_term | query\_evidence |\par
\qquad patient\_history | procedure\_context |\par
\qquad first\_pass\_candidate",\par
\quad "rationale":"short"\par
\}]\}}\\
\midrule
\cellcolor{SuppGray}\textbf{Dynamic payload}
& Excluded current codes; second-pass query plan; up to 24,000 characters of readable query evidence; and up to 8,000 characters of patient history. Maximum 60 candidates after normalization and dictionary validation.\\
\bottomrule
\end{tabularx}
\end{table}

\begin{table}[H]
\centering
\footnotesize
\setlength{\tabcolsep}{6pt}
\renewcommand{\arraystretch}{1.12}
\caption{Exact static default MIMIC-IV Iteration~2 judge prompt (archived
Stage~2). The same function is called for the initial pass and the targeted
repair pass; only the opening prefix and previous-decision payload differ.}
\label{tab:s-v2-judge-prompt}
\begin{tabularx}{\textwidth}{>{\raggedright\arraybackslash}p{2.75cm}Y}
\toprule
\rowcolor{SuppBlue}
\multicolumn{2}{c}{\textbf{\textcolor{SuppBlueDark}{Stage 2C: Default Three-Part Add Judge}}}\\
\midrule
\cellcolor{SuppGray}\textbf{System message}
& Judge patient anchor, literature bridge, and exact clinical-code mapping separately. JSON only.\\
\midrule
\cellcolor{SuppGray}\textbf{Opening instruction}
& ``Second-stage'' (initial call) or ``Evidence-repair'' (repair call) clinical-code add editor using a three-part evidence contract. Assess only the supplied candidates. Do not use ground truth, target/post codes, or hidden pretrain tail.\\
\midrule
\cellcolor{SuppOrange}\textbf{Patient anchor}
& PATIENT ANCHOR comes only from this EHR: repeated code, active diagnosis/treatment, durable procedure/device/status, explicit plan, or direct acute/lab event.\\
\midrule
\cellcolor{SuppOrange}\textbf{Literature bridge}
& LITERATURE BRIDGE comes from candidate-bound E or QEV sources and verifies the claimed medical persistence, recurrence, treatment consequence, diagnostic threshold, or durable-status relation. It does not need to prove that this individual patient will be coded next.\\
\midrule
\cellcolor{SuppOrange}\textbf{Exact mapping}
& EXACT MAPPING requires the patient anchor to match the candidate entity, subtype, site, cause, acuity, and status.\\
\midrule
\cellcolor{SuppGreen}\textbf{Clinical policy}
& Durable procedure/device status can persist unless removal is documented. Chronic disease needs repetition or an active treatment/problem clue. Acute events, infections, injuries, laboratory abnormalities, and precise complications require a direct patient event, threshold, or explicit plan; general risk is insufficient. Literature cannot create a patient diagnosis.\\
\midrule
\cellcolor{SuppPurple}\textbf{Exact output contract}
& One record per assessed candidate with fields \texttt{code}, \texttt{accept}, \texttt{confidence}, \texttt{support\_type}, \texttt{patient\_anchor\_present}, \texttt{patient\_anchor\_type}, \texttt{literature\_bridge\_supported}, \texttt{exact\_mapping\_supported}, \texttt{evidence\_ids}, \texttt{query\_evidence\_ids}, and a brief role-separated \texttt{rationale}.\\
\midrule
\cellcolor{SuppGray}\textbf{Dynamic payload}
& Current codes; Stage~1 accepted additions; previous decision when present; the last 8,000 history characters; up to 15,000 characters of candidates; up to 26,000 characters of candidate-bound readable evidence; and up to 16,000 characters of readable discovery evidence. The assembled prompt is capped at 30,000 characters.\\
\midrule
\cellcolor{SuppGreen}\textbf{Runtime gate}
& Post-processing retains only \texttt{accept=true} records for which all three booleans are true, confidence is at least the configured threshold (0.68 in the frozen logic), support is not historical/risk-only, and evidence IDs resolve to the candidate or matched query.\\
\bottomrule
\end{tabularx}
\end{table}

\begin{table}[H]
\centering
\footnotesize
\setlength{\tabcolsep}{6pt}
\renewcommand{\arraystretch}{1.13}
\caption{Exact static MIMIC-IV verdict-completion prompt used when a long Stage~2 response (Iteration~2 in the method) omitted candidate verdicts. It is a repair/completion call, not the default Stage~2 policy.}
\label{tab:s-completion-judge-prompt}
\begin{tabularx}{\textwidth}{>{\raggedright\arraybackslash}p{2.75cm}Y}
\toprule
\rowcolor{SuppBlue}
\multicolumn{2}{c}{\textbf{\textcolor{SuppBlueDark}{Stage 2D: Untruncated Verdict-Completion Judge}}}\\
\midrule
\cellcolor{SuppGray}\textbf{System message}
& Judge every supplied clinical-code candidate. Return one verdict per code. JSON only.\\
\midrule
\cellcolor{SuppGray}\textbf{Core instruction}
& You are the legacy second-stage clinical-code add judge. Assess every supplied candidate independently. Do not omit a candidate and do not rank candidates against one another. Patient history is primary evidence. Literature may verify code meaning, persistence, recurrence, treatment consequence, or a bridge from the patient context to the candidate, but literature cannot create a patient diagnosis.\\
\midrule
\cellcolor{SuppOrange}\textbf{Acceptance rule}
& Add chronic or persistent conditions only with an explicit prior diagnosis, repeated code, active treatment, durable status/procedure, or other concrete ongoing clue. Add acute events, laboratory abnormalities, injuries, infections, external causes, or precise complications only with a direct current event, lab/treatment context, or strong recurrent pattern. A prior code is a clue but may support a persistent chronic condition when the candidate and patient history match exactly. Reject a related but different subtype. DNR alone does not establish a palliative-care encounter.\\
\midrule
\cellcolor{SuppPurple}\textbf{Structured output}
& \promptbox{%
\{"add":[\{\par
\quad "code":"ICD\_...", "direct\_support":true,\par
\quad "support\_type":"persistent\_current|procedure\_supported|\par
\qquad active\_context|prior\_only|speculative",\par
\quad "evidence\_ids":["E1"],\par
\quad "query\_evidence\_ids":["QEV\_E1"],\par
\quad "rationale":"brief"\par
\}]\}}\\
\midrule
\cellcolor{SuppOrange}\textbf{Interface limitation}
& Unlike Table~\ref{tab:s-v2-judge-prompt}, this exact archived response does
not contain \texttt{accept}, \texttt{confidence},
\texttt{literature\_bridge\_supported}, or
\texttt{exact\_mapping\_supported}. No unreported field mapping is assumed;
a uniform-gate release must rejudge these fixed proposals with the default
three-part contract.\\
\midrule
\cellcolor{SuppGray}\textbf{Dynamic payload}
& Current codes; up to 8,000 characters of patient history; and every candidate paired with its dictionary title, candidate-bound evidence, and semantically matched discovery evidence. Missing verdicts are retried per ICD until every candidate has one output record.\\
\midrule
\cellcolor{SuppGreen}\textbf{Observed enforcement gap}
& The natural-language policy is stricter than the saved decisions: row 59 accepted several candidates whose stated anchors are absent from the supplied structured trajectory. This motivates immutable anchor spans rather than free-text anchor claims.\\
\bottomrule
\end{tabularx}
\end{table}

\begin{table}[H]
\centering
\footnotesize
\setlength{\tabcolsep}{6pt}
\renewcommand{\arraystretch}{1.10}
\caption{Exact static MIMIC-III verdict-completion prompts. The batch call retries omitted codes; any code still missing after those retries receives the fixed-candidate fallback call.}
\label{tab:s-mimic3-completion-judge-prompt}
\begin{tabularx}{\textwidth}{>{\raggedright\arraybackslash}p{2.75cm}Y}
\toprule
\rowcolor{SuppBlue}
\multicolumn{2}{c}{\textbf{\textcolor{SuppBlueDark}{MIMIC-III ICD-9 Verdict Completion and Fixed-Candidate Fallback}}}\\
\midrule
\cellcolor{SuppGray}\textbf{Batch system message}
& Judge patient anchor, literature bridge, and exact ICD-9 mapping separately. JSON only.\\
\midrule
\cellcolor{SuppGray}\textbf{Batch opening}
& You are a strict future-admission ICD-9 prediction judge using a three-part evidence contract. Everything in PATIENT HISTORY happened before the unknown target admission. Judge each role independently.\\
\midrule
\cellcolor{SuppOrange}\textbf{Three exact roles}
& 1. PATIENT ANCHOR comes only from this EHR. It must be a repeated code, active diagnosis/treatment, durable procedure/device/status, explicit plan, or direct acute/lab precursor.\par
2. LITERATURE BRIDGE comes from candidate-bound Reader evidence. It verifies the claimed general medical persistence, recurrence, treatment consequence, diagnostic threshold, or durable-status relation. It does not need to prove that this individual patient will be coded next.\par
3. EXACT MAPPING requires the patient anchor to match the candidate entity, subtype, site, cause, acuity, and status.\\
\midrule
\cellcolor{SuppGreen}\textbf{Batch policy}
& Durable procedure/device status can persist unless removal is documented. Chronic disease can be supported by either repetition or a concrete active treatment/problem clue. Acute symptoms, infection, injury, lab abnormalities, and precise complications require direct patient evidence; general risk is insufficient. Literature cannot create a patient diagnosis. Return one record for every assessed candidate. JSON only.\\
\midrule
\cellcolor{SuppPurple}\textbf{Batch contract}
& \texttt{add[code, accept, confidence, support\_type, patient\_anchor\_present, patient\_anchor\_type, literature\_bridge\_supported, exact\_mapping\_supported, evidence\_ids, rationale]}; dynamic sections are \texttt{CURRENT CODES}, \texttt{CANDIDATES}, the last 8,500 history characters, and \texttt{CANDIDATE-BOUND READER EVIDENCE}.\\
\midrule
\cellcolor{SuppGray}\textbf{Fallback system message}
& Return one explicit accept or reject verdict for the fixed ICD-9 candidate. JSON only.\\
\midrule
\cellcolor{SuppOrange}\textbf{Fallback user instruction}
& Judge whether this ICD-9 candidate should be added to the prediction for the patient's final target admission. Use only the pre-target patient history as the clinical anchor. The attached literature may validate meaning, persistence, recurrence, treatment consequence, or a medical bridge, but cannot invent a patient diagnosis. Chronic and durable conditions need an explicit prior diagnosis, active treatment, repeated code, durable procedure/status, or another concrete ongoing clue. Acute or precise conditions need a direct event or strong recurrent pattern. Reject association-only, risk-only, mismatched-subtype, and unsupported candidates.\\
\midrule
\cellcolor{SuppPurple}\textbf{Fallback contract}
& You must return a verdict even when the candidate is invalid or unsupported. Do not return the ICD code. JSON only:\par
\promptbox{%
\{"accept":false, "confidence":0.0,\par
\quad "support\_type":"unsupported",\par
\quad "patient\_anchor\_present":false,\par
\quad "literature\_bridge\_supported":false,\par
\quad "exact\_mapping\_supported":false,\par
\quad "evidence\_ids":[],\par
\quad "rationale":"brief medical rationale"\}}\\
\bottomrule
\end{tabularx}
\end{table}

\begin{table}[H]
\centering
\footnotesize
\setlength{\tabcolsep}{6pt}
\renewcommand{\arraystretch}{1.13}
\caption{Exact static MIMIC-IV and MIMIC-III research-confidence prompts plus
the deterministic Union+RRF selector replacement reported as
ICD-Deepresearch w/o Evidence-Aware Reranker. Each confidence agent scores
every dictionary-valid candidate; RRF combines the resulting research rank
with the independent Direct GPT-5 rank.}
\label{tab:s-rank-fusion}
\begin{tabularx}{\textwidth}{>{\raggedright\arraybackslash}p{2.75cm}Y}
\toprule
\rowcolor{SuppBlue}
\multicolumn{2}{c}{\textbf{\textcolor{SuppBlueDark}{Research Confidence Ranking and Candidate-Union Fusion}}}\\
\midrule
\cellcolor{SuppGray}\textbf{System message}
& You are a calibrated next-visit ICD-10-CM diagnosis ranking model. Score candidates independently, then calibrate the ranking jointly. JSON only.\\
\midrule
\cellcolor{SuppOrange}\textbf{MIMIC-IV instruction}
& Rank every supplied ICD-10-CM candidate for the patient's unknown final visit. The history below is the only patient-specific information available before the target visit. Predict which exact diagnoses are likely to be coded at that next visit. A past code can recur, persist, resolve, or be irrelevant. General retrieved evidence may support a medical transition but does not prove that it will occur in this patient. Use one calibrated 0--100 probability scale across all candidates. Reward active chronic disease, durable status, repeated diagnoses, plausible complications, and coherent care transitions. Penalize resolved events, unsupported specificity, distant associations, and diagnoses supported only by a generic code-description page.\\
\midrule
\cellcolor{SuppPurple}\textbf{MIMIC-IV output}
& \promptbox{%
\{"eov\_threshold":50,\par
\quad "scores":[\{"code":"ICD\_...", "confidence":0\}]\}}\par
Return one patient-specific \texttt{eov\_threshold} (0--100). The complete variable-length prediction contains candidates with confidence $\geq$ \texttt{eov\_threshold}. Return JSON only, include every supplied code exactly once, and introduce no new code.\\
\midrule
\cellcolor{SuppGray}\textbf{MIMIC-III system}
& You are a calibrated next-admission ICD-9 ranking model. Evaluate candidates independently, then calibrate them jointly. JSON only.\\
\midrule
\cellcolor{SuppOrange}\textbf{MIMIC-III instruction}
& Score every candidate ICD-9-CM diagnosis for the unknown final admission. The PATIENT HISTORY contains only information available before the target admission. Do not infer that a historical diagnosis must recur. Estimate the probability that the exact candidate code, at its stated specificity, will be assigned during the next admission. Distinguish persistent active disease, durable status, recurrent disease, likely complications, isolated past events, and speculative associations. Retrieved sources support general medical transitions but do not prove this patient's future diagnosis. Use one consistent 0--100 scale across all candidates: 90--100 near-certain; 70--89 likely; 45--69 plausible but uncertain; 20--44 weak; 0--19 unlikely.\\
\midrule
\cellcolor{SuppPurple}\textbf{MIMIC-III output}
& Also return one patient-specific \texttt{eov\_threshold} from 0--100. A candidate with confidence $\geq$ \texttt{eov\_threshold} is in the variable-length diagnosis prediction set before EOV. The threshold should balance missing likely diagnoses against unsupported additions. Return JSON only. Include every supplied code exactly once and do not introduce new codes:\par
\promptbox{%
\{"eov\_threshold":0,\par
\quad "scores":[\{"code":"ICD\_...",\par
\qquad "confidence":0,\par
\qquad "include\_before\_eov":false\}]\}}\\
\midrule
\cellcolor{SuppGreen}\textbf{RRF selector replacement}
& For candidate $c$, $s(c)=\mathbb{1}[c\in D]/(60+r_D(c))+\mathbb{1}[c\in R]/(60+r_R(c))$. Codes are dictionary validated and the union is sorted by $s(c)$; evaluation uses the leading $K$ entries. No LLM call occurs in this step.\\
\midrule
\cellcolor{SuppGray}\textbf{Interpretive boundary}
& RRF rewards cross-branch agreement and can promote complementary mid-ranked codes, but its score is not a probability and does not establish that retrieval caused either branch's ranking.\\
\bottomrule
\end{tabularx}
\end{table}

\begin{table}[H]
\centering
\footnotesize
\setlength{\tabcolsep}{6pt}
\renewcommand{\arraystretch}{1.13}
\caption{Exact static Direct GPT forecasting prompt used for Direct Forecast
Only and as the independent direct branch of the RRF selector replacement.
The model name changes by arm; no retrieval tool, candidate pool, baseline
prediction, dictionary content, or other system output is supplied.}
\label{tab:s-direct-prompt}
\begin{tabularx}{\textwidth}{>{\raggedright\arraybackslash}p{2.75cm}Y}
\toprule
\rowcolor{SuppBlue}
\multicolumn{2}{c}{\textbf{\textcolor{SuppBlueDark}{Independent Direct Forecast Prompt}}}\\
\midrule
\cellcolor{SuppGray}\textbf{System message}
& Independently forecast ICD-10-CM diagnoses from the supplied pre-target patient history. Return JSON only.\\
\midrule
\cellcolor{SuppGray}\textbf{Opening and restriction}
& Forecast the ICD-10-CM diagnosis codes most likely to be assigned during this patient's unknown next/final admission. Use ONLY the observed pre-target patient history below and your own medical/coding knowledge. No future encounter information is available. You have no candidate list, baseline predictions, retrieved sources, external tools, or outputs from another system.\\
\midrule
\cellcolor{SuppOrange}\textbf{Prediction requirements}
& Independently generate 20 to 30 unique ICD-10-CM diagnosis codes. Do not output ICD-9-CM, CPT, procedure codes, or prose in the code field. Rank them from most to least likely for the next/final admission. Give each code a confidence from 0 to 100. Confidence must be non-increasing down the list and should reflect uncertainty rather than forcing every item to look likely. Forecast future coding rather than copying every historical code. Balance persistent diseases/statuses, recurrent conditions, likely acute complications, and procedure/treatment consequences. Preserve exact ICD-10-CM specificity. Do not invent patient facts or assume access to the held-out admission.\\
\midrule
\cellcolor{SuppPurple}\textbf{Structured output}
& \promptbox{%
\{"ranked":[\{\par
\quad "code":"ICD-10-CM code", "confidence":0\par
\}]\}}\\
\midrule
\cellcolor{SuppGray}\textbf{Dynamic payload}
& \textbf{Observed pre-target patient history:} \texttt{<row.pre\_text>}. On validation failure the exact suffix is: ``Prior structural errors: \texttt{<feedback>}. Regenerate the complete ranking without any external candidate list.''\\
\midrule
\cellcolor{SuppGreen}\textbf{MIMIC-III resolution}
& The same executable template resolves \texttt{codeSystem} to ``ICD-9-CM'' and \texttt{otherCodeSystem} to ``ICD-10-CM''; all other static sentences and the structural retry suffix are unchanged.\\
\bottomrule
\end{tabularx}
\end{table}

\begin{table}[H]
\centering
\footnotesize
\setlength{\tabcolsep}{6pt}
\renewcommand{\arraystretch}{1.12}
\caption{Exact static GPT-5 + Web Search (Standalone) baseline prompt. Unlike ICD-Deepresearch, this agent chooses its own search count and directly produces an unbounded code ranking from OpenAI Web Search results.}
\label{tab:s-websearch-prompt}
\begin{tabularx}{\textwidth}{>{\raggedright\arraybackslash}p{2.75cm}Y}
\toprule
\rowcolor{SuppBlue}
\multicolumn{2}{c}{\textbf{\textcolor{SuppBlueDark}{GPT-5 + Web Search (Standalone)}}}\\
\midrule
\cellcolor{SuppGray}\textbf{System message}
& Independently produce an exhaustive, unbounded ICD-10-CM forecast from pre-target history. Use Web Search for general medical and coding evidence, maximize recall, and return JSON only.\\
\midrule
\cellcolor{SuppGray}\textbf{Opening and restriction}
& Forecast every distinct ICD-10-CM diagnosis code that could plausibly be assigned during the unknown target encounter represented by the final empty \texttt{icd\_list\_begin} marker. Use ONLY the observed pre-target patient history below, your medical and coding knowledge, and the provided Web Search tool. No future encounter information is available. You have no baseline predictions, candidate list, dictionary entries, or outputs from another system.\\
\midrule
\cellcolor{SuppOrange}\textbf{Search policy}
& You MUST use Web Search before finalizing the ranking: Decide the number of searches yourself from the complexity and uncertainty of this patient's trajectory; there is no minimum or maximum search-count target beyond using the tool at least once. Search each clinically distinct unresolved trajectory when doing so could change candidate inclusion, code specificity, confidence, or rank. Continue searching while another focused query is likely to add decision-relevant evidence; stop when additional searches would only repeat information already found. Search for diagnosis progression, common next-encounter diagnoses, treatment/complication relations, and exact ICD-10-CM terminology when useful. Do not include patient identifiers or long excerpts from the patient history in a query. Use retrieved pages as general medical or coding evidence, not as hidden patient facts.\\
\midrule
\cellcolor{SuppGreen}\textbf{Unbounded objective}
& The objective is exhaustive recall. There is deliberately NO prediction-count budget: Do not target or stop at 20, 30, or any other preset number. Generate as many distinct, medically defensible codes as the trajectory can support, including lower-confidence possibilities. Continue until further codes would be unrelated to this patient or would merely duplicate a more appropriate code already listed.\\
\midrule
\cellcolor{SuppPurple}\textbf{Ranking/output}
& 1. Rank all generated codes from most to least likely for the unknown target encounter. 2. Assign confidence from 0 to 100 in non-increasing order. Low-confidence candidates are allowed and should remain low. 3. Cover the full trajectory, including persistent and recurrent diagnoses, durable status/history codes, treatment consequences, plausible acute developments, comorbidities, and follow-up-related diagnoses. 4. Preserve exact ICD-10-CM specificity when justified. Do not invent hidden patient facts. 5. Do not output ICD-9-CM, CPT, ICD-10-PCS procedure codes, or prose in the code field. 6. Return JSON only: \texttt{\{"ranked":[\{"code":"ICD-10-CM code","confidence":0\}]\}}.\\
\midrule
\cellcolor{SuppGray}\textbf{Tool and retry}
& OpenAI Responses API tool: \texttt{web\_search\_preview}, \texttt{tool\_choice="required"}, \texttt{search\_context\_size="low"}. The exact retry suffix is: ``Prior structural errors: \texttt{<feedback>}. Regenerate the complete exhaustive list without imposing a count limit.'' The MIMIC-III resolution substitutes the ICD-9-CM target sentence and its exact opposite-vocabulary exclusion.\\
\bottomrule
\end{tabularx}
\end{table}

\begin{table}[H]
\centering
\footnotesize
\setlength{\tabcolsep}{6pt}
\renewcommand{\arraystretch}{1.12}
\caption{Exact static Direct-model explanation prompt, distinct from the
evidence-linked explanation writer.}
\label{tab:s-direct-explanation-prompt}
\begin{tabularx}{\textwidth}{>{\raggedright\arraybackslash}p{2.75cm}Y}
\toprule
\rowcolor{SuppBlue}
\multicolumn{2}{c}{\textbf{\textcolor{SuppBlueDark}{Direct Locked-Prediction Explanation Writer}}}\\
\midrule
\cellcolor{SuppGray}\textbf{System message}
& Explain locked next-admission ICD-10-CM predictions using only the supplied patient history. Return JSON only.\\
\midrule
\cellcolor{SuppGray}\textbf{User instruction}
& Explain the following locked ICD-10-CM predictions for the patient's unknown next/final admission. Use ONLY the observed pre-target patient history and your own medical/coding knowledge. The prediction set is already locked: do not add, remove, replace, or reorder codes. No gold answer, candidate pool, baseline prediction, retrieval result, external source, or other system output is available.\\
\midrule
\cellcolor{SuppOrange}\textbf{Medical output}
& For every code, write one concise English medical rationale linking observed history to possible next-admission coding, plus one concise uncertainty statement. Do not claim that the future diagnosis is known. Do not mention models, prompts, experiments, candidates, or evaluation.\\
\midrule
\cellcolor{SuppPurple}\textbf{Exact JSON}
& \promptbox{%
\{"explanations":[\{\par
\quad "code":"locked code",\par
\quad "rationale":"medical rationale",\par
\quad "uncertainty":"specific limitation"\par
\}]\}}\\
\midrule
\cellcolor{SuppGray}\textbf{Dynamic payload/retry}
& Ten locked predictions per parallel explanation chunk and the full observed pre-target history. Validation requires one explanation per locked code. The exact retry suffix is: ``Prior structural errors: \texttt{<feedback>}. Return corrected complete JSON.''\\
\midrule
\cellcolor{SuppGreen}\textbf{MIMIC-III resolution}
& The same executable template replaces ``ICD-10-CM'' with ``ICD-9-CM'' in the system, opening, and output-schema context; the remaining static prose is unchanged.\\
\bottomrule
\end{tabularx}
\end{table}

\begin{table}[H]
\centering
\footnotesize
\setlength{\tabcolsep}{6pt}
\renewcommand{\arraystretch}{1.13}
\caption{Exact static evidence-linked explanation prompt for ICD-Deepresearch
(RRF-selected; legacy artifact label AF), segmented for readability.
Predictions and order are finalized before this call, so the explanation agent
cannot improve prediction metrics; this prompt was not evaluated on the full
ICD-Deepresearch outputs produced by the Evidence-Aware Reranker (legacy
label FR). Verbatim ``locked'' language below reflects the archived $K=20$
implementation.}
\label{tab:s-explanation-prompt}
\begin{tabularx}{\textwidth}{>{\raggedright\arraybackslash}p{2.75cm}Y}
\toprule
\rowcolor{SuppBlue}
\multicolumn{2}{c}{\textbf{\textcolor{SuppBlueDark}{Locked-Prediction Explanation Agent}}}\\
\midrule
\cellcolor{SuppGray}\textbf{System message}
& Write evidence-linked medical explanations for locked next-admission ICD-10-CM forecasts. Return JSON only.\\
\midrule
\cellcolor{SuppGray}\textbf{Prediction lock}
& Write concise English medical explanations for the 20 locked ICD-10-CM forecasts for this patient's unknown next/final admission. The prediction set and ranking are fixed. Do not add, remove, replace, or reorder codes. Use only the observed pre-target history, the candidate-specific clues, and the supplied evidence ledger. The held-out admission and gold codes are not available.\\
\midrule
\cellcolor{SuppOrange}\textbf{Medical requirements}
& 1. Explain medicine only. Never mention a model, system, baseline, candidate-generation stage, rank, prompt, experiment, or evaluation. 2. For every exact code, state a patient-specific medical bridge from observed history to possible next-admission documentation. Similar diagnoses may share concise reasoning, but each code must receive its own explanation. 3. Preserve exact disease, organ, acuity, status, device, complication, and laterality. Do not substitute a nearby diagnosis. 4. Chronic/status/history codes may persist when their anchor remains relevant. Acute diagnoses, symptoms, injuries, and abnormalities require recurrence or renewed documentation; do not portray them as certain. 7. Give one medically specific uncertainty statement. Do not invent future symptoms, tests, procedures, treatments, or diagnoses. 8. Return every locked code exactly once as JSON.\\
\midrule
\cellcolor{SuppGreen}\textbf{Citation contract}
& 5. Use a definition reference only to establish code meaning. Use a relation source only for a claim actually supported by its excerpt. General literature cannot prove this patient's future event. 6. Cite only reference IDs assigned to that same code. Put bracketed citations such as \texttt{[1][2]} next to the supported claim and return those same integers in \texttt{citation\_ids}. Every explanation must cite at least its definition reference.\\
\midrule
\cellcolor{SuppPurple}\textbf{Structured output}
& \promptbox{%
\{"explanations":[\{\par
\quad "code":"locked code",\par
\quad "rationale":"medical rationale with [id]",\par
\quad "uncertainty":"specific limitation",\par
\quad "citation\_ids":[1]\par
\}]\}}\\
\midrule
\cellcolor{SuppGray}\textbf{Dynamic payload}
& Observed history; locked predictions with code-specific clues and allowed reference IDs; and numbered evidence ledger. Structural validation requires all 20 codes exactly once and rejects cross-code or missing citations. The exact retry suffix is: ``Prior structural errors: \texttt{<feedback>}. Return a corrected complete response.''\\
\midrule
\cellcolor{SuppGreen}\textbf{MIMIC-III resolution}
& The same executable template replaces ``ICD-10-CM'' with ``ICD-9-CM'' in its two code-system slots; all numbered requirements, payload labels, and retry text are unchanged.\\
\bottomrule
\end{tabularx}
\end{table}

\begin{table}[H]
\centering
\footnotesize
\setlength{\tabcolsep}{6pt}
\renewcommand{\arraystretch}{1.12}
\caption{Exact static candidate-controlled shared-pool selector prompts. MIMIC-IV and MIMIC-III use separate executable templates, reproduced here rather than described as a simple vocabulary substitution.}
\label{tab:s-shared-selector-prompt}
\begin{tabularx}{\textwidth}{>{\raggedright\arraybackslash}p{2.75cm}Y}
\toprule
\rowcolor{SuppBlue}
\multicolumn{2}{c}{\textbf{\textcolor{SuppBlueDark}{Evaluation-Only Shared-Pool Selector}}}\\
\midrule
\cellcolor{SuppGray}\textbf{System message}
& You rank a fixed candidate pool for prospective ICD-10-CM forecasting. Return JSON only.\\
\midrule
\cellcolor{SuppGray}\textbf{Opening}
& Predict exactly 20 unique ICD-10-CM diagnoses for the patient's unknown next encounter. This is prospective next-encounter forecasting, not coding a future note. The held-out target is unavailable. Select only from the supplied valid candidate pool. Candidate order is deterministically randomized and carries no rank signal.\\
\midrule
\cellcolor{SuppOrange}\textbf{Requirements}
& Return exactly 20 different \texttt{candidate\_id} values, ordered from highest to lowest confidence. Copy IDs exactly; do not add, duplicate, substitute, or omit a slot. Rank persistent active disease, recurrent conditions, near-term treatment consequences, and strongly anchored follow-up needs above generic medical possibilities. Preserve exact ICD entity, acuity, anatomy, status, device, laterality, and complication. Do not invent patient facts or assume access to the held-out encounter.\\
\midrule
\cellcolor{SuppGreen}\textbf{Controlled contrast}
& Ledger arm: ``Use the supplied target-blind research ledger critically. A definition or general relation does not by itself prove next-encounter occurrence.'' No-ledger arms: ``Use only the observed trajectory and candidate code identities; no research ledger is available.''\\
\midrule
\cellcolor{SuppPurple}\textbf{MIMIC-IV output/payload}
& Return \texttt{\{"ranked":[\{"candidate\_id":"C01","confidence":0\}]\}}. The dynamic payload contains the pre-target trajectory and the same randomized pool; only the ledger arm receives candidate-specific research fields. The exact retry suffix is: ``Prior response errors: \texttt{<feedback>}. Return corrected JSON.''\\
\midrule
\cellcolor{SuppGray}\textbf{MIMIC-III system}
& You produce an exact-cardinality next-admission ICD-9 ranking. Return JSON only.\\
\midrule
\cellcolor{SuppOrange}\textbf{MIMIC-III opening}
& Predict exactly 20 unique ICD-9-CM diagnoses for the patient's unknown next/final admission. This is next-encounter forecasting, not extraction from a future note. The gold answer is unavailable. Select only from the supplied valid candidate pool. The list order is randomized and contains no rank signal.\\
\midrule
\cellcolor{SuppOrange}\textbf{MIMIC-III requirements}
& 1. Return exactly 20 different \texttt{candidate\_id} values, ordered best to worst. Copy IDs exactly from the pool. Never add, duplicate, substitute, or omit a slot. 2. For each selected \texttt{candidate\_id}, provide confidence 0--100. Confidence and list order must agree. 3. Distinguish persistent active disease/status, recurrent conditions, likely complications, isolated historical events, and unsupported possibilities. 4. Preserve the exact ICD entity, acuity, anatomy, status, device, laterality, and complication. Do not invent patient facts. 5. Do not mention models, systems, prompts, candidates, stages, experiments, or evaluation.\\
\midrule
\cellcolor{SuppGreen}\textbf{MIMIC-III contrast/output}
& Ledger arm: ``6. Use the supplied patient clues and sources critically. Definitions establish identity, not future occurrence; general literature does not prove this patient will receive the code.'' No-ledger arms: ``6. Use only the observed trajectory and code identities; no retrieved sources are available.'' Requirement 7 returns the same exact \texttt{ranked[candidate\_id,confidence]} JSON. The exact retry suffix is: ``Prior ranking errors: \texttt{<feedback>}. Return corrected complete JSON.''\\
\bottomrule
\end{tabularx}
\end{table}

\begin{table}[H]
\centering
\footnotesize
\setlength{\tabcolsep}{6pt}
\renewcommand{\arraystretch}{1.10}
\caption{Exact static MIMIC-III shared-pool explanation prompt. These
post-selection reasons were generated by the $K=20$ shared-pool runner but do
not enter the candidate-controlled prediction metrics. Verbatim prompt text is
preserved below.}
\label{tab:s-mimic3-shared-explanation-prompt}
\begin{tabularx}{\textwidth}{>{\raggedright\arraybackslash}p{2.75cm}Y}
\toprule
\rowcolor{SuppBlue}
\multicolumn{2}{c}{\textbf{\textcolor{SuppBlueDark}{MIMIC-III Shared-Pool Locked-Prediction Explainer}}}\\
\midrule
\cellcolor{SuppGray}\textbf{System message}
& You explain locked next-admission ICD-9 predictions with concise medical reasoning. Return JSON only.\\
\midrule
\cellcolor{SuppGray}\textbf{Opening}
& Write an English medical next-encounter explanation for each of the \texttt{<chunk size>} locked ICD-9-CM predictions below. The predictions and gold answer are unavailable for revision: do not add, remove, substitute, or reorder codes. Explain the forecast from only the observed pre-target record and supplied method-specific information.\\
\midrule
\cellcolor{SuppOrange}\textbf{Requirements 1--5}
& 1. Return all \texttt{<chunk size>} locked codes exactly once. 2. For each, write one concise patient-specific medical rationale and one concise uncertainty statement. 3. Preserve exact ICD identity and avoid invented patient or future-visit facts. 4. For weak predictions, say specifically why evidence is insufficient instead of making the code sound certain. 5. Do not mention models, systems, prompts, candidates, stages, ranks, experiments, or evaluation.\\
\midrule
\cellcolor{SuppGreen}\textbf{Method-specific rule}
& Ledger arm: ``6. \texttt{citation\_ids} may contain only supplied reference IDs assigned to that exact code. Include at least its definition reference. General literature does not establish patient-specific future occurrence.'' No-ledger arms: ``6. \texttt{citation\_ids} must be empty. Do not invent citations, references, or URLs.''\\
\midrule
\cellcolor{SuppPurple}\textbf{Output/payload}
& 7. Return JSON only: \texttt{\{"explanations":[\{"code":"ICD\_...","rationale":"one concise medical sentence","uncertainty":"one concise caveat","citation\_ids":[]\}]\}}. Payload headings are \texttt{OBSERVED PRE-TARGET PATIENT CONTEXT}, \texttt{LOCKED TOP-20 PREDICTIONS}, \texttt{METHOD-SPECIFIC INFORMATION FOR THOSE CODES}, and, for the ledger arm, \texttt{SUPPLIED REFERENCES}.\\
\midrule
\cellcolor{SuppGray}\textbf{Chunk retry}
& Two ten-code chunks run in parallel. The exact retry suffix is: ``Prior chunk \texttt{<chunk index>} errors: \texttt{<feedback>}. Return corrected complete JSON.''\\
\bottomrule
\end{tabularx}
\end{table}

\begin{table}[H]
\centering
\footnotesize
\setlength{\tabcolsep}{6pt}
\renewcommand{\arraystretch}{1.12}
\caption{Exact static source-metadata proxy judge prompt. The archived response schema uses utility-related field names, but this audit sees only source metadata and patient history, not page bodies, system identity, predictions, or gold labels; its outputs are not source-utility measurements.}
\label{tab:s-source-judge-prompt}
\begin{tabularx}{\textwidth}{>{\raggedright\arraybackslash}p{2.75cm}Y}
\toprule
\rowcolor{SuppBlue}
\multicolumn{2}{c}{\textbf{\textcolor{SuppBlueDark}{Label-Blind Retrieval-Item Utility Judge}}}\\
\midrule
\cellcolor{SuppGray}\textbf{System message}
& You are a blinded clinical retrieval evaluator. Return JSON only.\\
\midrule
\cellcolor{SuppGray}\textbf{Task/blinding}
& Evaluate one retrieved webpage for a next-visit diagnosis prediction system. TASK: Given only the patient's pre-target history, predict diagnosis codes documented at the immediately next hospital visit. BLINDING: You are not told which retrieval system selected this page. Do not infer quality from the query style. Gold diagnoses and model predictions are intentionally absent.\\
\midrule
\cellcolor{SuppOrange}\textbf{Exact definitions}
& \texttt{credible}: URL/domain and title indicate a medically or officially credible source.\par
\texttt{patient\_relevant}: the page addresses a diagnosis, mechanism, complication, or follow-up issue grounded in this patient's history.\par
\texttt{forecast\_useful}: it supplies a concrete clinical bridge that could help rank a diagnosis for the immediate next visit. A generic ICD definition alone is NOT forecast-useful.\par
\texttt{coding\_only}: useful only for code identity/specificity, without improving which diagnosis is likely next.\par
\texttt{direct}: concrete patient-grounded next-visit bridge; \texttt{indirect}: relevant background or coding help but weak temporal discrimination; \texttt{none}: irrelevant/noisy/unhelpful.\\
\midrule
\cellcolor{SuppGreen}\textbf{Dynamic payload}
& Patient history (first 5,000 characters), retrieval query (first 600), page title (first 300), and page URL. The exact final instruction is: ``Return a strict, conservative judgment. The reason must be one short sentence.''\\
\midrule
\cellcolor{SuppPurple}\textbf{JSON fields}
& \texttt{credible}, \texttt{patient\_relevant}, \texttt{forecast\_useful}, and \texttt{coding\_only} booleans; \texttt{utility\_level} in \texttt{direct|indirect|none}; and one-sentence \texttt{reason}.\\
\bottomrule
\end{tabularx}
\end{table}

\begin{table}[H]
\centering
\footnotesize
\setlength{\tabcolsep}{6pt}
\renewcommand{\arraystretch}{1.10}
\caption{Exact static prompt for the automated, gold-aware GPT-5 Reason5
semantic audit, segmented for readability. All prediction and explanation
outputs are frozen before the complete gold answer is supplied to this
evaluation-only judge. Items hide method identity, but the audit is not
independent clinical adjudication.}
\label{tab:s-eval-prompts}
\begin{tabularx}{\textwidth}{>{\raggedright\arraybackslash}p{2.75cm}Y}
\toprule
\rowcolor{SuppBlue}
\multicolumn{2}{c}{\textbf{\textcolor{SuppBlueDark}{Gold-Aware Item-Blind Five-Dimension Explanation Judge}}}\\
\midrule
\cellcolor{SuppGray}\textbf{System message}
& You are a strict blinded clinical forecast-reason adjudicator. Return JSON only.\\
\midrule
\cellcolor{SuppGray}\textbf{Opening}
& You are a blinded clinical adjudicator evaluating individually anonymized reasons for exact Top-20 next-admission ICD-10-CM forecasts. The complete held-out gold code set is supplied only for post-generation evaluation. All generators were target-blind.\\
\midrule
\cellcolor{SuppOrange}\textbf{Symmetric set rule}
& Prediction-set evaluation is fixed and symmetric: every false prediction and every missed gold code receives zero reason utility. Do not let fluent prose, confidence, citations, or medical plausibility rescue either error. Score only reasons attached to true-positive predicted codes because only a correct set prediction can earn reason utility.\\
\midrule
\cellcolor{SuppOrange}\textbf{Item blinding}
& Each \texttt{anonymous\_reason\_item} is independently shuffled and may come from any system. The opaque \texttt{item\_id} contains no model identity. You are not shown the originating system's prediction set, TP count, FP count, FN count, rank, or confidence. Do not group items by writing style or infer generator identity. Score each reason on its own content.\\
\midrule
\cellcolor{SuppGreen}\textbf{Five exact dimensions}
& \textbf{Clinical validity:} Are the medical relationships, mechanisms, and claims clinically sound and free of substantive contradiction or hallucination? Gold occurrence proves that the code appeared, not that the stated medical mechanism is correct.\par
\textbf{Patient grounding:} Are all patient-specific claims explicitly supported by the observed pre-target trajectory, without importing held-out encounter facts or inventing symptoms, tests, diagnoses, treatments, or plans?\par
\textbf{Next-encounter relevance:} Does the reason explain why this code is likely to be documented in the immediately held-out admission, instead of merely defining the code, restating history, or asserting generic chronic carry-forward?\par
\textbf{ICD alignment:} Does the reason match the exact ICD entity and its level of specificity, rather than a neighboring diagnosis, symptom, complication, procedure, or broader code family?\par
\textbf{Evidential calibration:} Does the reason clearly separate observed patient facts, general medical knowledge or cited relations, and the forecast hypothesis; avoid treating association as patient-specific certainty; and express uncertainty proportionate to the available support? If citations are supplied, judge whether they entail the claim attributed to them. Citation absence alone must not lower this score when the trajectory itself provides adequate support.\\
\midrule
\cellcolor{SuppPurple}\textbf{Scale/output}
& Shared rubric: 5 complete, precise, and without a substantive flaw; 4 strong with one minor limitation; 3 plausible but partial or generic; 2 major gaps, weak support, or notable mismatch; 1 materially misleading; 0 absent, unusable, contradicted, or incompatible with the trajectory or exact ICD entity. Score dimensions independently, use the full scale, and do not infer system or model identity. Assess each listed true-positive reason exactly once and return JSON only. The strict response schema requires all five integers and one \texttt{concise\_reason} per opaque item ID.\\
\midrule
\cellcolor{SuppGray}\textbf{Dynamic payload/retry}
& Code system, observed pre-target context, complete gold set with titles, and independently shuffled anonymous true-positive reasons. The judge sees no method names. Validation failures append: ``Prior validation errors: \texttt{<feedback>}. Return corrected complete JSON.''\\
\bottomrule
\end{tabularx}
\end{table}

\clearpage

\section{Case Studies}
\label{app:case-studies}

The selected execution traces preserve the frozen row identifier, functional role, prediction outcome, and quoted or reconstructed output used in the analysis. Intervening candidate lists are marked when omitted for length. The cases illustrate candidate expansion, joint ranking, explanation alignment, and safety failures.
All explanation traces labeled ICD-Deepresearch are RRF-selected archival
outputs; they do not represent the full Evidence-Aware selector.

\setlength\LTleft{0pt}
\setlength\LTright{0pt}
\setlength\LTcapwidth{\textwidth}
\setlength{\tabcolsep}{4pt}
\renewcommand{\arraystretch}{1.08}
\begin{longtable}{>{\raggedright\arraybackslash}p{1.15cm}>{\raggedright\arraybackslash}p{2.55cm}>{\raggedright\arraybackslash}p{13.0cm}}
\multicolumn{3}{>{\raggedright\arraybackslash}p{16.7cm}}{\small\textbf{Table \thetable.} Selected qualitative forecasting and explanation traces. Panel A compares ICD-Deepresearch: Direct Forecast Only with ICD-Deepresearch (RRF-selected) on a shared true-positive code; the latter has the legacy label AF/Union+RRF in the saved records. Panels B--C follow productive and failed Research Expansion; Panel D shows how Direct Forecast Only, Research Path Only, and ICD-Deepresearch w/o Evidence-Aware Reranker contribute different true positives. Metrics and gold membership are revealed only in the evaluation rows.\label{tab:s-qualitative-traces}}\\[0.7ex]
\toprule
\textbf{Trace} & \textbf{Function / view} & \textbf{Selected output and interpretation}\\
\midrule
\endfirsthead
\multicolumn{3}{l}{\small\itshape Supplementary Table \thetable\ continued from the previous page.}\\
\toprule
\textbf{Trace} & \textbf{Function / view} & \textbf{Selected output and interpretation}\\
\midrule
\endhead
\midrule
\multicolumn{3}{r}{\small\itshape Continued on the next page.}\\
\endfoot
\bottomrule
\endlastfoot

\rowcolor{SuppBlue}
\multicolumn{3}{>{\raggedright\arraybackslash}p{16.7cm}}{\textbf{\textcolor{SuppBlueDark}{Panel A. Explanation case, MIMIC-IV row 45 --- correcting a fluent ICD-entity error}}}\\
\midrule
\rowcolor{SuppGray}
A0 & \textbf{Observed record}
& A 32-year-old woman has repeated pregnancy-related records containing uterine leiomyoma (D25.9), maternal care for benign uterine tumor in the second trimester (O34.12), and later third-trimester obstetric care. O34.13 is a held-out true positive; neither explanation generator saw the gold set.\\
\rowcolor{SuppOrange}
A1 & \textbf{GPT-5 (Direct Forecasting)}
& ``Maternal care for uterine scar from prior surgery was previously noted (O34.12), so a future third-trimester pregnancy could warrant this code if the scar is not specifically attributed to cesarean or is coded under the broader uterine-scar category.'' The explanation treats O34.12 as a uterine-scar code rather than maternal care for a benign uterine tumor.\\
\rowcolor{SuppGreen}
A2 & \textbf{ICD-Deepresearch (RRF-selected)}
& ``She has leiomyoma and previously required maternal care for a benign uterine tumor; if pregnant again into the third trimester with fibroids affecting management, O34.13 would apply [15].'' Source [15] resolves to the exact local ICD-10-CM title. The uncertainty statement correctly conditions the forecast on a future third-trimester pregnancy and active management.\\
\rowcolor{SuppPurple}
A3 & \textbf{Automated gold-aware Reason5 judge}
& ICD-Deepresearch (RRF-selected) receives clinical validity/patient grounding/next-encounter relevance/ICD alignment/evidential-calibration scores of 5/5/4/5/5; GPT-5 (Direct Forecasting) receives 0/1/1/0/1. After outputs are frozen, the method-blind GPT-5 judge receives the gold set, identifies the scar--tumor conflation, and assigns higher scores to the richer-input ICD-Deepresearch explanation.\\
\rowcolor{SuppGray}
A4 & \textbf{Interpretation}
& The ICD-Deepresearch writer receives the canonical entity represented by Source [15], while the observed obstetric trajectory supplies the patient and temporal bridge. This selected-case contrast does not isolate which additional input field causes the score difference.\\

\midrule
\rowcolor{SuppBlue}
\multicolumn{3}{>{\raggedright\arraybackslash}p{16.7cm}}{\textbf{\textcolor{SuppBlueDark}{Panel B. Productive expansion, MIMIC-IV row 52 --- four hits among eight additions}}}\\
\midrule
\rowcolor{SuppGray}
B0 & \textbf{Research Round 1 state}
& The Candidate Generation Only $K=20$ prediction matches 5/21 held-out codes. Research Round 1 expands the set to 50 candidates and 11 held-out matches, raising natural recall from .238 to .524.\\
\rowcolor{SuppGray}
B1 & \textbf{Research Round 2 planner}
& Selected relation queries include \texttt{Waldenstrom macroglobulinemia relapse progression outcomes}, \texttt{pancytopenia chemotherapy infection complications}, \texttt{hypernatremia dehydration neurologic complication timing}, and \texttt{hypomagnesemia arrhythmia complication incidence}.\\
\rowcolor{SuppGreen}
B2 & \textbf{Mapper and judge}
& Eight candidates are accepted. I50.32 chronic diastolic heart failure, I11.0 hypertensive heart disease with heart failure, R73.9 hyperglycemia, and R91.1 solitary pulmonary nodule are held-out matches. G62.0, N05.8, T38.0X5A, and D73.89 are false additions. The saved discovery-evidence list is empty, identifying this as a candidate-construction and validation trace.\\
\rowcolor{SuppGreen}
B3 & \textbf{Explanation writer}
& For I50.32: ``Explicitly documented in recent history alongside hypertension and hypertensive heart disease, indicating a persistent chronic HFpEF condition that is likely to be redocumented [2].'' The uncertainty notes that reassessment or quiescence may alter documentation. Citation [2] defines the code; the persistence claim is attributed to the pre-target record.\\
\rowcolor{SuppPurple}
B4 & \textbf{Evaluator}
& The final natural set has 15/21 held-out codes: P/R/F1=.259/.714/.380. The 4/8 marginal hit rate improves F1. Direct Forecast Only, Research Path Only, and ICD-Deepresearch w/o Evidence-Aware Reranker each recover 11/21 at $K=20$, illustrating the distinction between natural-cardinality expansion and fixed-budget ranking.\\

\midrule
\rowcolor{SuppBlue}
\multicolumn{3}{>{\raggedright\arraybackslash}p{16.7cm}}{\textbf{\textcolor{SuppBlueDark}{Panel C. Failed expansion, MIMIC-IV row 59 --- credible relations attached to invented anchors}}}\\
\midrule
\rowcolor{SuppGray}
C0 & \textbf{Research Round 2 planner}
& The initial plan concerns hepatitis progression, COPD recurrence, migraine chronicity, autonomic dysfunction, substance relapse, stimulant complications, and constipation. The accepted candidates later drift toward lipids, GERD, back pain, diabetes, and hypertension.\\
\rowcolor{SuppOrange}
C1 & \textbf{Mapper and judge}
& E78.5 is accepted from the claimed anchor ``LDL 162 mg/dL; atorvastatin started and continued'' plus readable PMC articles on long-term statin persistence. The structured patient input contains no LDL, medication, or visit text. The sources support population treatment persistence, not this patient's premise.\\
\rowcolor{SuppOrange}
C2 & \textbf{Mapper and judge}
& K21.9 uses an absent ``heartburn $>$3 months; daily omeprazole'' anchor despite a medically relevant GERD natural-history source. M54.5 uses an absent multi-year low-back-pain history despite a review of recurrent back pain. In both cases a valid relation is attached to an unresolved patient claim.\\
\multicolumn{3}{>{\raggedright\arraybackslash}p{16.7cm}}{\itshape Six further false additions are omitted for length: J02.9, E66.09, R07.9, E11.9, I10, and G47.00. Several rely only on dictionary definitions or reverse the immediate-horizon direction.}\\
\rowcolor{SuppPurple}
C3 & \textbf{Evaluator}
& All nine Research Round 2 additions are absent from the held-out encounter. Natural recall remains .846, while precision falls .224$\rightarrow$.190 and F1 falls .355$\rightarrow$.310 as set size grows 49$\rightarrow$58.\\
\rowcolor{SuppGray}
C4 & \textbf{Required correction}
& Resolve every candidate anchor to an immutable event identifier or exact pre-target record span before search, then verify source entailment and immediate-horizon direction as separate checks.\\

\midrule
\rowcolor{SuppBlue}
\multicolumn{3}{>{\raggedright\arraybackslash}p{16.7cm}}{\textbf{\textcolor{SuppBlueDark}{Panel D. Direct forecast and RRF-fusion case, MIMIC-IV row 62 --- complementary rank paths}}}\\
\midrule
\rowcolor{SuppOrange}
D0 & \textbf{ICD-Deepresearch: Direct Forecast Only}
& E27.40 adrenocortical insufficiency is ranked 19th by Direct Forecast Only but 25th by Research Path Only. Its saved reason states that the diagnosis appears repeatedly and may be coded again while acknowledging that imminent adrenal decompensation is not established. Direct Forecast Only obtains 5 true positives and F1@20=.286.\\
\rowcolor{SuppGreen}
D1 & \textbf{ICD-Deepresearch: Research Path Only}
& F32.9 major depression and Z87.891 personal history of nicotine dependence are ranked 5th and 10th by research but only 28th and 27th by Direct Forecast Only. Research Path Only obtains 6 true positives and F1@20=.343.\\
\rowcolor{SuppPurple}
D2 & \textbf{ICD-Deepresearch w/o Evidence-Aware Reranker}
& E27.40 enters through the direct rank; F32.9 and Z87.891 enter through research; E87.0 hyperosmolality/hypernatremia is promoted to rank 20 from moderate component ranks 25/23 by agreement.\\
\rowcolor{SuppGray}
D3 & \textbf{Evaluator}
& ICD-Deepresearch w/o Evidence-Aware Reranker obtains 8 true positives and P/R/F1=.400/.533/.457. The trace illustrates candidate and rank complementarity in this case.\\
\end{longtable}

\clearpage
\subsection{Complete Paired Explanation Output}
\label{app:complete-explanation-case}

Table~\ref{tab:s-complete-explanations} reproduces the complete $K=20$ explanation output for GPT-5 (Direct Forecasting) and ICD-Deepresearch (RRF-selected) in the selected MIMIC-IV case. Wording is preserved from the saved records; only Unicode quotation marks and dash characters are normalized for \LaTeX. TP and FP are post-hoc annotations added after generation. Both explanation writers use \texttt{gpt-5-2025-08-07}; the supplied prediction and evidence states differ.

\setlength\LTleft{0pt}
\setlength\LTright{0pt}
\setlength\LTcapwidth{\textwidth}
\setlength{\tabcolsep}{4pt}
\renewcommand{\arraystretch}{1.07}
\small
\begin{longtable}{>{\raggedright\arraybackslash}p{0.85cm}>{\raggedright\arraybackslash}p{4.05cm}>{\raggedright\arraybackslash}p{11.8cm}}
\multicolumn{3}{>{\raggedright\arraybackslash}p{16.7cm}}{\small\textbf{Table \thetable.} Complete explanation outputs for MIMIC-IV row 45. The exact same pre-target record is used by GPT-5 (Direct Forecasting) and ICD-Deepresearch (RRF-selected), while the ICD-Deepresearch writer also receives canonical code information, candidate clues, and ledger fields; the held-out target contains 15 ICD-10-CM codes. Every one of the 20 code-level rationales and uncertainty statements is shown. The ICD-Deepresearch source IDs are expanded in place so that each citation is inspectable.\label{tab:s-complete-explanations}}\\[0.7ex]
\toprule
\textbf{ID} & \textbf{Predicted code / outcome} & \textbf{Complete explanation output}\\
\midrule
\endfirsthead
\multicolumn{3}{l}{\small\itshape Supplementary Table \thetable\ continued from the previous page.}\\
\toprule
\textbf{ID} & \textbf{Predicted code / outcome} & \textbf{Complete explanation output}\\
\midrule
\endhead
\midrule
\multicolumn{3}{r}{\small\itshape Continued on the next page.}\\
\endfoot
\bottomrule
\endlastfoot

\rowcolor{SuppBlue}
\multicolumn{3}{>{\raggedright\arraybackslash}p{16.7cm}}{\textbf{\textcolor{SuppBlueDark}{Panel A. Frozen input and post-hoc target}}}\\*
\midrule
\rowcolor{SuppGray}
A0 & \textbf{Pre-target history} &
Age 32, female. Visit 1: \texttt{O34.32}, \texttt{D25.9}, \texttt{D57.3}, \texttt{Z3A.26}, \texttt{Z37.0}, \texttt{O32.8XX0}, \texttt{O26.852}, \texttt{O42.912}, and \texttt{O34.12}; procedure \texttt{10D00Z1}. Visit 2: \texttt{O86.12}, \texttt{O86.89}, \texttt{R78.81}, and \texttt{B95.61}. Visit 3: \texttt{O34.33}, \texttt{O26.893}, \texttt{Z3A.36}, \texttt{O99.013}, and \texttt{D57.3}.\\
\rowcolor{SuppPurple}
A1 & \textbf{Held-out gold} &
\texttt{O34.211}, \texttt{Z37.0}, \texttt{N85.8}, \texttt{O69.81X0}, \texttt{O32.6XX0}, \texttt{O70.1}, \texttt{O34.13}, \texttt{D25.1}, \texttt{D25.2}, \texttt{O99.03}, \texttt{D57.3}, \texttt{O76}, \texttt{O99.89}, \texttt{R55}, \texttt{Z3A.38}. The gold set is disclosed only for this retrospective case audit.\\

\midrule
\rowcolor{SuppBlue}
\multicolumn{3}{>{\raggedright\arraybackslash}p{16.7cm}}{\textbf{\textcolor{SuppBlueDark}{Panel B. GPT-5 (Direct Forecasting): all 20 explanations}}}\\*
\midrule
\rowcolor{SuppOrange}
D01 & \textbf{\texttt{O34.33}}\par Maternal care for cervical incompetence, third trimester\par \textbf{Post-hoc: FP} &
\textbf{Rationale.} Prior records show maternal care for scar from previous cesarean (O34.32, O34.33) and a documented cesarean delivery, so a future third-trimester pregnancy could require care for a prior C-section scar.\par
\textbf{Uncertainty.} Future pregnancy and trimester at the next admission are unknown, so this condition may not be applicable.\par
\textbf{Evidence returned.} None (the frozen output contains an empty citation-ID list).\\
\midrule
\rowcolor{SuppGray}
D02 & \textbf{\texttt{Z98.891}}\par History of uterine scar from previous surgery\par \textbf{Post-hoc: FP} &
\textbf{Rationale.} History includes a cesarean section (procedure 10D00Z1) and repeated documentation of uterine/cesarean scar, supporting a personal history of uterine scar from prior surgery.\par
\textbf{Uncertainty.} Whether the next admission will document or require coding of this historical condition is uncertain.\par
\textbf{Evidence returned.} None (the frozen output contains an empty citation-ID list).\\
\midrule
\rowcolor{SuppOrange}
D03 & \textbf{\texttt{D25.9}}\par Leiomyoma of uterus, unspecified\par \textbf{Post-hoc: FP} &
\textbf{Rationale.} Leiomyoma of uterus (D25.9) is documented previously and typically persists, making it reasonable to appear again on a subsequent admission if relevant to care.\par
\textbf{Uncertainty.} The presence or clinical relevance of fibroids at the next encounter is not guaranteed.\par
\textbf{Evidence returned.} None (the frozen output contains an empty citation-ID list).\\
\midrule
\rowcolor{SuppGray}
D04 & \textbf{\texttt{O34.13}}\par Maternal care for benign tumor of corpus uteri, third trimester\par \textbf{Post-hoc: TP} &
\textbf{Rationale.} Maternal care for uterine scar from prior surgery was previously noted (O34.12), so a future third-trimester pregnancy could warrant this code if the scar is not specifically attributed to cesarean or is coded under the broader uterine-scar category.\par
\textbf{Uncertainty.} It is unclear whether a future pregnancy will occur or whether coding will favor the cesarean-scar-specific code instead.\par
\textbf{Evidence returned.} None (the frozen output contains an empty citation-ID list).\\
\midrule
\rowcolor{SuppOrange}
D05 & \textbf{\texttt{D57.3}}\par Sickle-cell trait\par \textbf{Post-hoc: TP} &
\textbf{Rationale.} Sickle-cell trait (D57.3) is repeatedly documented, a chronic lifelong condition likely to be recoded if pertinent to a future admission or pregnancy.\par
\textbf{Uncertainty.} It may not be assessed or clinically relevant during the next admission.\par
\textbf{Evidence returned.} None (the frozen output contains an empty citation-ID list).\\
\midrule
\rowcolor{SuppGray}
D06 & \textbf{\texttt{O99.019}}\par Anemia complicating pregnancy, unspecified trimester\par \textbf{Post-hoc: FP} &
\textbf{Rationale.} Anemia complicating pregnancy in the third trimester (O99.013) was recorded previously, so anemia complicating pregnancy (unspecified trimester) could recur or be documented if trimester is not specified.\par
\textbf{Uncertainty.} Next admission may not involve pregnancy or anemia, and documentation specificity may differ.\par
\textbf{Evidence returned.} None (the frozen output contains an empty citation-ID list).\\
\midrule
\rowcolor{SuppOrange}
D07 & \textbf{\texttt{D50.9}}\par Iron deficiency anemia, unspecified\par \textbf{Post-hoc: FP} &
\textbf{Rationale.} Anemia in pregnancy was noted previously; iron-deficiency anemia (D50.9) is a common underlying cause that might be identified or recoded on a subsequent admission.\par
\textbf{Uncertainty.} No direct prior code for iron deficiency exists in the record, so etiology at the next admission is uncertain.\par
\textbf{Evidence returned.} None (the frozen output contains an empty citation-ID list).\\
\midrule
\rowcolor{SuppGray}
D08 & \textbf{\texttt{O32.8XX0}}\par Maternal care for other malpresentation of fetus, not applicable or unspecified\par \textbf{Post-hoc: FP} &
\textbf{Rationale.} Prior pregnancy included fetal malpresentation (O32.8XX0), which can recur in later pregnancies and prompt maternal care for malpresentation.\par
\textbf{Uncertainty.} Future pregnancy and fetal presentation are unknown.\par
\textbf{Evidence returned.} None (the frozen output contains an empty citation-ID list).\\
\midrule
\rowcolor{SuppOrange}
D09 & \textbf{\texttt{O42.90}}\par Premature rupture of membranes, unspecified as to length of time between rupture and onset of labor, unspecified weeks of gestation\par \textbf{Post-hoc: FP} &
\textbf{Rationale.} History includes premature rupture of membranes (O42.912); PROM can recur in subsequent pregnancies and may be coded with unspecified timing if details are unclear.\par
\textbf{Uncertainty.} Whether a future pregnancy will have PROM, and whether timing will be documented, is uncertain.\par
\textbf{Evidence returned.} None (the frozen output contains an empty citation-ID list).\\
\midrule
\rowcolor{SuppGray}
D10 & \textbf{\texttt{O85}}\par Puerperal sepsis\par \textbf{Post-hoc: FP} &
\textbf{Rationale.} Postpartum infections including endometritis (O86.12), other puerperal infection (O86.89), and bacteremia (R78.81) with MSSA (B95.61) were documented, indicating risk for severe puerperal infection such as puerperal sepsis in a future delivery admission.\par
\textbf{Uncertainty.} Sepsis may not occur in a subsequent puerperium, and infection severity cannot be predicted from prior history alone.\par
\textbf{Evidence returned.} None (the frozen output contains an empty citation-ID list).\\
\midrule
\rowcolor{SuppOrange}
D11 & \textbf{\texttt{B95.61}}\par Methicillin susceptible Staphylococcus aureus infection as the cause of diseases classified elsewhere\par \textbf{Post-hoc: FP} &
\textbf{Rationale.} Prior records show bacteremia and obstetric sepsis with likely genitourinary source; E. coli is a common etiologic organism for postpartum and pregnancy-related UTIs/wound infections, so a related organism code could recur on readmission.\par
\textbf{Uncertainty.} No culture results or organism documentation for the future admission are available to confirm E. coli causation.\par
\textbf{Evidence returned.} None (the frozen output contains an empty citation-ID list).\\
\midrule
\rowcolor{SuppGray}
D12 & \textbf{\texttt{R78.81}}\par Bacteremia\par \textbf{Post-hoc: FP} &
\textbf{Rationale.} History includes bacteremia during the peripartum period, placing the patient at risk for recurrence or evaluation for bloodstream infection on a subsequent admission.\par
\textbf{Uncertainty.} No future blood culture data or systemic infection signs are provided to verify bacteremia.\par
\textbf{Evidence returned.} None (the frozen output contains an empty citation-ID list).\\
\midrule
\rowcolor{SuppOrange}
D13 & \textbf{\texttt{O23.40}}\par Unspecified infection of urinary tract in pregnancy, unspecified trimester\par \textbf{Post-hoc: FP} &
\textbf{Rationale.} Pregnancy at 26 and 36 weeks was documented, and UTIs are common in pregnancy, especially with prior sepsis/possible E. coli, supporting a potential coding of genitourinary infection complicating pregnancy (unspecified trimester).\par
\textbf{Uncertainty.} Trimester and clinical confirmation of a pregnancy-associated UTI at the next visit are not documented.\par
\textbf{Evidence returned.} None (the frozen output contains an empty citation-ID list).\\
\midrule
\rowcolor{SuppGray}
D14 & \textbf{\texttt{N39.0}}\par Urinary tract infection, site not specified\par \textbf{Post-hoc: FP} &
\textbf{Rationale.} Given prior bacteremia and likely genitourinary source, a nonspecific urinary tract infection could be the focus of a subsequent admission.\par
\textbf{Uncertainty.} No urinalysis, culture, or urinary symptoms from the future encounter are provided.\par
\textbf{Evidence returned.} None (the frozen output contains an empty citation-ID list).\\
\midrule
\rowcolor{SuppOrange}
D15 & \textbf{\texttt{Z37.0}}\par Single live birth\par \textbf{Post-hoc: TP} &
\textbf{Rationale.} The chart shows late pregnancy (36 weeks) and prior singleton live birth; the next/final admission could be for another delivery resulting in a single liveborn outcome.\par
\textbf{Uncertainty.} Actual delivery outcome and plurality for the upcoming admission are unknown.\par
\textbf{Evidence returned.} None (the frozen output contains an empty citation-ID list).\\
\midrule
\rowcolor{SuppGray}
D16 & \textbf{\texttt{O26.899}}\par Other specified pregnancy related conditions, unspecified trimester\par \textbf{Post-hoc: FP} &
\textbf{Rationale.} Patient had multiple pregnancy-related complications coded previously (e.g., O26.893/O26.852), so another nonspecific pregnancy-related condition may be coded on a later admission when details are limited.\par
\textbf{Uncertainty.} The specific nature and timing of any new pregnancy-related complication are not provided.\par
\textbf{Evidence returned.} None (the frozen output contains an empty citation-ID list).\\
\midrule
\rowcolor{SuppOrange}
D17 & \textbf{\texttt{O26.90}}\par Pregnancy related conditions, unspecified, unspecified trimester\par \textbf{Post-hoc: FP} &
\textbf{Rationale.} Given recurring pregnancy complication codes in prior encounters, an unspecified pregnancy-related condition code may again be used if the condition is present but not further specified on admission.\par
\textbf{Uncertainty.} No explicit clinical problem is described to justify an unspecified pregnancy complication at the future visit.\par
\textbf{Evidence returned.} None (the frozen output contains an empty citation-ID list).\\
\midrule
\rowcolor{SuppGray}
D18 & \textbf{\texttt{Z3A.36}}\par 36 weeks gestation of pregnancy\par \textbf{Post-hoc: FP} &
\textbf{Rationale.} Gestational age of 36 weeks was recorded in late pregnancy; a near-term readmission around the same gestation often carries a Z3A36 dating code.\par
\textbf{Uncertainty.} Exact gestational age at the time of the next/final admission is not known.\par
\textbf{Evidence returned.} None (the frozen output contains an empty citation-ID list).\\
\midrule
\rowcolor{SuppOrange}
D19 & \textbf{\texttt{O86.00}}\par Infection of obstetric surgical wound, unspecified\par \textbf{Post-hoc: FP} &
\textbf{Rationale.} Previous cesarean delivery and puerperal sepsis increase risk for postpartum/obstetric surgical wound infection, which may prompt readmission and coding of an unspecified obstetric wound infection.\par
\textbf{Uncertainty.} No wound findings or postoperative exam from the future admission confirm a surgical site infection.\par
\textbf{Evidence returned.} None (the frozen output contains an empty citation-ID list).\\
\midrule
\rowcolor{SuppGray}
D20 & \textbf{\texttt{O90.81}}\par Anemia of the puerperium\par \textbf{Post-hoc: FP} &
\textbf{Rationale.} Following cesarean delivery and puerperal complications, additional specified puerperal complications may occur and require admission, aligning with an O90.81-type postpartum complication code.\par
\textbf{Uncertainty.} The specific postpartum condition, timing, and diagnostic evidence for O90.81 at the next encounter are not detailed.\par
\textbf{Evidence returned.} None (the frozen output contains an empty citation-ID list).\\

\midrule
\rowcolor{SuppBlue}
\multicolumn{3}{>{\raggedright\arraybackslash}p{16.7cm}}{\textbf{\textcolor{SuppBlueDark}{Panel C. ICD-Deepresearch (RRF-selected): all 20 explanations}}}\\*
\midrule
\rowcolor{SuppGreen}
R01 & \textbf{\texttt{D25.9}}\par Leiomyoma of uterus, unspecified\par \textbf{Post-hoc: FP} &
\textbf{Rationale.} She previously carried uterine leiomyoma (fibroids), which can remain clinically relevant across pregnancies and postpartum care; documentation of leiomyoma again would align with D25.9 if noted on the next admission [1].\par
\textbf{Uncertainty.} Only applicable if fibroids are re-documented as an active condition or influencing care on the next encounter.\par
\textbf{Source [1].} D25.9 Leiomyoma of uterus, unspecified; \url{https://www.icd10data.com/search?s=D25.9}. \textit{Recorded role: icd\_definition.}\\
\midrule
\rowcolor{SuppGray}
R02 & \textbf{\texttt{O34.33}}\par Maternal care for cervical incompetence, third trimester\par \textbf{Post-hoc: FP} &
\textbf{Rationale.} She had prior care for cervical incompetence (second trimester O34.32) and later third-trimester care (O34.33) was recorded, so a future pregnancy complicated by cervical incompetence in the third trimester could warrant this code if managed again [2].\par
\textbf{Uncertainty.} Requires a subsequent pregnancy with documented cervical insufficiency care in the third trimester.\par
\textbf{Source [2].} O34.33 Maternal care for cervical incompetence, third trimester; \url{https://www.icd10data.com/search?s=O34.33}. \textit{Recorded role: icd\_definition.}\\
\midrule
\rowcolor{SuppGreen}
R03 & \textbf{\texttt{R78.81}}\par Bacteremia\par \textbf{Post-hoc: FP} &
\textbf{Rationale.} Bacteremia was documented previously, and a new episode would be coded again if blood cultures show bacteria in the bloodstream on a future admission [3].\par
\textbf{Uncertainty.} Only coded if recurrent or new bloodstream bacteremia is documented.\par
\textbf{Source [3].} R78.81 Bacteremia; \url{https://www.icd10data.com/search?s=R78.81}. \textit{Recorded role: icd\_definition.}\\
\midrule
\rowcolor{SuppGray}
R04 & \textbf{\texttt{B95.61}}\par Methicillin susceptible Staphylococcus aureus infection as the cause of diseases classified elsewhere\par \textbf{Post-hoc: FP} &
\textbf{Rationale.} MSSA (B95.61) was previously identified as the causative organism; if a future infection is attributed to MSSA, this organism code would again accompany the primary infectious diagnosis [4].\par
\textbf{Uncertainty.} Contingent on a new infection being microbiologically linked to MSSA.\par
\textbf{Source [4].} B95.61 Methicillin susceptible Staphylococcus aureus infection as the cause of diseases classified elsewhere; \url{https://www.icd10data.com/search?s=B95.61}. \textit{Recorded role: icd\_definition.}\\
\midrule
\rowcolor{SuppGreen}
R05 & \textbf{\texttt{D57.3}}\par Sickle-cell trait\par \textbf{Post-hoc: TP} &
\textbf{Rationale.} Sickle-cell trait was documented on multiple encounters and is a lifelong hematologic status that commonly remains on problem lists or prenatal records when relevant [5].\par
\textbf{Uncertainty.} Will appear only if clinicians carry forward or reconfirm the trait as pertinent to care.\par
\textbf{Source [5].} D57.3 Sickle-cell trait; \url{https://www.icd10data.com/search?s=D57.3}. \textit{Recorded role: icd\_definition.}\\
\midrule
\rowcolor{SuppGray}
R06 & \textbf{\texttt{Z37.0}}\par Single live birth\par \textbf{Post-hoc: TP} &
\textbf{Rationale.} She previously had a single live birth; if the next admission is a delivery resulting in one liveborn infant, Z37.0 would document that outcome again [6].\par
\textbf{Uncertainty.} Applicable only if the next hospitalization is for delivery with a singleton liveborn outcome.\par
\textbf{Source [6].} Z37.0 Single live birth; \url{https://www.icd10data.com/search?s=Z37.0}. \textit{Recorded role: icd\_definition.}\\
\midrule
\rowcolor{SuppGreen}
R07 & \textbf{\texttt{O32.8XX0}}\par Maternal care for other malpresentation of fetus, not applicable or unspecified\par \textbf{Post-hoc: FP} &
\textbf{Rationale.} Prior records included antenatal care for fetal malpresentation; recurrence of a non-specific malpresentation in a future pregnancy could prompt this unspecified malpresentation care code [7].\par
\textbf{Uncertainty.} Depends on a future pregnancy with documented malpresentation requiring maternal care.\par
\textbf{Source [7].} O32.8XX0 Maternal care for other malpresentation of fetus, not applicable or unspecified; \url{https://www.icd10data.com/search?s=O32.8XX0}. \textit{Recorded role: icd\_definition.}\\
\midrule
\rowcolor{SuppGray}
R08 & \textbf{\texttt{Z3A.36}}\par 36 weeks gestation of pregnancy\par \textbf{Post-hoc: FP} &
\textbf{Rationale.} She was previously seen at 36 weeks' gestation; if a future encounter occurs at 36 weeks, gestational age would again be captured with Z3A.36 [8].\par
\textbf{Uncertainty.} Only used if the timing of a subsequent pregnancy admission is specifically at 36 weeks.\par
\textbf{Source [8].} Z3A.36 36 weeks gestation of pregnancy; \url{https://www.icd10data.com/search?s=Z3A.36}. \textit{Recorded role: icd\_definition.}\\
\midrule
\rowcolor{SuppGreen}
R09 & \textbf{\texttt{O85}}\par Puerperal sepsis\par \textbf{Post-hoc: FP} &
\textbf{Rationale.} She had postpartum infections and bacteremia after delivery in prior care; if a future postpartum admission meets criteria for puerperal sepsis, O85 would capture that complication [9].\par
\textbf{Uncertainty.} Requires new postpartum sepsis; prior infection alone does not ensure recurrence.\par
\textbf{Source [9].} O85 Puerperal sepsis; \url{https://www.icd10data.com/search?s=O85}. \textit{Recorded role: icd\_definition.}\\
\midrule
\rowcolor{SuppGray}
R10 & \textbf{\texttt{O86.12}}\par Endometritis following delivery\par \textbf{Post-hoc: FP} &
\textbf{Rationale.} Endometritis following delivery was previously coded; a subsequent postpartum endometritis episode after a future birth would again be captured with O86.12 [12].\par
\textbf{Uncertainty.} Only applicable if a new post-delivery endometritis is documented.\par
\textbf{Source [12].} O86.12 Endometritis following delivery; \url{https://www.icd10data.com/search?s=O86.12}. \textit{Recorded role: icd\_definition.}\\
\midrule
\rowcolor{SuppGreen}
R11 & \textbf{\texttt{O86.89}}\par Other specified puerperal infections\par \textbf{Post-hoc: FP} &
\textbf{Rationale.} Other specified puerperal infections were in her history; a future postpartum infection not classified elsewhere could be coded again with O86.89 [13].\par
\textbf{Uncertainty.} Requires documentation of a new specified puerperal infection postpartum.\par
\textbf{Source [13].} O86.89 Other specified puerperal infections; \url{https://www.icd10data.com/search?s=O86.89}. \textit{Recorded role: icd\_definition.}\\
\midrule
\rowcolor{SuppGray}
R12 & \textbf{\texttt{Z98.891}}\par History of uterine scar from previous surgery\par \textbf{Post-hoc: FP} &
\textbf{Rationale.} History of a uterine surgical scar (e.g., prior cesarean or myomectomy) is a persistent status; if such a scar is documented, this status code may be carried forward on future obstetric admissions [14].\par
\textbf{Uncertainty.} Only appropriate if records confirm a prior uterine surgery resulting in a scar.\par
\textbf{Source [14].} Z98.891 History of uterine scar from previous surgery; \url{https://www.icd10data.com/search?s=Z98.891}. \textit{Recorded role: icd\_definition.}\\
\midrule
\rowcolor{SuppGreen}
R13 & \textbf{\texttt{O34.13}}\par Maternal care for benign tumor of corpus uteri, third trimester\par \textbf{Post-hoc: TP} &
\textbf{Rationale.} She has leiomyoma and previously required maternal care for a benign uterine tumor; if pregnant again into the third trimester with fibroids affecting management, O34.13 would apply [15].\par
\textbf{Uncertainty.} Depends on third-trimester pregnancy with active management implications from fibroids.\par
\textbf{Source [15].} O34.13 Maternal care for benign tumor of corpus uteri, third trimester; \url{https://www.icd10data.com/search?s=O34.13}. \textit{Recorded role: icd\_definition.}\\
\midrule
\rowcolor{SuppGray}
R14 & \textbf{\texttt{O99.013}}\par Anemia complicating pregnancy, third trimester\par \textbf{Post-hoc: FP} &
\textbf{Rationale.} Anemia complicating pregnancy, third trimester, was previously documented; recurrence in a future third-trimester pregnancy would again warrant O99.013 [16].\par
\textbf{Uncertainty.} Requires a subsequent pregnancy with third-trimester anemia impacting care.\par
\textbf{Source [16].} O99.013 Anemia complicating pregnancy, third trimester; \url{https://www.icd10data.com/search?s=O99.013}. \textit{Recorded role: icd\_definition.}\\
\midrule
\rowcolor{SuppGreen}
R15 & \textbf{\texttt{O99.019}}\par Anemia complicating pregnancy, unspecified trimester\par \textbf{Post-hoc: FP} &
\textbf{Rationale.} If anemia complicates a future pregnancy but trimester is not specified in documentation, O99.019 captures that unspecified-trimester complication [17].\par
\textbf{Uncertainty.} Used only if pregnancy-related anemia is present and the trimester is not documented.\par
\textbf{Source [17].} O99.019 Anemia complicating pregnancy, unspecified trimester; \url{https://www.icd10data.com/search?s=O99.019}. \textit{Recorded role: icd\_definition.}\\
\midrule
\rowcolor{SuppGray}
R16 & \textbf{\texttt{D50.9}}\par Iron deficiency anemia, unspecified\par \textbf{Post-hoc: FP} &
\textbf{Rationale.} If future evaluation shows iron deficiency as the cause of anemia, D50.9 would document unspecified iron deficiency anemia alongside pregnancy-related codes as appropriate [18].\par
\textbf{Uncertainty.} Applicable only with documentation or labs supporting iron deficiency anemia.\par
\textbf{Source [18].} D50.9 Iron deficiency anemia, unspecified; \url{https://www.icd10data.com/search?s=D50.9}. \textit{Recorded role: icd\_definition.}\\
\midrule
\rowcolor{SuppGreen}
R17 & \textbf{\texttt{O42.90}}\par Premature rupture of membranes, unspecified as to length of time between rupture and onset of labor, unspecified weeks of gestation\par \textbf{Post-hoc: FP} &
\textbf{Rationale.} Should a future pregnancy be complicated by premature rupture of membranes with unspecified latency and gestational week, O42.90 would capture that event [19].\par
\textbf{Uncertainty.} Contingent on clear PROM documentation meeting this unspecified description.\par
\textbf{Source [19].} O42.90 Premature rupture of membranes, unspecified as to length of time between rupture and onset of labor, unspecified weeks of gestation; \url{https://www.icd10data.com/search?s=O42.90}. \textit{Recorded role: icd\_definition.}\\
\midrule
\rowcolor{SuppGray}
R18 & \textbf{\texttt{O34.12}}\par Maternal care for benign tumor of corpus uteri, second trimester\par \textbf{Post-hoc: FP} &
\textbf{Rationale.} She previously had maternal care for a benign uterine tumor in the second trimester; recurrence in the second trimester of a future pregnancy could again be coded O34.12 [20].\par
\textbf{Uncertainty.} Requires a subsequent second-trimester pregnancy with fibroid-related management.\par
\textbf{Source [20].} O34.12 Maternal care for benign tumor of corpus uteri, second trimester; \url{https://www.icd10data.com/search?s=O34.12}. \textit{Recorded role: icd\_definition.}\\
\midrule
\rowcolor{SuppGreen}
R19 & \textbf{\texttt{O34.32}}\par Maternal care for cervical incompetence, second trimester\par \textbf{Post-hoc: FP} &
\textbf{Rationale.} Maternal care for cervical incompetence in the second trimester was recorded before; if cervical insufficiency again requires care in a future second trimester, O34.32 would apply [21].\par
\textbf{Uncertainty.} Only applicable with renewed documentation of second-trimester cervical incompetence care.\par
\textbf{Source [21].} O34.32 Maternal care for cervical incompetence, second trimester; \url{https://www.icd10data.com/search?s=O34.32}. \textit{Recorded role: icd\_definition.}\\
\midrule
\rowcolor{SuppGray}
R20 & \textbf{\texttt{O23.40}}\par Unspecified infection of urinary tract in pregnancy, unspecified trimester\par \textbf{Post-hoc: FP} &
\textbf{Rationale.} If a future pregnancy is complicated by a urinary tract infection without trimester specified, O23.40 would document that condition [22].\par
\textbf{Uncertainty.} Used only if a pregnancy-associated UTI is diagnosed and trimester remains unspecified.\par
\textbf{Source [22].} O23.40 Unspecified infection of urinary tract in pregnancy, unspecified trimester; \url{https://www.icd10data.com/search?s=O23.40}. \textit{Recorded role: icd\_definition.}\\
\end{longtable}
\normalsize

\begin{table}[H]
\centering
\footnotesize
\setlength{\tabcolsep}{4.5pt}
\renewcommand{\arraystretch}{1.13}
\caption{Gold-aware automated comparison of the complete row-45 explanation outputs. Reason5 values are means over the three shared true-positive codes (\texttt{O34.13}, \texttt{D57.3}, and \texttt{Z37.0}) under the frozen item-blind GPT-5 judge. The semantic audit counts explicit title--rationale contradictions visible in Table~\ref{tab:s-complete-explanations}; ICD-Deepresearch (RRF-selected) receives the complete code-bound explanation input.}
\label{tab:s-complete-explanation-analysis}
\begin{tabularx}{\textwidth}{>{\raggedright\arraybackslash}p{3.25cm}>{\raggedright\arraybackslash}p{4.25cm}>{\raggedright\arraybackslash}p{4.25cm}Y}
\toprule
\rowcolor{SuppBlue}
\textbf{\textcolor{SuppBlueDark}{Comparison axis}} &
\textbf{\textcolor{SuppBlueDark}{GPT-5 (Direct Forecasting)}} &
\textbf{\textcolor{SuppBlueDark}{ICD-Deepresearch (RRF-selected)}} &
\textbf{\textcolor{SuppBlueDark}{What the automated audit shows}}\\
\midrule
\cellcolor{SuppGray}\textbf{Prediction outcome}
& 3/20 hits: \texttt{O34.13}, \texttt{D57.3}, \texttt{Z37.0}. P/R/F1=15.00/20.00/17.14\%.
& 3/20 hits: \texttt{D57.3}, \texttt{Z37.0}, \texttt{O34.13}. P/R/F1=15.00/20.00/17.14\%.
& Prediction quality is tied in this case; explanation quality must therefore be evaluated separately.\\
\midrule
\cellcolor{SuppOrange}\textbf{Clinical validity}
& 3.33/5
& \textbf{5.00/5}
& The automated-score difference coincides with the Direct \texttt{O34.13} scar--tumor conflation; both configurations score 5/5 on the other two shared hits.\\
\midrule
\cellcolor{SuppGreen}\textbf{Patient grounding}
& 3.67/5
& \textbf{5.00/5}
& The richer-input ICD-Deepresearch explanation links \texttt{O34.13} to observed leiomyoma/\texttt{O34.12}; GPT-5 (Direct Forecasting) links it to an incorrectly inferred uterine-scar entity.\\
\midrule
\cellcolor{SuppPurple}\textbf{Next-encounter relevance}
& 3.33/5
& \textbf{4.00/5}
& Both outputs are conditional forecasts; the score difference reflects the specificity of the observed transition rationale.\\
\midrule
\cellcolor{SuppOrange}\textbf{ICD alignment}
& 3.33/5
& \textbf{5.00/5}
& GPT-5 (Direct Forecasting) misdescribes \texttt{O34.33}, \texttt{O34.13}, \texttt{B95.61}, and \texttt{O90.81}; the 20 ICD-Deepresearch rows preserve title--rationale consistency in this selected case.\\
\midrule
\cellcolor{SuppGreen}\textbf{Evidential calibration}
& 3.67/5
& \textbf{5.00/5}
& Code-bound source IDs make the asserted ICD entity and definition traceable.\\
\midrule
\cellcolor{SuppPurple}\textbf{Five-axis mean}
& 3.47/5
& \textbf{4.80/5}
& On the three shared hits, the frozen GPT-5 judge assigns the richer-input ICD-Deepresearch explanation a 1.33-point higher mean.\\
\midrule
\cellcolor{SuppGray}\textbf{Citation trace}
& 0/20 explanations return a source ID.
& 20/20 explanations return an inspectable code-bound source ID.
& The ICD-Deepresearch source IDs make code identity inspectable. In this case, all cited sources are local ICD definitions, so the citation trace is confined to code identity.\\
\midrule
\cellcolor{SuppOrange}\textbf{Remaining errors}
& 17/20 predictions are false positives; several explanations infer unobserved causes or future complications.
& 17/20 predictions are also false positives; many rationales remain conditional and weakly tied to the immediate encounter.
& The richer-input ICD-Deepresearch configuration receives higher automated
alignment and calibration scores, while both methods retain substantial
false-positive and horizon errors.\\
\bottomrule
\end{tabularx}
\end{table}

\paragraph{Case interpretation.}
GPT-5 (Direct Forecasting) produces internally contradictory entity descriptions: it calls \texttt{O34.33} and \texttt{O34.13} uterine-scar codes, describes \texttt{B95.61} as an \emph{E.\ coli} organism code, and treats \texttt{O90.81} as a generic puerperal complication. The ICD-Deepresearch (RRF-selected) writer receives the selected code title and exposes that title through a resolvable citation, connecting the observed fibroid trajectory to the correct \texttt{O34.13} entity. The automated, gold-aware, method-blind GPT-5 Reason5 judge assigns the three shared hits a higher five-axis mean (4.80 versus 3.47). Both systems recover the same three gold codes at $K=20$; the observed score difference is concentrated in code alignment and evidential calibration and does not isolate the effect of any single additional input field.

\begin{longtable}{>{\raggedright\arraybackslash}p{1.45cm}>{\raggedright\arraybackslash}p{2.75cm}>{\raggedright\arraybackslash}p{12.5cm}}
\multicolumn{3}{>{\raggedright\arraybackslash}p{16.7cm}}{\small\textbf{Table \thetable.} Selected paired retrieval traces from the metadata audit. Within each panel, Research Expansion and GPT-5 + Web Search (Standalone) operate on the same patient row. The rows reproduce recorded queries, source metadata, and metadata-proxy labels.\label{tab:s-source-paired-cases}}\\[0.7ex]
\toprule
\textbf{Case} & \textbf{Retriever / judge} & \textbf{Selected retrieval trace}\\
\midrule
\endfirsthead
\multicolumn{3}{l}{\small\itshape Supplementary Table \thetable\ continued from the previous page.}\\
\toprule
\textbf{Case} & \textbf{Retriever / judge} & \textbf{Selected retrieval trace}\\
\midrule
\endhead
\midrule
\multicolumn{3}{r}{\small\itshape Continued on the next page.}\\
\endfoot
\bottomrule
\endlastfoot

\rowcolor{SuppBlue}
\multicolumn{3}{>{\raggedright\arraybackslash}p{16.7cm}}{\textbf{\textcolor{SuppBlueDark}{Panel A. MIMIC-III row 48 --- newborn jaundice trajectory}}}\\
\midrule
\rowcolor{SuppGreen}
A1 & \textbf{Archived research-path metadata}
& \textbf{Query:} \texttt{bilirubin total peak newborn phototherapy}. \textbf{Source:} \emph{Guidelines for Phototherapy}, Stanford Newborn Nursery, \url{https://med.stanford.edu/newborns/professional-education/jaundice-and-phototherapy/guidelines-for-phototherapy.html}.\\
\rowcolor{SuppGreen}
A2 & \textbf{Archived proxy label}
& Credible=yes; patient-relevant=yes; forecast-useful=yes; coding-only=no; utility=direct. The metadata judge labels the phototherapy-threshold source as a possible near-term bridge from the query, title, URL, and pre-target record.\\
\rowcolor{SuppOrange}
A3 & \textbf{GPT-5 + Web Search (Standalone) metadata}
& \textbf{Query:} \texttt{ICD-9-CM 771.81 septicemia of newborn code definition | 771.82 urinary tract infection | V29.0 observation}. \textbf{Source:} Reddit Emergency Medicine thread, \url{https://www.reddit.com/r/emergencymedicine/comments/1jx5rdi}.\\
\rowcolor{SuppOrange}
A4 & \textbf{Archived proxy label}
& Credible=no; patient-relevant=no; forecast-useful=no; coding-only=no; utility=none. The metadata judge labels this forum source as lacking an authoritative next-visit bridge.\\

\midrule
\rowcolor{SuppBlue}
\multicolumn{3}{>{\raggedright\arraybackslash}p{16.7cm}}{\textbf{\textcolor{SuppBlueDark}{Panel B. MIMIC-IV row 78 --- diabetic foot-ulcer care}}}\\
\midrule
\rowcolor{SuppGreen}
B1 & \textbf{Archived research-path metadata}
& \textbf{Query:} \texttt{poor glycemic control ulcer healing delay}. \textbf{Source:} \emph{Reasonable Glycemic Control Would Help Wound Healing During the Treatment of Diabetic Foot Ulcers}, PMC, \url{https://pmc.ncbi.nlm.nih.gov/articles/PMC6349287/}.\\
\rowcolor{SuppGreen}
B2 & \textbf{Archived proxy label}
& Credible=yes; patient-relevant=yes; forecast-useful=yes; coding-only=no; utility=indirect. The metadata judge labels the proposed glycemic-control/wound-healing relation as potentially relevant.\\
\rowcolor{SuppOrange}
B3 & \textbf{GPT-5 + Web Search (Standalone) metadata}
& \textbf{Query:} \texttt{ICD-10-CM L97.418 definition | L97.419 definition | L02.611 definition | L03.115 definition}. \textbf{Source:} Reddit Phlebotomy thread, \url{https://www.reddit.com/r/phlebotomy/comments/1tk6dk6/can_somebody_please_help_me_understand_icd_codes/}.\\
\rowcolor{SuppOrange}
B4 & \textbf{Archived proxy label}
& Credible=no; patient-relevant=no; forecast-useful=no; coding-only=no; utility=none. The metadata judge labels this code-oriented forum source as lacking a concrete next-encounter transition.\\

\midrule
\rowcolor{SuppBlue}
\multicolumn{3}{>{\raggedright\arraybackslash}p{16.7cm}}{\textbf{\textcolor{SuppBlueDark}{Panel C. MIMIC-IV row 62 --- central-line surveillance}}}\\
\midrule
\rowcolor{SuppGreen}
C1 & \textbf{Archived research-path metadata}
& \textbf{Query:} \texttt{indwelling central venous catheter surveillance outcomes}. \textbf{Source:} \emph{A Two-Year Surveillance of Central Line-Associated Bloodstream Infections}, PMC, \url{https://pmc.ncbi.nlm.nih.gov/articles/PMC10577095/}.\\
\rowcolor{SuppGreen}
C2 & \textbf{Archived proxy label}
& Credible=yes; patient-relevant=yes; forecast-useful=yes; coding-only=no; utility=direct. The metadata judge labels catheter-infection surveillance as potentially relevant.\\
\rowcolor{SuppOrange}
C3 & \textbf{GPT-5 + Web Search (Standalone) metadata}
& \textbf{Query:} \texttt{ICD-10-CM hyperammonemia | metabolic encephalopathy G93.41 | esophageal varices | fluid overload E87.7}. \textbf{Source:} Reddit AskDocs thread, \url{https://www.reddit.com/r/AskDocs/comments/163f3rn}.\\
\rowcolor{SuppOrange}
C4 & \textbf{Archived proxy label}
& Credible=no; patient-relevant=no; forecast-useful=no; coding-only=no; utility=none. The metadata judge labels this forum source as lacking authoritative next-visit guidance.\\

\midrule
\rowcolor{SuppGray}
D & \textbf{Aggregate metadata audit}
& Across 50 sampled retrieval occurrences per system/dataset cell, the forecast-proxy label is positive for 22\% versus 6\% on MIMIC-III and 36\% versus 4\% on MIMIC-IV. The audit unit is an occurrence represented by its query, title, URL, and pre-target record.\\
\end{longtable}

\section{Cross-Case Error Analysis}
\label{sec:supp-error-analysis}

\begin{table}[H]
\centering
\footnotesize
\setlength{\tabcolsep}{5pt}
\renewcommand{\arraystretch}{1.12}
\caption{Failure mechanisms and corresponding validation requirements derived from the saved case traces.}
\label{tab:s-error-taxonomy}
\begin{tabularx}{\textwidth}{>{\raggedright\arraybackslash}p{2.35cm}>{\raggedright\arraybackslash}p{3.2cm}YY}
\toprule
\rowcolor{SuppBlue}
\textbf{\textcolor{SuppBlueDark}{Mechanism}} &
\textbf{\textcolor{SuppBlueDark}{Observed trace}} &
\textbf{\textcolor{SuppBlueDark}{Current validation gap}} &
\textbf{\textcolor{SuppBlueDark}{Required correction}}\\
\midrule
\cellcolor{SuppOrange}\textbf{Free-text anchor hallucination}
& Row 59 invents LDL, HbA1c, medications, BP values, and symptom visits absent from its structured input.
& The judge receives the candidate's anchor statement without an immutable source-span link.
& Candidate must return an event ID or exact quoted pre-target span; deterministic validation rejects unresolved anchors before search.\\
\midrule
\cellcolor{SuppGreen}\textbf{Query--candidate drift}
& Row 59's initial plan concerns hepatitis, COPD, migraine, autonomic dysfunction, substance relapse, and constipation; accepted additions later concern lipids, GERD, back pain, diabetes, and hypertension.
& Mapping is allowed to ``discover a supported consequence,'' weakening traceability to the planned hypothesis.
& Assign a hypothesis ID and require every candidate, query, source, and verdict to preserve that ID; new hypotheses must trigger a new planning pass.\\
\midrule
\cellcolor{SuppPurple}\textbf{Definition--forecast conflation}
& Row 52 true additions and several row 59 false additions cite dictionary pages as their only accepted evidence.
& Exact code validation is represented beside clinical evidence and may be read as additive support.
& Store \texttt{definition}, \texttt{patient\_anchor}, and \texttt{relation\_evidence} in separate channels; definitions contribute zero forecast confidence.\\
\midrule
\cellcolor{SuppOrange}\textbf{Semantic mapping mismatch}
& R73.9 is described as abnormal liver enzymes, G62.0 as alcoholic neuropathy, and D73.89 as anemia despite conflicting dictionary titles.
& Validation checks that a code exists, not that model term, hypothesis, anchor, and dictionary entity agree.
& Add bidirectional semantic entailment checks over code title $\leftrightarrow$ candidate term $\leftrightarrow$ patient anchor; reject mismatched site, cause, status, or entity.\\
\midrule
\cellcolor{SuppGray}\textbf{Horizon-polarity error}
& Row 59 accepts acute pharyngitis while its own hypothesis and sources describe a self-limited episode.
& The current relation label conflates persistence or recurrence with resolution.
& Verifier must label the relation as persistence, recurrence, emergence, resolution, or contraindication and require positive immediate-horizon polarity.\\
\midrule
\cellcolor{SuppGreen}\textbf{Uncontrolled cardinality}
& Row 52's 4/8 Research Round~2 yield improves F1; row 59's 0/9 Research Round~2 yield necessarily lowers F1.
& Candidate acceptance is local; the observed break-even statistic is available only after held-out targets are revealed.
& Learn any retention rule on disjoint development data, freeze it before evaluation, and compare it with fixed-budget controls; preserve failed-query and negative-evidence states for ranking.\\
\bottomrule
\end{tabularx}
\end{table}

The cases diagnose internal failure modes of Research Expansion and
support a three-link contract: \emph{resolvable pre-target anchor}
$\rightarrow$ \emph{source-entailed directional relation} $\rightarrow$
\emph{exact immediate-horizon ICD candidate}. Row 52 recovers candidates but
exposes mapping errors and definition-only support; row 59 breaks the anchor
and horizon links; and in row 62 the RRF selector retains true positives
available at complementary ranks in the two branches. The final observation
is a single-case illustration, not an estimate of a general RRF effect. The
taxonomy complements the component controls by locating errors in anchoring,
relation direction, code mapping, and retention.

\flushbottom

\end{document}